\documentclass{thesisclass}
\pdfoutput=1

\makeatletter

\renewcommand\paragraph{\@startsection{paragraph}{4}{\z@}%
                                    {3.25ex \@plus1ex \@minus.2ex}%
                                    {-1em}%
                                    {\normalfont\normalsize}}
\makeatother

\newcommand{\totalCollectedRecordings}{166898}  

\newcommand{\trainingsetsize}{134804}
\newcommand{\validtionsetsize}{15161}
\newcommand{\testsetsize}{17012}

\newcommand{\totalClassesAnalyzed}{369}
\newcommand{\totalClassesAboveFifty}{680}
\newcommand{\totalClassesNotAnalyzedBelowFifty}{431}
\newcommand{\detexifyPercentage}{$\SI{91.93}{\percent}$}
\newcommand{\recordingsWithDots}{$\SI{2.77}{\percent}$}  
\newcommand{\recordingsWithDotsSizechange}{$\SI{0.85}{\percent}$}  

\newcommand{\myname}{Martin Thoma}
\newcommand{\mytitle}{On-line Recognition of Handwritten Mathematical Symbols}
\newcommand{\myinstitute}{Institute for Anthropomatics and Robotics (IAR)}

\newcommand{\reviewerone}{Prof.\ Dr.\ Alexander Waibel}
\newcommand{\reviewertwo}{Dr.\ Sebastian Stüker}
\newcommand{\advisor}{Kevin Kilgour}
\newcommand{\advisortwo}{Prof.\ Dr.\ Florian Metze}

\newcommand{\timestart}{June 2014}
\newcommand{\timeend}{November 2014}
\newcommand{\submissiontime}{07.11.2014}
\newcommand{\submissionplace}{Karlsruhe}

\hypersetup{
    pdfauthor={\myname},
    pdftitle={\mytitle},
    pdfsubject={Handwriting recognition of mathematical symbols},
    pdfkeywords={Machine Learning; classification; neural networks}
}

\newlist{problemenumd}{enumerate}{1}
\setlist[problemenumd]{label=D\arabic{problemenumdi},ref=D\arabic{problemenumdi}}
\crefname{problemenumdi}{problem}{problems}
\Crefname{problemenumdi}{Problem}{Problems}

\newlist{problemenumh}{enumerate}{1}
\setlist[problemenumh]{label=H\arabic{problemenumhi},ref=H\arabic{problemenumhi}}
\crefname{problemenumhi}{problem}{problems}
\Crefname{problemenumhi}{Problem}{Problems}

\newlist{problemenumo}{enumerate}{1}
\setlist[problemenumo]{label=O\arabic{problemenumoi},ref=O\arabic{problemenumoi}}
\crefname{problemenumoi}{problem}{problems}
\Crefname{problemenumoi}{Problem}{Problems}

\DeclareRobustCommand{\inlinelist}[1]{\begin{inparaenum}[(1)] #1 \end{inparaenum}}

\VerbatimFootnotes%

\begin{document}
\selectlanguage{english}

\frontmatter
\pagenumbering{roman}

\newcommand{\diameter}{20}
\newcommand{\xone}{-15}
\newcommand{\xtwo}{160}
\newcommand{\yone}{15}
\newcommand{\ytwo}{-253}

\begin{titlepage}
\begin{tikzpicture}[overlay]
\draw[color=gray]
          (\xone mm, \yone mm)
  -- (\xtwo mm, \yone mm)
 arc (90:0:\diameter pt)
  -- (\xtwo mm + \diameter pt , \ytwo mm)
    -- (\xone mm + \diameter pt , \ytwo mm)
 arc (270:180:\diameter pt)
    -- (\xone mm, \yone mm);
\end{tikzpicture}
    \begin{textblock}{10}[0,0](4,2.5)
        \includegraphics[width=.3\textwidth]{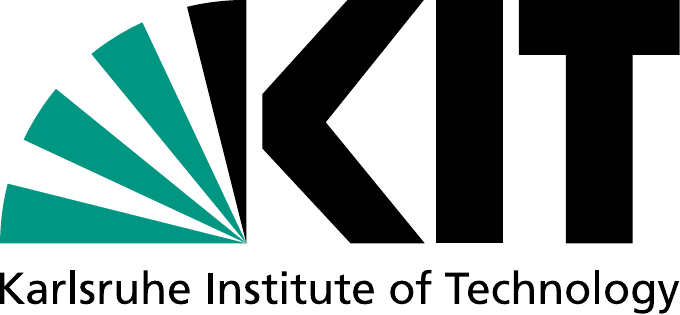}
    \end{textblock}
    \begin{textblock}{10}[0,0](13,2.5)
        \includegraphics[width=.3\textwidth]{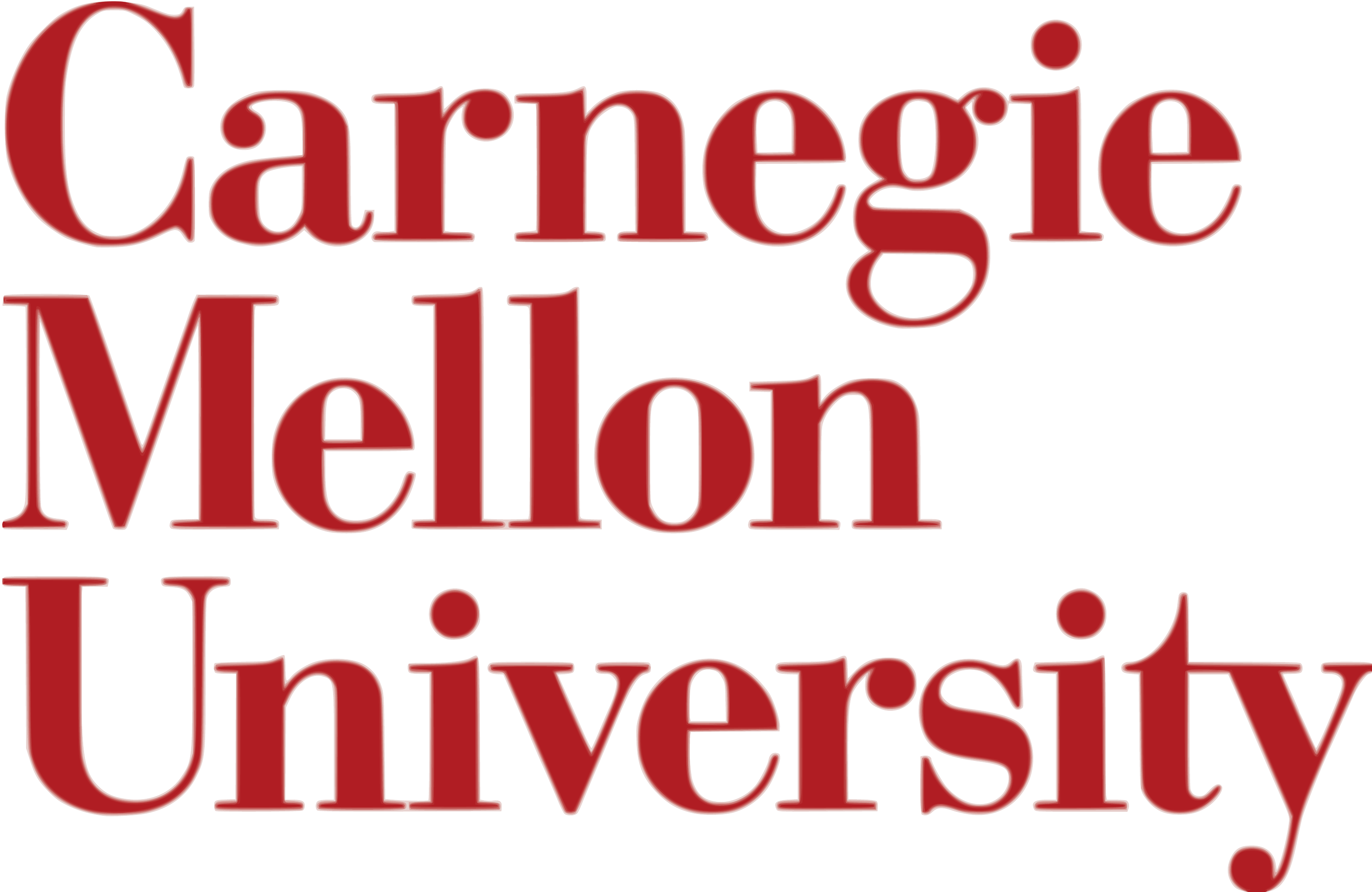}
    \end{textblock}
    \changefont{phv}{m}{n}
    \vspace*{3.5cm}
    \begin{center}
        \Huge{\mytitle}
        \vspace*{2cm}\\
        \Large{
            \iflanguage{english}{Bachelor's Thesis of}
                                                  {Bachelorarbeit\\von}
        }\\
        \vspace*{1cm}
        \huge{\myname}\\
        \vspace*{1cm}
        \Large{
            \iflanguage{english}{At the Department of Informatics}
                                                    {An der Fakultät für Informatik}
            \\
            \myinstitute\\
            Karlsruhe Institute of Technology (KIT)\\
            Karlsruhe, Germany\\
        }
        \vspace*{1cm}
        \Large{
            \iflanguage{english}{School of Computer Science}
                                                    {An der Fakultät für Informatik}
            \\
            Interactive Systems Lab (ISL)\\
            Carnegie Mellon University (CMU)\\
            Pittsburgh, United States
        }
    \end{center}
    \vspace*{0.5cm}
\Large{
\begin{center}
\begin{tabular}[ht]{l c l}
  \iflanguage{english}{Reviewer}{Erstgutachter}: & \hfill  & \reviewerone\\
  \iflanguage{english}{Second reviewer}{Zweitgutachter}: & \hfill  & \reviewertwo\\
  \iflanguage{english}{Advisor}{Betreuender Mitarbeiter}: & \hfill  & \advisor\\
  \iflanguage{english}{Second advisor}{Zweiter betreuender Mitarbeiter}: & \hfill  & \advisortwo\\
\end{tabular}
\end{center}
}

\vspace{2cm}
\begin{center}
{\large\iflanguage{english}{Duration}{Bearbeitungszeit}: \timestart{} -- \timeend{}}
\end{center}

\begin{textblock}{10}[0,0](4,16.8)
\tiny{
    \iflanguage{english}
        {KIT -- University of the State of Baden-Wuerttemberg and National Research Center of the Helmholtz Association}
        {KIT -- Universit\"at des Landes Baden-W\"urttemberg und nationales Forschungszentrum in der Helmholtz-Gemeinschaft}
}
\end{textblock}

\begin{textblock}{10}[0,0](14,16.75)
\large{
    \textbf{www.kit.edu}
}
\end{textblock}

\end{titlepage}

\cleardoublepage%
\vspace*{36\baselineskip}
\hbox to \textwidth{\hrulefill} 
\par
\iflanguage{english}{I declare that I have developed and written the enclosed
thesis completely by myself, and have not used sources or means without
declaration in the text.}{Ich versichere wahrheitsgemäß, die Arbeit
selbstständig angefertigt, alle benutzten Hilfsmittel vollständig und genau
angegeben und alles kenntlich gemacht zu haben, was aus Arbeiten anderer
unverändert oder mit Abänderungen entnommen wurde.}

\textbf{\submissionplace, \submissiontime}
\vspace{1.5cm}

\dotfill\hspace*{8.0cm}\\
\hspace*{2cm} (\textbf{\myname}) 

\thispagestyle{empty}

\chapter*{Acknowledgement}
Daniel Kirsch published the data collected with Detexify under the
ODbL.\footnote{\url{https://github.com/kirel/detexify-data}} This dataset made
it possible to evaluate many algorithms. Thank you Daniel!

My advisors Kevin Kilgour and Sebastian Stüker told me to make use of GPUs
which boosted neural network training a lot. Thank you!

The StackExchange community helped me with very specific questions I had when I
got problems with my implementation (StackOverflow) or regarding
\LaTeX{} (tex.stackexchange). Especially David Carlisle, Enrico Gregorio and
percusse helped me to understand how \LaTeX{} works, to get some of the
diagrams to compile and helped me with a formulation in the introduction. Thank
you!

Lara Martin and Anna Blomley helped me to notably improve the language in the
first two chapters and the last chapter. It is now much easier to read and
sounds much better. Additionally, I've learned a little bit about punctuation.
Thank you, Lara and Anna!

The Baden-Württemberg Stiftung and interACT gave me the great possibility to
write this bachelor's thesis at Carnegie Mellon University. Thank you!

\begin{figure}[h]
    \centering
    \subfloat{
        \includegraphics*[width=0.48\linewidth, keepaspectratio]{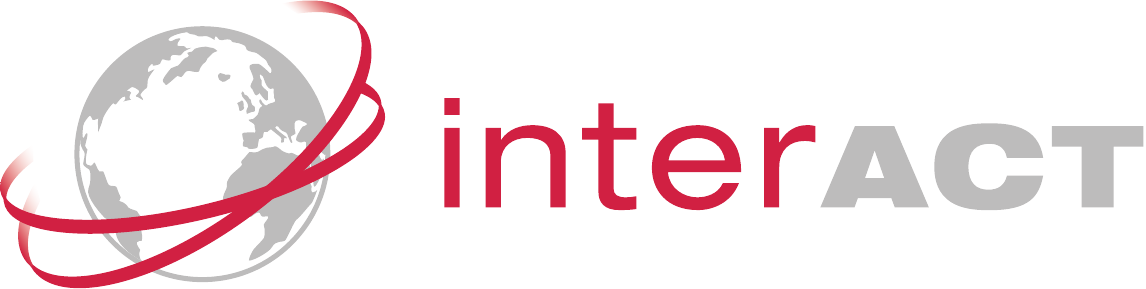}
    }%
    \subfloat{
        \includegraphics*[width=0.48\linewidth, keepaspectratio]{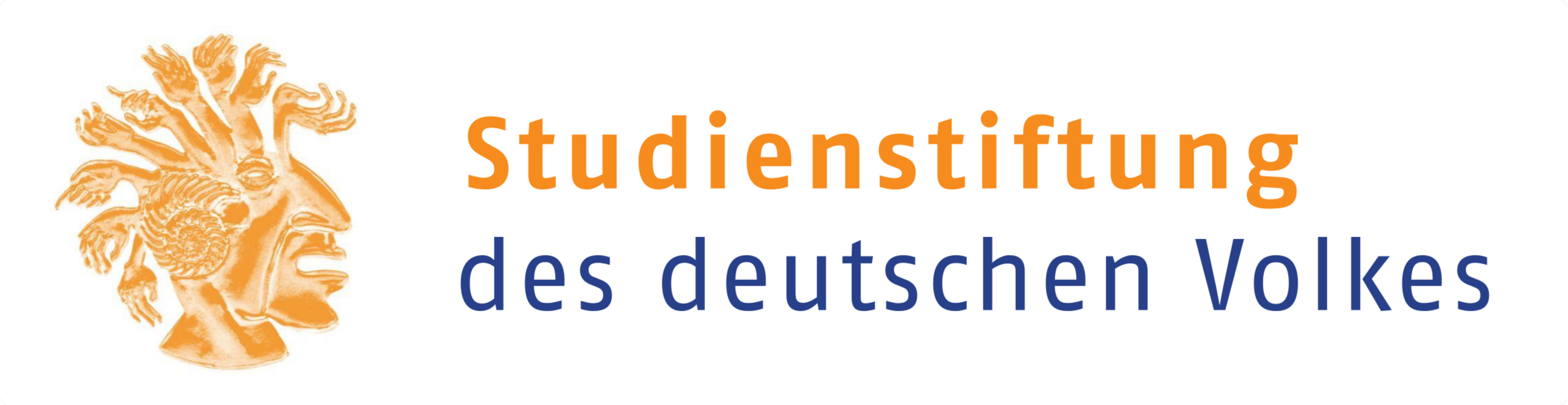}
    }%
\end{figure}

\begin{figure}[h]
    \centering
    \includegraphics*[width=\linewidth, keepaspectratio]{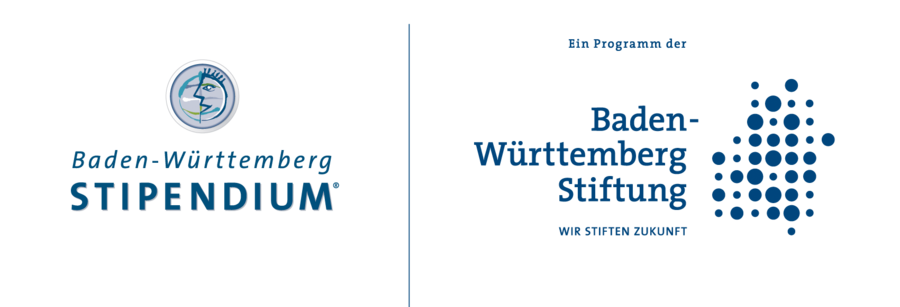}
\end{figure}

\clearpage

This work can be cited the following way:

\begin{verbatim}
@MastersThesis{Thoma:2014,
    Title     = {On-line {Recognition} of {Handwritten} {Mathematical} {Symbols}},
    Author    = {Martin Thoma},
    School    = {Karlsruhe Institute of Technology},
    Year      = {2014},

    Address   = {Karlsruhe, Germany},
    Month     = nov,
    Type      = {Bachelor's Thesis},

    Keywords  = {handwriting recognition; on-line; machine learning;
                 artificial neural networks; mathematics; classification;
                 supervised learning; MLP; multilayer perceptrons; hwrt;
                 write-math},
    Url       = {http://martin-thoma.com/write-math}
}
\end{verbatim}

A DVD with a digital version of this bachelor's thesis and the source code as
well as the used data is part of this work.
\chapter*{Abstract}
Finding the name of an unknown symbol is often hard, but writing the symbol is
easy. This bachelor's thesis presents multiple systems that use the pen
trajectory to classify handwritten symbols. Five preprocessing steps, one data
augmentation algorithm, five features and five variants for multilayer
Perceptron training were evaluated using $\num{\totalCollectedRecordings}$
recordings which were collected with two crowdsourcing projects. The evaluation
results of these 21~experiments were used to create an optimized recognizer
which has a TOP1 error of less than $\SI{17.5}{\percent}$ and a TOP3 error of
$\SI{4.0}{\percent}$. This is an improvement of $\SI{18.5}{\percent}$ for the
TOP1 error and $\SI{29.7}{\percent}$ for the TOP3 error.

\blankpage%

\tableofcontents
\blankpage%

\mainmatter%
\pagenumbering{arabic}
\chapter{Introduction}\label{ch:Introduction}

Euclid's \textit{Elements} is one of the oldest mathematical texts that is still
available. It was written in 300~BC by the ancient Greek mathematician Euclid.
At that time, it was not possible to replicate information fast. Since a person
had to copy the book by hand, its creation was relatively simple regarding the
technology being used, but it was difficult to spread information.

The invention of the printing press changed this, and in~1482, Euclid's Elements
was first set in type. By using a plate, ink, and a press, one could easily make
hundreds of copies. However, the creation of the plate was difficult. It was
made with a combination of movable metal types that could be reused for other
texts and wooden templates for formulas and drawings. In summary, it can be
said that the printing press made it easy to replicate information once the
plate was created, but creating it was hard.

The creation of the original text became easier with the invention and evolution
of computers, and the possibilities for replication became cheaper and more
effective. With computers, one can easily restructure chapters with just a few
keystrokes. Words, and even whole paragraphs, can simply be inserted or deleted
wherever the author wants. Modern, low-priced printers can easily print 20 pages
per minute, and the Internet can be used to spread information on a scale that
was unimaginable before. \TeX{}, a language that allows typesetting of almost
arbitrary content was initially released by Donald Knuth in 1978. It got
extended by \LaTeX{} and is still available for free. It offers to people the
possibility, not only to create texts themselves, but also typeset them to a
high standard without knowledge of typesetting algorithms.\\
Despite all of this progress, there is still a lot of potential to improve the
process of writing.
\LaTeX{} code is written using a keyboard in a combination of Latin script and
special characters like \texttt{\{}, \texttt{\}}, and \texttt{\textbackslash{}}
to form commands such as \verb+\begin{equation}+ or \verb+\alpha+. One tedious
task that all people learning \LaTeX{} have to do to find the code for the
symbol they want to write. This can be done by looking in symbol tables.
However, as touch devices become ubiquitous, systems can be created to let users
write a symbol, record it, and output the \LaTeX{} command of the recognized
symbol. This task of finding a proper textual representation of a given
handwritten \gls{symbol} is called \gls{HWR}. If the recognition software only
uses the pixel image of the recording, it is called off-line
\gls{HWR}. On-line \gls{HWR} can use information from how the
symbols were written, which includes the pen trajectory.

On-line \gls{HWR} can use techniques of off-line \gls{HWR}, but studies have
shown that on-line information notably improves recognition rates
and simplifies algorithms~\cite{Becker72,Guyon91}.

This thesis is about on-line \gls{HWR}. The type of machine learning task is a
classification task, meaning that the set of symbols which should be recognized
is provided.

\section{Symbols, Glyphs and \LaTeX{} Codes}\label{sec:what-is-a-symbol}

A \textit{symbol} is an atomic semantic entity which has exactly one visual
appearance when it is handwritten. Examples of symbols are: $\alpha, \propto,
\cdot, x, \int, \sigma, \dots$\footnote{The first symbol is an \verb+\alpha+,
the second one is a \verb+\propto+.}

While a symbol is a single semantic entity with a given visual appearance, a
glyph is a single typesetting entity. Symbols, glyphs and \LaTeX{} commands do
not relate:

\begin{itemize}
    \item Two different symbols might have the same glyph. For example, the symbols
\verb+\sum+ and \verb+\Sigma+ both render to $\Sigma$, but they have different
semantics and hence they are different symbols. Other symbols that have the same
or similar glyphs can be found in \cref{table:difficult-symbols}.
    \item Two different glyphs might correspond to the same semantic entity. An example is
\verb+\varphi+ ($\varphi$) and \verb+\phi+ ($\phi$): Both represent the small
Greek letter \enquote{phi}, but they exist in two different variants. Hence
\verb+\varphi+ and \verb+\phi+ are two different symbols.
    \item Examples for different \LaTeX{} commands that represent the same symbol are
          \verb+\alpha+ ($\alpha$) and \verb+\upalpha+ ($\upalpha$): Both have the same
semantics and are hand-drawn the same way. This is the case for all \verb+\up+
variants of Greek letters.
\end{itemize}

It is also worth noting that \LaTeX{} commands are neither always glyphs nor
always single symbols. The \LaTeX{} command \verb+\ll+ renders to $\ll$, which
are two symbols. More examples of \LaTeX{} commands that generate two or more
symbols can be found in \cref{table:post-processing-formula-recognizer} on
\cpageref{table:post-processing-formula-recognizer}.

\section{MathML and \LaTeX{}}

The task of symbol recognition is independent of the recognized symbol's output
language as long as the output language is powerful enough.

Both MathML and \LaTeX{} can be used to express a lot of formulas. The difference between
them is how they were meant to be used. \LaTeX{} was developed as an input
language, that is, people should be able to easily write what they want to express.
MathML, on the other hand, is an XML format and hence is easier for programs to parse.

Converters can transform one format into the other. A simple \LaTeX{}-to-MathML
converter can be found at \url{http://www.mathtowebonline.com} and a
MathML-to-\LaTeX{} converter is given by XSLT at
\url{http://code.google.com/p/web-xslt/source/browse/trunk/pmml2tex/}.

\LaTeX{} is used in this bachelor's thesis because it is easier to read. One
can expect readers to understand the \LaTeX{} command \verb+\varphi+ but not the
Unicode code point \verb+\u03C6+. As one aim of this bachelor's thesis is to
provide a symbol recognition system that can be used to find the code for a
hand-drawn symbol, the semantically meaningful output \verb+\varphi+ is of
higher use for the user than \verb+\u03C6+.

Also, \LaTeX{} can be used to express any mathematical formula due to its
powerful extension system. Every common symbol can be expected to be in at least
one package, as \LaTeX{} has been around for over 30 years now and --- as shown by
submissions to \url{arxiv.org} --- is still used a lot.

A notable downside of \LaTeX{} is that parsing it is hard. Even simple tasks ---
like checking if a symbol appears in the rendered output of a given text --- is not
trivial with \LaTeX{}.

\section{Steps in Handwriting Recognition}

One possible way in which handwriting recognizers can work is by performing the
following steps in order to recognize characters, symbols, or words. Not every
recognizer uses all of these steps.

\begin{enumerate}
    \item \textbf{Preprocessing}: Recorded data is never perfect. Devices have
          errors and people make mistakes while using devices. To tackle
          these problems there are preprocessing algorithms to clean the data.
          The preprocessing algorithms can also remove unnecessary variations of
          the data that do not help classify but hide what is important.
          Having slightly different sizes of the same symbol is an example of such a
          variation. Nine preprocessing algorithms that clean or normalize
          recordings are explained in
          \cref{sec:preprocessing}.
    \item \textbf{Data augmentation}: Learning algorithms need lots of data
          to learn internal parameters. If there is not enough data available,
          domain knowledge can be considered to create new artificial
          data from the original data. Ideas for data augmentation in the
          domain of on-line handwriting recognition can be found in
          \cref{sec:Data-augmentation}.
    \item \textbf{Segmentation}: The task of formula recognition can eventually
          be reduced to the task of symbol recognition combined with symbol
          placement. Before symbol recognition can be done, the formula has
          to be segmented. As this bachelor's thesis is only about single-symbol
          recognition, this step was not evaluated.
    \item \textbf{Feature computation}: A feature is high-level information
          derived from the raw data after preprocessing. Some systems like
          \gls{Detexify}, which was presented in~\cite{Kirsch}, simply take the
          result of the preprocessing step, but many compute new features. This
          might have the advantage that less training data is needed since the
          developer can use knowledge about handwriting to compute highly
          discriminative features. Various features are explained in
          \cref{sec:features}.
    \item \textbf{Feature enhancement}: Applying \gls{PCA}, \gls{LDA}, or
          feature standardization might change the features in ways that
          improve the performance of learning algorithms.
          \Cref{sec:feature-enhancement} describes feature standardization.
\end{enumerate}

After these steps, we are faced with a classification learning task which consists of
two parts:
\begin{enumerate}
    \item \textbf{Learning} parameters for a given classifier. This process is
          also called \textit{training}.
    \item \textbf{Classifying} new recordings, sometimes called
          \textit{evaluation}. This should not be confused with the evaluation
          of the classification performance which is done for multiple
          topologies, preprocessing queues, and features in \Cref{ch:Evaluation}.
\end{enumerate}

Two fundamentally different systems for classification of time series data were
evaluated. One uses greedy time warping, which has a very easy, fast learning
algorithm which only stores some of the seen training examples. The other one is
based on neural networks, taking longer to train, but is much faster in
recognition and also leads to better recognition results.

\section{Limitations of Single-Symbol Recognition}

The recognition capabilities of single-symbol classifiers have some limitations
that multi-symbol classifiers do not have. There are symbols such as the
multiplication dot \enquote{$\cdot$} versus the point \enquote{.}, or zero
\enquote{0} versus the capital and the small Latin letter \enquote{O} and \enquote{o}
which can be distinguished by context and the availability of a baseline, but are
extremely hard if not impossible, to distinguish without context. For example, a
preceding \enquote{1} can indicate if the current symbol is a \enquote{0} or an
\enquote{O}. More examples of symbols that look identical without context are
given in \cref{table:difficult-symbols}.

As the design of \href{http://write-math.com}{write-math.com} was set up without
a ruled writing space, it is impossible to distinguish symbols that only differ
in size or their relative position to a baseline. A baseline, and some context
in terms of size and position, could have been established with a user interface
like the one shown in \cref{fig:draft-single-symbol-context-interface}. However,
this was not done for two reasons: On the one hand, most data which was used is
from the Detexify project which has neither this kind of single-symbol context
nor a baseline. On the other hand, users with mobile devices should not be
forced to write at an uncomfortably small size.

\begin{figure} [ht]
    \centering
    \includegraphics[width=0.45\linewidth, keepaspectratio]{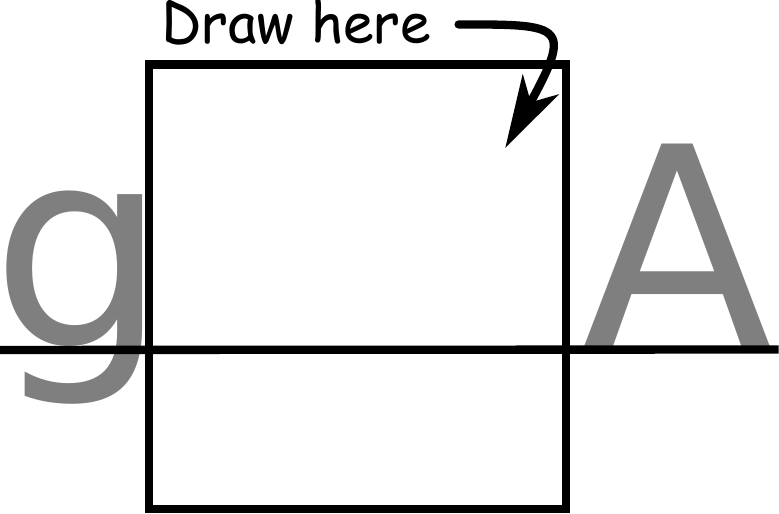}
    \caption{An example of how the user interface for single-symbol recognition
             could be designed. The advantage of this design over a simple empty
             box is that it gives the user some context as to how big he should
             write and where the baseline is. This information could later be
             used by a recognizer to distinguish \enquote{$\cdot$} from
             \enquote{.} or \enquote{o} from \enquote{O}.}%
\label{fig:draft-single-symbol-context-interface}
\end{figure}

\chapter{Related Work}\label{ch:related-work}

On-line handwriting recognition has been a field of study since T.~L.~Dimond
developed a device for reading handwritten characters in 1958~\cite{Dimond1958}.
In the past 56 years, technology changed a lot. Computers went down in size from
hundreds of square meters to less than half a square meter. The energy
consumption and the weight were also notably reduced. At the same time,
computing power grew exponentially. Computers became available for everybody.
Multi-core processors started to spread in the early 2000s, and more data than
ever were produced and stored in the world wide web. \Gls{GPU}-accelerated
computing became usable with the \gls{CUDA} platform, which was initially
released in 2007, boosting the practical capabilities of neural networks.
Combined with the enormous amount of data that is available through the Internet
and services like Amazon Mechanical Turk, it becomes possible to design systems
which learn from large amounts of data.

Meanwhile, there was also progress in the field of on-line handwritten
mathematical formulas:

In 1966, G.~F.~Groner proposed a real-time recognizer that made use of a tablet that
had a time-resolution of $\SI{4}{\milli\second}$ and an accuracy of about
$\SI{0.1}{\milli\metre}$~\cite{Groner66}. The system recognized symbols by
comparing sequences of the directions of strokes with labeled training data and
applying manually-designed tests to features. His system was capable of
recognizing 53 symbols, but only 52 symbols were used in the evaluation. The
evaluation showed that the average recognition rate was at $\SI{87}{\percent}$,
but the users were instructed on how to use the system before the evaluation was
done. This implies that the way users entered the symbols was perhaps not
always the way they would normally write.

In the following years, a lot of work was done in cursive handwriting
recognition. Jaeger, Manke, and Waibel described in~\cite{Manke00,Manke01} a
system that uses a multi-state \gls{TDNN}, which achieved recognition rates of
over $\SI{90}{\percent}$ with context bitmaps for individual lowercase letters (a--z),
individual uppercase letters (A--Z), or digits 
(0--9). Context bitmaps show a $3 \times 3$ bitmap of the proximity of a point.

One of the early works done in on-line handwriting recognition for mathematical
formulas is~\cite{Belaid1984}. Their system used a combination of sequence
vectors and a feature that was the ratio of the distance from the starting point
to the end point and the symbol height. Only 35 different symbols were
evaluated. With those settings, a recognition rate of $\SI{93}{\percent}$ was
achieved.

In 1998, A.~Kosmala and G.~Rigoll designed a system for on-line mathematical
handwriting recognition which was trained to recognize 100 different symbols
\cite{Kosmala98}. This included the $52$ lowercase and uppercase letters, $23$
mathematical symbols, $11$ lower-case Greek letters, and $6$ parentheses. The
system was designed to recognize complete formulas, although the symbols of the
formula had to be drawn in a predefined order. It applied \glspl{HMM} for symbol
segmentation. The data was resampled, but no other preprocessing was described.
A sampled bitmap was used as a feature, as well as on-line features like the
writing direction. One-hundred common mathematical and physical formulas were
used as a training set, and $30$ additional formulas as a test set. They claimed
to get recognition rates of $\SI{96.3}{\percent}$. However, this seems to be
very high since in the \gls{CROHME} from 2013, the best team achieved
recognition rates of $\SI{60.36}{\percent}$ and the second best team achieved
recognition rates of $\SI{23.40}{\percent}$ as documented
in~\cite{Mouchere2013}. The first two competitions, \cite{Mouchere2012}
and~\cite{Mouchere2011}, also did not receive any submissions that had
recognition rates over $\SI{65}{\percent}$.

Daniel Kirsch used over $1000$ symbols in his diploma thesis~\cite{Kirsch}. He
evaluated a very simple recognition system called \textit{Detexify} which was
--- and still is, at the time of this work --- accessible through the
web.\footnote{\url{http://detexify.kirelabs.org/}} Many people can use the
recognition system as nothing else than a browser and internet access is
required, while providing a huge number of symbols that can get classified.
However, in his evaluation, Kirsch used only a randomly-chosen subset of $100$
symbols. He claimed to get a TOP1 error of $\SI{26.12}{\percent}$ and a TOP3
error of less than $\SI{10}{\percent}$ with standard
\gls{DTW} and other variants of \gls{DTW}.

The aim of this work is to build a recognition system that is as accessible as
Detexify, but is faster, can recognize more symbols, and has a higher
recognition rate. The presented system is able to classify
\totalClassesAnalyzed{} symbols. Furthermore, this work contributes to the first
publicly-available dataset for  on-line handwriting recognition, with more than
$\num{\totalCollectedRecordings}$ recordings. This will help to make experiments
and different classifiers comparable. The symbol classifier~A (see
\cref{sec:system-a-greedy-matching}) can be tested on \url{write-math.com}, and
the comparably better classifier~B (see \cref{sec:complex-recognizer}) is currently not
publicly available but is planned to be released in the near future.


\chapter{Domain Specific Classification Steps}

Taking a close look at the collected data might give relevant insights into
problems one has to deal with and eventually imagine preprocessing steps that
can reduce those problems. It could also lead to ideas for features that are
invariant to variations that occur in the dataset.

This chapter explains classification steps that are specific for on-line
\gls{HWR}, whereas the next chapter explains the rather domain independent task
of classification of time series data.


\section{Data}
The data that was used for all experiments was collected with
\href{http://write-math.com}{write-math.com} and
\href{http://detexify.kirelabs.org/classify.html}{detexify.kirelabs.org}
(see~\cite{Kirsch}). write-math.com is a website designed by me for this 
bachelor's thesis whereas Detexify was created by Daniel Kirsch. Both websites
use HTML and JavaScript to gather data and both websites store the same data,
but in a slightly different data format.

\href{http://write-math.com}{write-math.com} makes use of HTML5 canvas elements. Those elements can be used in
combination with JavaScript to track fingers or a mouse cursor touching the
canvas, moving and lifting. Every point is specified by two integer coordinates
$(x, y)$. The origin $(0, 0)$ is at the upper left corner of the rectangular
canvas element and $x$ values get bigger to the right and $y$ values get bigger
to the bottom. \Cref{fig:canvas-plane} shows such an HTML5 canvas plane.
JavaScript asynchronously triggers events that contain the information where on
the canvas the cursor or finger currently is. Those points are called
\textit{control points} in the following.
For the mouse, the information if the mouse button is currently pressed down is
also available. So when the position is recorded, the stroke-wise segmentation
is automatically given for both, mouse and fingers. A list of such user
generated control points together with the information which points belong to
the same stroke and the information when the point was recorded is called a
\textit{recording}. An example of a recording is
\cref{fig:handwritten-symbol}.

\begin{figure}[ht]
    \centering
    \subfloat[HTML5 canvas plane]{
        \adjustbox{max width=0.5\textwidth,max height=5cm}{
            \newcommand*{\xMin}{0}%
\newcommand*{\xMax}{6}%
\newcommand*{\yMin}{0}%
\newcommand*{\yMax}{6}%
\begin{tikzpicture}[y=-1cm]
    \foreach \i in {\xMin,...,\xMax} {
        \draw [very thin,gray!25] (\i,\yMin) -- (\i,\yMax)  node [below] at (\i,\yMin) {};
    }
    \foreach \i in {\yMin,...,\yMax} {
        \draw [very thin,gray!25] (\xMin,\i) -- (\xMax,\i) node [left] at (\xMin,\i) {};
    }
    \draw[->, very thick] (0,0) -- (1,0) node[above] {$x$};
    \draw[->, very thick] (0,0) -- (0,1) node[left] {$y$};
    \foreach \x in {\xMin,...,5} {
        \foreach \y in {\yMin,...,5} {
            \node at ({\x+0.5},{\y+0.5}) {$(\x, \y)$};
        }
    }
\end{tikzpicture}
        }
\label{fig:canvas-plane}
    }%
    \subfloat[Recorded sequence of points]{
        \adjustbox{max width=0.5\textwidth,max height=5cm}{
            \begin{tikzpicture}
    \tikzstyle{point}=[circle,thick,draw=black,fill=black,inner sep=0pt,minimum width=4pt,minimum height=4pt]
    \tikzstyle{smallpoint}=[circle,thick,draw=black,fill=black,inner sep=0pt,minimum width=0.5pt,minimum height=0.5pt]
    \draw [gray!50] (0.91, 6.90)-- (0.91, 6.75)-- (0.83, 6.15)-- (0.83, 5.85)-- (0.76, 5.62)-- (0.68, 4.65)-- (0.53, 3.98)-- (0.53, 3.45)-- (0.46, 2.85)-- (0.46, 2.48)-- (0.23, 1.35)-- (0.23, 0.75)-- (0.01, 0.38)-- (0.01, 0.15)-- (0.01, 0.07);
    \draw [gray!50] (0.08, 0.00)-- (0.16, 0.00)-- (0.31, 0.00)-- (0.46, 0.00)-- (0.68, 0.00)-- (0.83, 0.00)-- (1.13, 0.00)-- (1.43, 0.07)-- (1.66, 0.15)-- (1.96, 0.15)-- (2.26, 0.15)-- (3.68, 0.38)-- (4.51, 0.53)-- (4.73, 0.53)-- (4.88, 0.53)-- (5.03, 0.60);
    \draw [gray!50] (0.91, 3.90)-- (1.13, 3.90)-- (1.36, 3.98)-- (1.88, 3.98)-- (3.01, 4.12)-- (4.81, 4.20)-- (5.56, 4.42)-- (5.63, 4.42)-- (5.78, 4.42)-- (5.86, 4.50);
    \draw [gray!50] (0.23, 7.58)-- (0.31, 7.58)-- (0.38, 7.58)-- (0.46, 7.58)-- (0.61, 7.58)-- (0.68, 7.58)-- (0.76, 7.58)-- (0.98, 7.58)-- (1.21, 7.58)-- (1.43, 7.50)-- (1.51, 7.50)-- (1.88, 7.50)-- (2.18, 7.50)-- (2.33, 7.50)-- (2.93, 7.50)-- (3.83, 7.50)-- (4.43, 7.50)-- (4.96, 7.50)-- (5.11, 7.50)-- (5.26, 7.50)-- (5.33, 7.50)-- (5.41, 7.50)-- (5.56, 7.58)-- (5.63, 7.58)-- (5.71, 7.58)-- (5.78, 7.65)-- (5.93, 7.73)-- (5.93, 7.80);

    \node (P)[point,label={[label distance=0cm]180:0}] at (0.91, 6.90) {};
    \node (P)[smallpoint] at (0.91, 6.75) {};
    \node (P)[smallpoint] at (0.83, 6.15) {};
    \node (P)[smallpoint] at (0.83, 5.85) {};
    \node (P)[smallpoint] at (0.76, 5.62) {};
    \node (P)[smallpoint] at (0.68, 4.65) {};
    \node (P)[smallpoint] at (0.53, 3.98) {};
    \node (P)[smallpoint] at (0.53, 3.45) {};
    \node (P)[smallpoint] at (0.46, 2.85) {};
    \node (P)[smallpoint] at (0.46, 2.48) {};
    \node (P)[point,label={[label distance=0cm]180:10}] at (0.23, 1.35) {};
    \node (P)[smallpoint] at (0.23, 0.75) {};
    \node (P)[smallpoint] at (0.01, 0.38) {};
    \node (P)[smallpoint] at (0.01, 0.15) {};
    \node (P)[point,label={[label distance=0cm]180:14}] at (0.01, 0.07) {};
    \node (P)[point,label={[label distance=0cm]-90:15}] at (0.08, 0.00) {};
    \node (P)[smallpoint] at (0.16, 0.00) {};
    \node (P)[smallpoint] at (0.31, 0.00) {};
    \node (P)[smallpoint] at (0.46, 0.00) {};
    \node (P)[smallpoint] at (0.68, 0.00) {};
    \node (P)[smallpoint] at (0.83, 0.00) {};
    \node (P)[smallpoint] at (1.13, 0.00) {};
    \node (P)[smallpoint] at (1.43, 0.07) {};
    \node (P)[smallpoint] at (1.66, 0.15) {};
    \node (P)[smallpoint] at (1.96, 0.15) {};
    \node (P)[point,label={[label distance=0cm]-90:25}] at (2.26, 0.15) {};
    \node (P)[smallpoint] at (3.68, 0.38) {};
    \node (P)[smallpoint] at (4.51, 0.53) {};
    \node (P)[smallpoint] at (4.73, 0.53) {};
    \node (P)[smallpoint] at (4.88, 0.53) {};
    \node (P)[point,label={[label distance=0cm]-90:30}] at (5.03, 0.60) {};
    \node (P)[point,label={[label distance=0cm]-90:31}] at (0.91, 3.90) {};
    \node (P)[smallpoint] at (1.13, 3.90) {};
    \node (P)[smallpoint] at (1.36, 3.98) {};
    \node (P)[smallpoint] at (1.88, 3.98) {};
    \node (P)[smallpoint] at (3.01, 4.12) {};
    \node (P)[smallpoint] at (4.81, 4.20) {};
    \node (P)[smallpoint] at (5.56, 4.42) {};
    \node (P)[smallpoint] at (5.63, 4.42) {};
    \node (P)[smallpoint] at (5.78, 4.42) {};
    \node (P)[point,label={[label distance=0cm]-90:40}] at (5.86, 4.50) {};
    \node (P)[point,label={[label distance=0cm]-90:41}] at (0.23, 7.58) {};
    \node (P)[smallpoint] at (0.31, 7.58) {};
    \node (P)[smallpoint] at (0.38, 7.58) {};
    \node (P)[smallpoint] at (0.46, 7.58) {};
    \node (P)[smallpoint] at (0.61, 7.58) {};
    \node (P)[smallpoint] at (0.68, 7.58) {};
    \node (P)[smallpoint] at (0.76, 7.58) {};
    \node (P)[smallpoint] at (0.98, 7.58) {};
    \node (P)[smallpoint] at (1.21, 7.58) {};
    \node (P)[smallpoint] at (1.43, 7.50) {};
    \node (P)[point,label={[label distance=0cm]-90:51}] at (1.51, 7.50) {};
    \node (P)[smallpoint] at (1.88, 7.50) {};
    \node (P)[smallpoint] at (2.18, 7.50) {};
    \node (P)[smallpoint] at (2.33, 7.50) {};
    \node (P)[smallpoint] at (2.93, 7.50) {};
    \node (P)[smallpoint] at (3.83, 7.50) {};
    \node (P)[smallpoint] at (4.43, 7.50) {};
    \node (P)[smallpoint] at (4.96, 7.50) {};
    \node (P)[smallpoint] at (5.11, 7.50) {};
    \node (P)[smallpoint] at (5.26, 7.50) {};
    \node (P)[point,label={[label distance=0cm]-90:61}] at (5.33, 7.50) {};
    \node (P)[smallpoint] at (5.41, 7.50) {};
    \node (P)[smallpoint] at (5.56, 7.58) {};
    \node (P)[smallpoint] at (5.63, 7.58) {};
    \node (P)[smallpoint] at (5.71, 7.58) {};
    \node (P)[smallpoint] at (5.78, 7.65) {};
    \node (P)[smallpoint] at (5.93, 7.73) {};
    \node (P)[point,label={[label distance=0cm]-90:68}] at (5.93, 7.80) {};
\end{tikzpicture}
        }
\label{fig:handwritten-symbol}
    }%
\label{fig:data}
    \caption{On the left side is an HTML5 canvas plane. Each coordinate
             $(x,y) \in \mathbb{N}_{0}^{2}$ is one pixel. Every coordinate has
             to be non-negative and an integer. On the right side is a
             visualization of a recording after preprocessing steps that reduced
             the number of points. The small and the large points are the
             control points. The large points are points with an annotation
             which indicates the order in which the points were recorded. Point
             $0$ is the first point that was recorded, point $68$ is the last
             one. Points of one stroke were connected with straight lines.
             \Cref{fig:handwritten-symbol} has 4 strokes in total.}
\end{figure}

\detexifyPercentage{} of the $\num{\totalCollectedRecordings{}}$ recordings that
were used in the evaluation were collected by Detexify.
The recordings are stored in \gls{JSON} format as a list of \glspl{stroke}. Each
stroke consists of tuples $(x(t), y(t), t)$ where $x$ and $y$ are canvas
coordinates and $t$ is a timestamp given in milliseconds since 1970. An example
of a recording in \gls{JSON} format is in \cref{appendix:raw-data-example}.

The time resolution between points as well as the resolution of the recording
depends on the device that was used. However, most recordings have a time
resolution of about $\SI{20}{\milli\second}$ and are within a bounding box of a
$\SI{250}{\pixel} \times \SI{250}{\pixel}$ square.
\Cref{fig:average-time-between-control-points} shows how the time between
control points is spread amongst the analyzed data. It shows that one can expect
a time resolution of $\SI{50}{\milli\second}$ and should eventually treat
control points of one stroke that take longer as errors.

$\num{\totalCollectedRecordings}$ recordings were collected for the
$\num{\totalClassesAnalyzed}$ classes which were tested.\footnote{Links to those
recordings and more are available at
\href{http://martin-thoma.com/write-math}{martin-thoma.com/write-math}.}

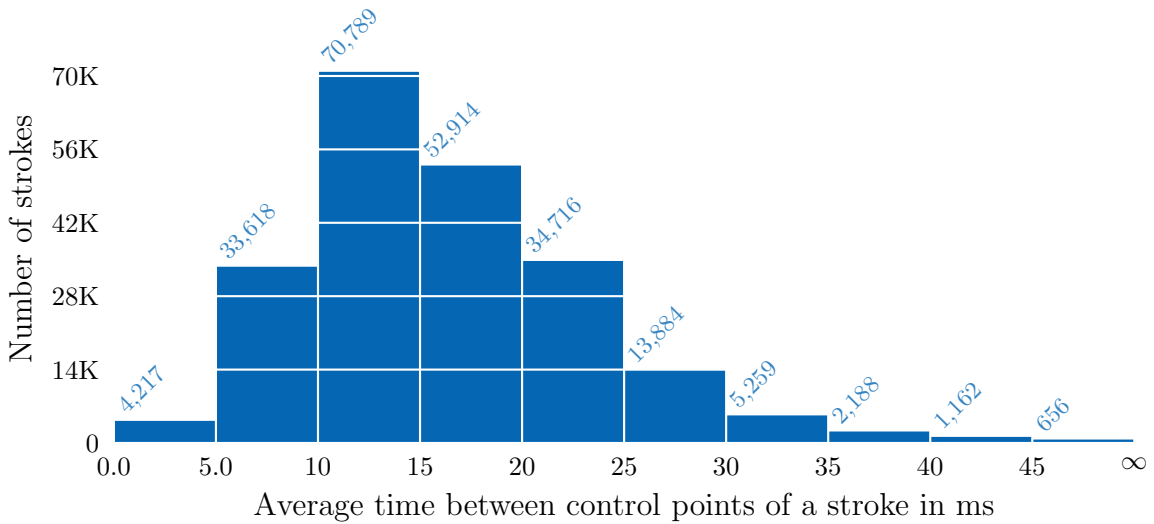
\begin{figure}[h!]
    \centering
    \newcommand\clipright[1][white]{
  \fill[#1](current axis.south east)rectangle(current axis.north-|current axis.outer east);
  \pgfresetboundingbox%
  \useasboundingbox(current axis.outer south west)rectangle([xshift=.5ex]current axis.outer north-|current axis.east);
}

\definecolor{mycolor}{rgb}{0.02,0.4,0.7}

\begin{tikzpicture}
    \begin{axis}[
        ymajorgrids,
        xmajorgrids,
        grid style={white,thick},
        axis on top,
        /tikz/ybar interval,
        tick align=outside,
        ymin=0,
        axis line style={draw opacity=0},
        tick style={draw=none},
        enlarge x limits=false,
        height=7cm,
        title style={font=\Large},
        xlabel={Average time between control points of a stroke in $\si{\milli\second}$},
        ylabel={Number of strokes},
        ytick={ 0,14000,28000,42000,56000,70000 },
        scaled ticks=false,
        yticklabels={ 0, 14K, 28K, 42K, 56K, 70K },
        xticklabels={ $0.0$, $5.0$, $10$, $15$, $20$, $25$, $30$, $35$, $40$, $45$, $\infty$ },
        width=\textwidth,
        xtick=data,
        label style={font=\large},
        ticklabel style={
            inner sep=1pt,
            font=\small
        },
        nodes near coords,
        every node near coord/.append style={
            fill=white,
            anchor=mid west,
            shift={(3pt,4pt)},
            inner sep=0,
            font=\footnotesize,
            rotate=45},
            ]
    \addplot[mycolor!80!white, fill=mycolor, draw=none] coordinates { (0, 4217) (1, 33618) (2, 70789) (3, 52914) (4, 34716) (5, 13884) (6, 5259) (7, 2188) (8, 1162) (9, 656) (10, 1480)  };
    \end{axis}
    \clipright
\end{tikzpicture}
    \caption{More than $\SI{98}{\percent}$ of all time values between two
             control points of the same stroke are less than
             $\SI{35}{\milli\second}$. More than $\SI{73}{\percent}$ are
             captured faster than in $\SI{20}{\milli\second}$.}
\label{fig:average-time-between-control-points}
\end{figure}

\subsection{Choice of Symbols}

The choice of symbols which the classifier was trained to recognize was
directly influenced by the number of obtained recordings per symbol. None of
the \totalClassesNotAnalyzedBelowFifty{}~symbols with less than 50~recordings
were evaluated, although some of them are used in mathematical formulas.

The following symbols or groups of symbols were then removed from the remaining
set of \totalClassesAboveFifty{}~symbols:

\begin{itemize}%
    \item Symbols that don't fit in the context of this work:
        \begin{itemize}
            \item Text mode-only symbols: \verb+\MVAt+ (\MVAt), $@$,
                  \verb+\textsurd+ (\textsurd), \dots
            \item Image-like symbols: \verb+\Bat+ (\Bat), \verb+\Mundus+ (\Mundus)
        \end{itemize}
    \item \LaTeX{} commands that are not symbols as defined before:
    \begin{itemize}
        \item \enquote{\textbackslash{}big\ldots} variants:
              \verb+\bigoplus+ ($\bigoplus$),
              \verb+\bigstar+ ($\bigstar$), \verb+\bigcup+ ($\bigcup$) \dots
        \item \enquote{\textbackslash{}Up} and \enquote{\textbackslash{}up}
              variants of Greek letters: \verb+\Upsigma+ ($\Upsigma$),
              \verb+\uppi+ ($\uppi$), \verb+\uplambda+ ($\uplambda$), \dots
        \item \enquote{\textbackslash{}thick} variants:
              \verb+\thicksim+ ($\thicksim$),
              \verb+\thickapprox+ ($\thickapprox$)
        \item \verb+\dotsb+ ($\dotsb$), but \verb+\dots+ was evaluated
        \item \verb+\cdotp+ ($\cdotp$) because it is the same as \verb+\cdot+
              ($\cdot$), except that it is used for punctuation whereas
              \verb+\cdotp+ is used for the binary math operator.
        \item \verb+\ocircle+ ($\ocircle$) because it is
              the same as the included symbol \verb+\circledcirc+
              ($\circledcirc$).
        \item Multiple-symbol \LaTeX{} commands like \verb+\ll+ ($\ll$) as
              shown in \cref{table:post-processing-formula-recognizer}. In a
              multiple-symbol classifier, these symbol sequences could be
              detected and replaced in a post-classification step.
    \end{itemize}
\end{itemize}

\begin{table}[ht]
    \centering
    \begin{tabular}{ll|ll} 
        \toprule
        \multicolumn{2}{c|}{Search}           & \multicolumn{2}{c}{Replace} \\
        \LaTeX{}            & Rendered       & \LaTeX{}      & Rendered \\\midrule
        \verb+\int\int+     & $\int\int$     & \verb+\iint+  & $\iint$ \\
        \verb+\int\int\int+ & $\int\int\int$ & \verb+\iiint+ & $\iiint$ \\
        \verb+\int\int\int\int+ & $\int\int\int\int$ & \verb+\iiiint+ & $\iiiint$ \\
        \verb+<<+           & $<<$           & \verb+\ll+    & $\ll$ \\
        \verb+<<<+          & $<<<$          & \verb+\lll+   & $\lll$ \\
        \verb+>>+           & $>>$           & \verb+\gg+    & $\gg$ \\
        \verb+>>>+          & $>>>$          & \verb+\ggg+   & $\ggg$ \\
        \verb+\int\cdots\int+ & $\int\cdots\int$ & \verb+\dotsint+   & $\dotsint$ \\
        \bottomrule
    \end{tabular}
    \caption{The single-symbol \LaTeX{} commands shown above have a pendant
             which renders to multiple symbols. A multiple-symbol classifier
             could search for those recognized patterns and replace them by a
             single \LaTeX{} command in order to get a better typesetted version of
             the text. That reduces the number of classes such a
             multiple-symbol classifier has to be able to recognize.}
\label{table:post-processing-formula-recognizer}
\end{table}

All symbols that were used to evaluate the algorithms are listed in
\crefrange{table:symbols-used-for-evaluation-0}{table:symbols-used-for-evaluation-8}.
This includes:

\begin{table}[H]
    \begin{tabular}{rlrll}
    $a-z$ & Small letters   & $\alpha-\omega$ & Small Greek letter    & $\rightarrow$, $\leftarrow$, $\Rightarrow$, $\Leftarrow$, $\Leftrightarrow$, $\dots$ \\
    $A-Z$ & Capital letters & $A-\Omega$      & Capital Greek letters & $=$, $\sim$, $\equiv$, $\approx$, $\dots$ \\
    $0-9$ & Digits          & \multicolumn{2}{c}{$+$, $-$, $\cdot$, $\sqrt{}$, $\cup$, $\cap$, $\dots$} & $\oplus$, $\star$, $\dots$ \\
    \end{tabular}
\end{table}

\subsection{Problems}
As the data was collected via crowdsourcing it has errors. Human classification
errors are only a problem for model training; a model trained with these might
make the same error as humans made before. Four different types of human
classification errors can be distinguished:

\begin{problemenumh}
    \item\label{itm:h1-confusion} \textit{Confusion}: Recordings were classified
              wrong, but the correct class looks similar to the chosen class,
              e.g. $\epsilon$, $\varepsilon$ and $\in$.
    \item\label{itm:h2-creativity} \textit{Creativity}: Drawings that should not have been entered
              in the first place were arbitrarily classified by the user.
              Some examples are shown in \cref{fig:creative-users}.
    \item\label{itm:h3-cherry-picking} \textit{Cherry-Picking}: Drawings of complete formulas were
              entered and classified as a class of a single symbol of that
              formula.
    \item\label{itm:h4-manipulation} \textit{Manipulation}: Obviously wrong classified symbols, e.g.
              $\epsilon$ that gets classified as $\alpha$.
\end{problemenumh}

Additionally to those human classification errors, there are errors that
are caused by the device or the human who uses it while drawing. Those errors
should be considered in preprocessing:

\begin{problemenumd}
    \item\label{itm:d1-wild-points} \textit{Wild points}: Points that appear
          randomly anywhere on the drawing plane.
    \item\label{itm:d2-missing-stroke} \textit{Missing strokes}: The user drew
          a stroke, but only the first point or the last point was captured.
          This might happen more often when the user tries to draw small strokes
          with his fingers. Examples are shown in \cref{fig:missing-strokes}.

          This problem could be confused with \cref{itm:d1-wild-points}.
            \begin{figure}[bt]
                \centering
                \subfloat[ID 288612 ($\in$)]{
                    \hspace*{2em}
                    \includegraphics[height=0.1\linewidth, keepaspectratio]{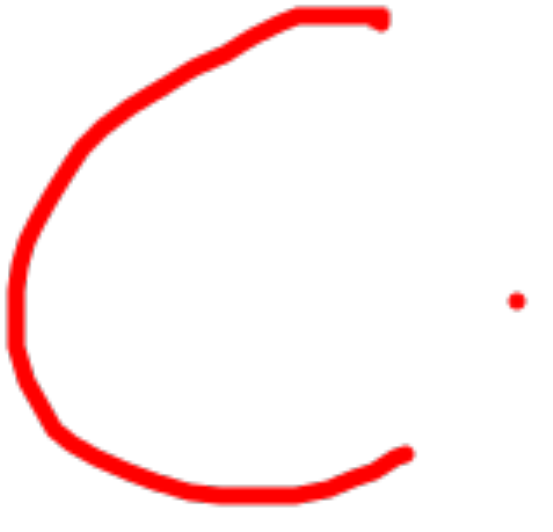}
                    \hspace*{2em}
\label{fig:missing-stroke-288612}
                }%
                \subfloat[ID 291939 ($\forall$)]{
                    \hspace*{2em}
                    \includegraphics[height=0.1\linewidth, keepaspectratio]{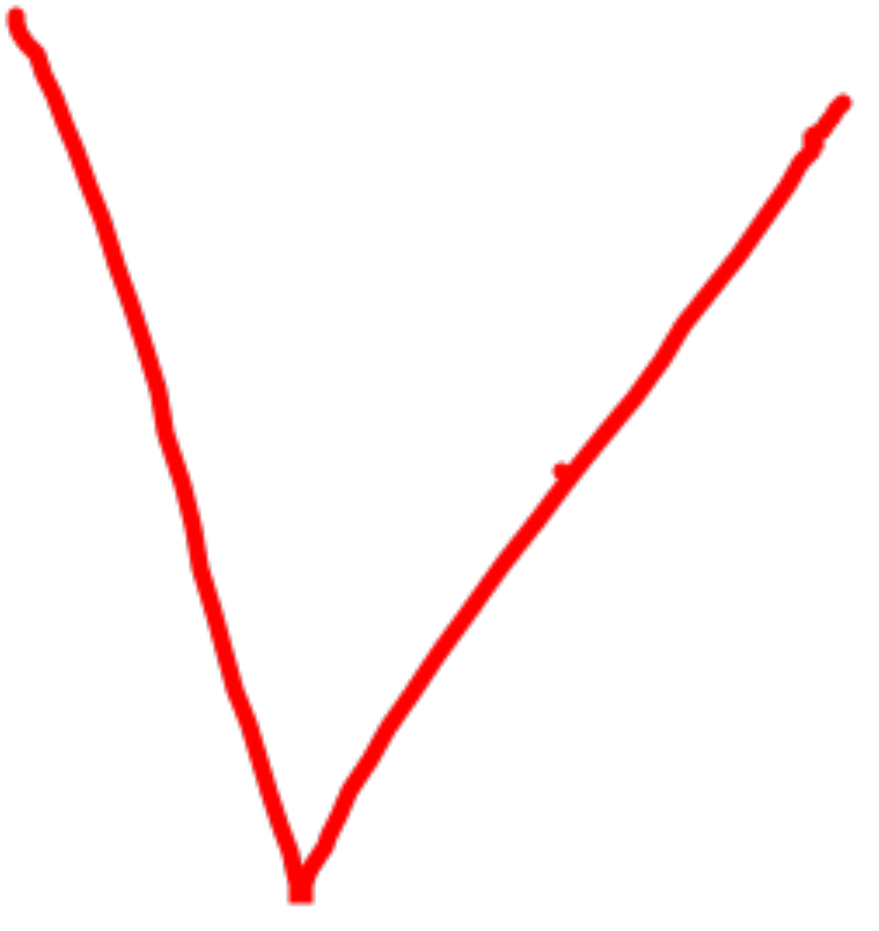}
                    \hspace*{2em}
\label{fig:missing-stroke-291939}
                }%
                \subfloat[ID 282212 ($\models$)]{
                    \hspace*{2em}
                    \includegraphics[height=0.1\linewidth, keepaspectratio]{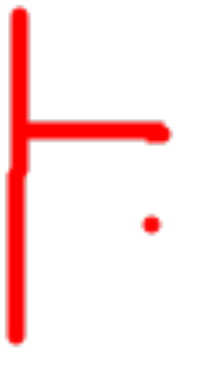}
                    \hspace*{2em}
\label{fig:missing-stroke-282212}
                }%
                \subfloat[ID 262502 ($\Pi$)]{
                    \hspace*{2em}
                    \includegraphics[height=0.1\linewidth, keepaspectratio]{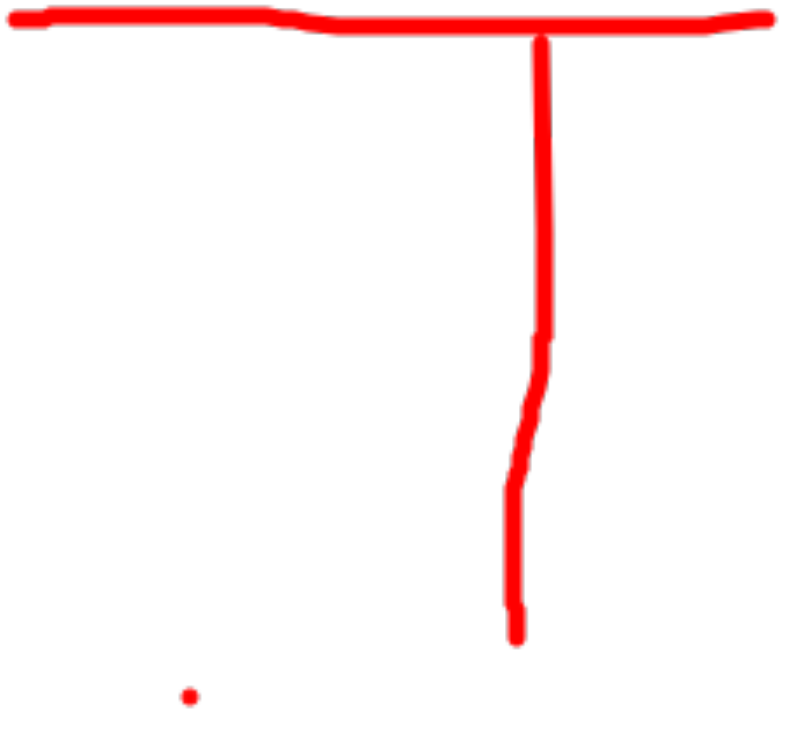}
                    \hspace*{2em}
\label{fig:missing-stroke-262502}
                }%
                \caption{Examples for missing strokes
                         (\cref{itm:d2-missing-stroke}). The classification was
                         added by the user who created the recording. It is not
                         possible to tell if the captured single point was the
                         last or the first point of a stroke.}
\label{fig:missing-strokes}
            \end{figure}
    \item\label{itm:d3-too-long-stroke} \textit{Too long strokes}: The user
          made a stroke much longer than he wanted to. Examples are shown in
          \cref{fig:too-long-lines}.
            \begin{figure}[bt]
                \centering
                \subfloat[ID 258177]{
                    \includegraphics[height=0.1\linewidth, keepaspectratio]{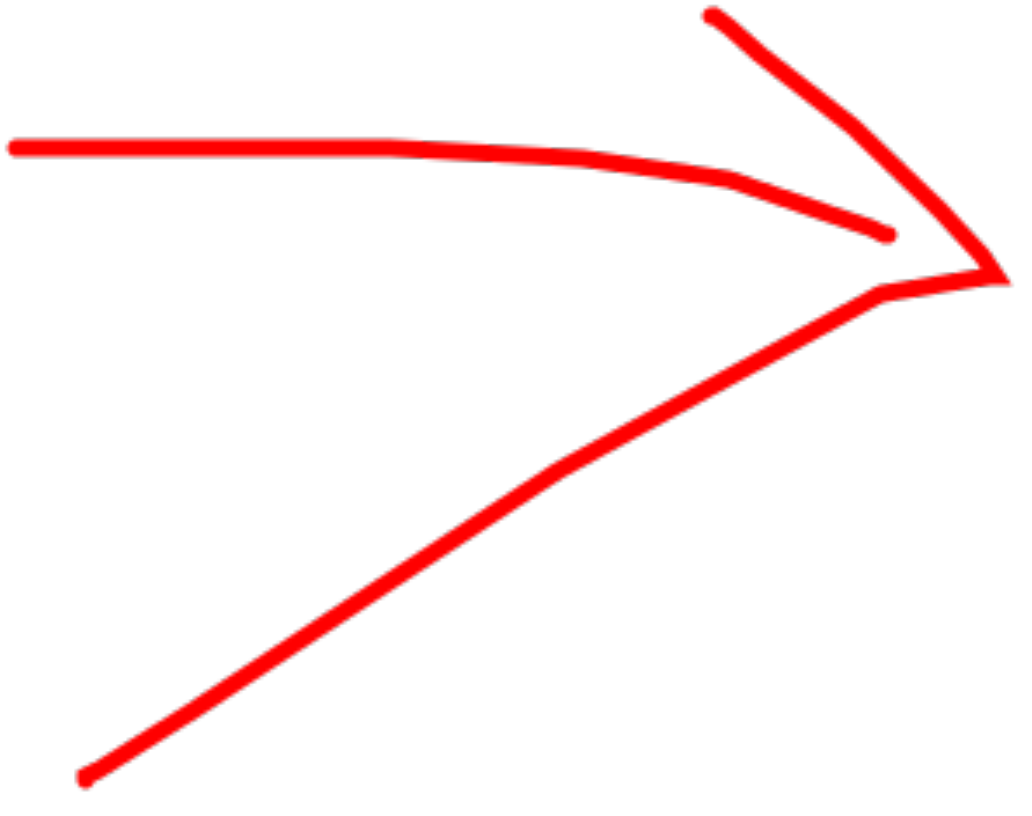}
\label{fig:too-long-line-258177}
                }%
                \subfloat[ID 270115]{
                    \includegraphics[height=0.1\linewidth, keepaspectratio]{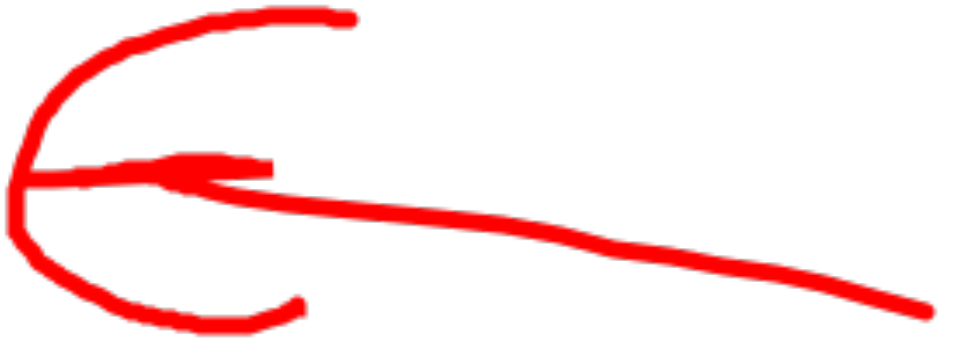}
\label{fig:too-long-line-270115}
                }%
                \subfloat[ID 286813]{
                    \includegraphics[height=0.15\linewidth, keepaspectratio]{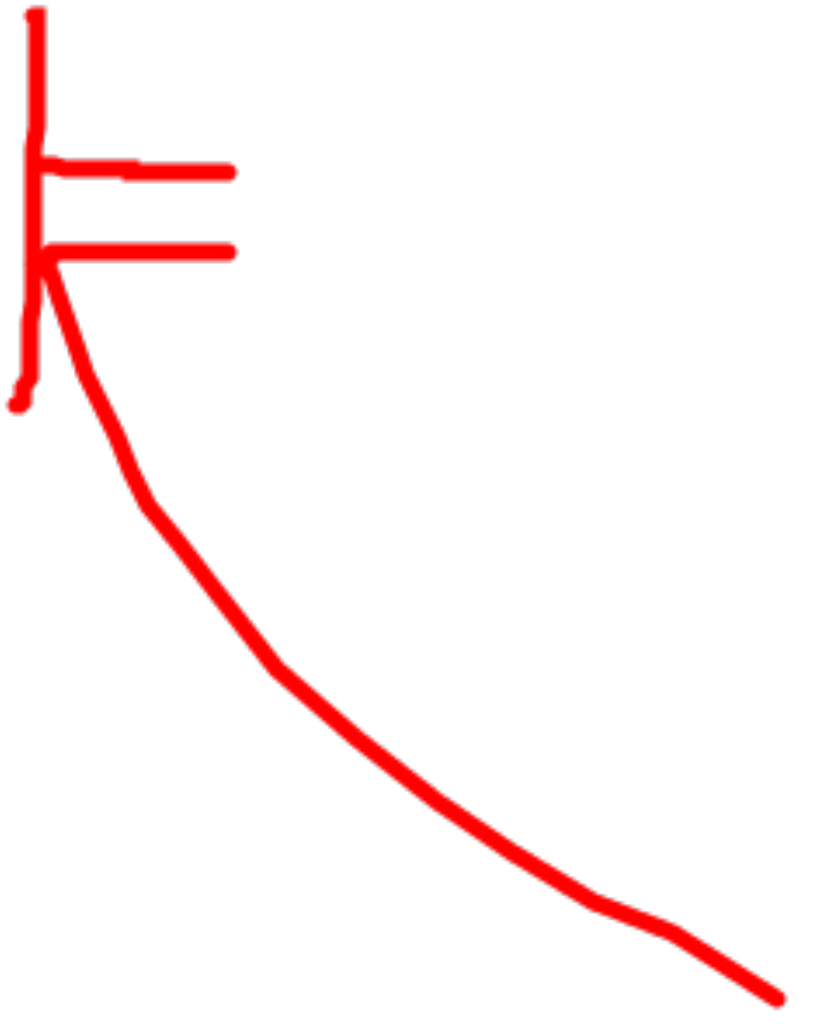}
\label{fig:too-long-line-286813}
                }%
                \subfloat[ID 249024]{
                    \includegraphics[height=0.1\linewidth, keepaspectratio]{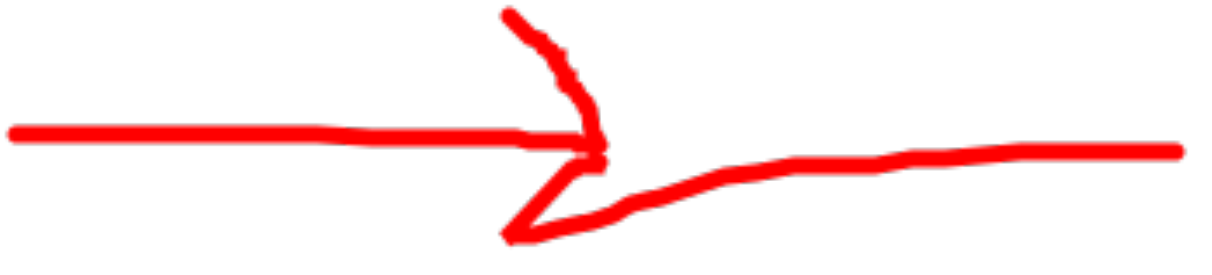}
\label{fig:too-long-line-249024}
                }%
                \caption{Examples for too long strokes
                         (\cref{itm:d3-too-long-stroke}) that users probably
                         did not want to make that long.}
\label{fig:too-long-lines}
            \end{figure}
    \item\label{itm:d4-hook} \textit{Hooks}: At the beginning or end of a stroke
          the user makes a hook, which he did not want to make. Examples are
          shown in \cref{fig:hooks}.
            \begin{figure}[tb]
                \centering
                \subfloat[ID 8350]{
                    \includegraphics[height=0.2\linewidth, keepaspectratio]{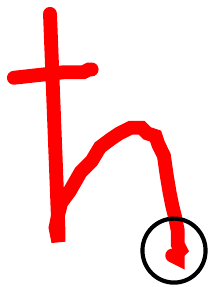}
\label{fig:hook-hbar}
                }%
                \subfloat[ID 11387]{
                    \includegraphics[height=0.2\linewidth, keepaspectratio]{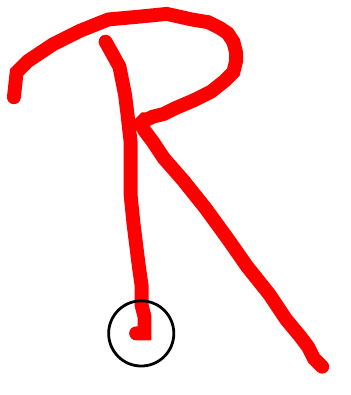}
\label{fig:hook-R}
                }%
                \caption{Examples for hooks at the end or the beginning of a
                         stroke that should not be there (\cref{itm:d4-hook}).}
\label{fig:hooks}
            \end{figure}
    \item\label{itm:d5-interrupted-stroke} \textit{Interrupted strokes}:
          Although the user drew one stroke, the stroke is interrupted and thus
          recorded as multiple strokes. See \cref{fig:interrupted-stroke} on
          \cpageref{fig:interrupted-stroke} for an example.
    \item\label{itm:d7-multiple-stroke-drawing} \textit{Multiple drawn strokes}:
          Some people draw strokes twice. This introduces new variants how
          symbols can be drawn. See \cref{fig:multiple-stroke-drawing} on
          \cpageref{fig:multiple-stroke-drawing} for an example.
    \item\label{itm:d6-wrong-time} \textit{Wrong timestamps}: Some of the data seems
          to have the wrong time. It seems highly unlikely that users took over
          10 minutes to draw a single symbol, yet alone over a day. A plot for
          which the mean recording time and the standard deviation of every
          symbol is shown in \cref{fig:time-mean-std-deviation} and the four most
          extreme values in \cref{table:time-mean-std-deviation}.
\end{problemenumd}

\begin{figure}[tb]
    \centering
    \includegraphics*[width=\textwidth]{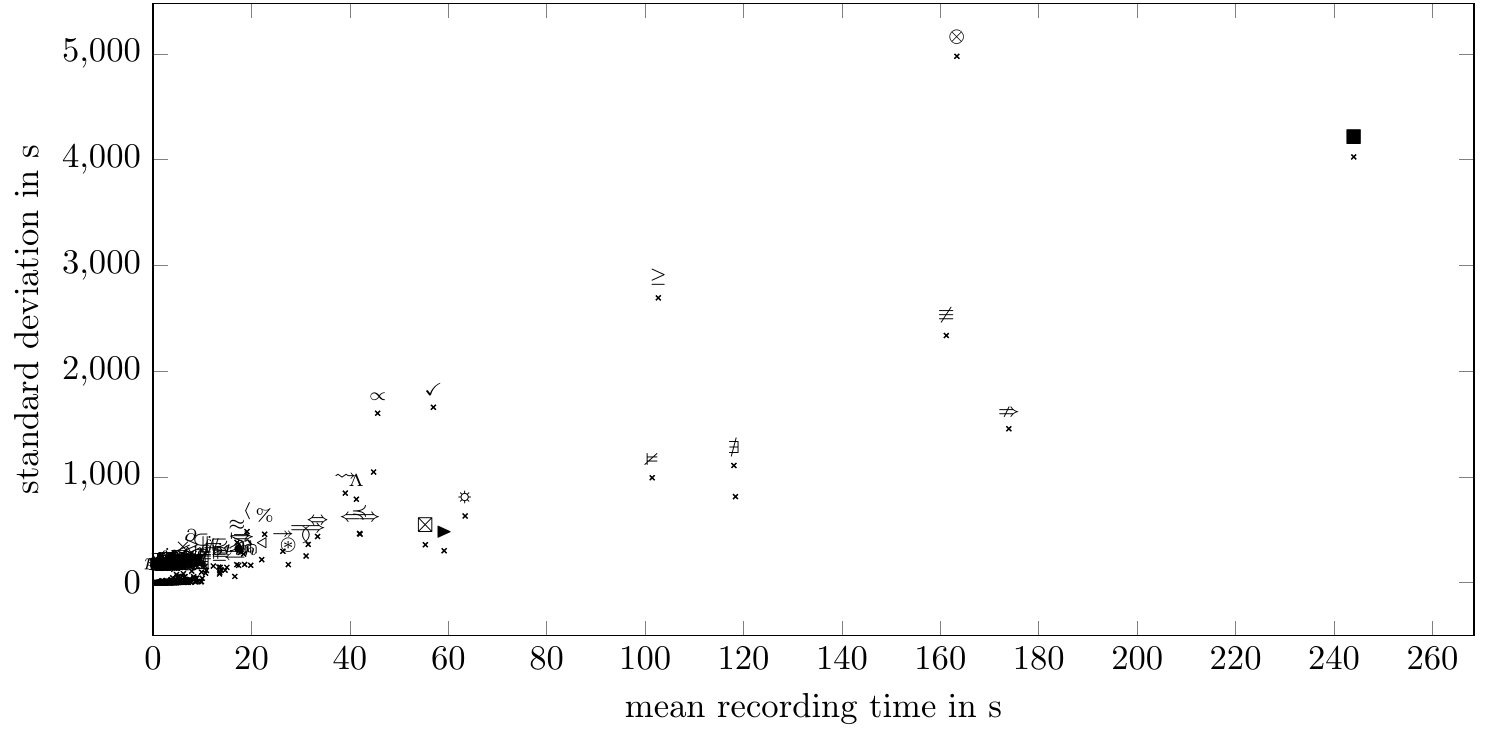}
    \caption{Mean and standard deviation of the recording time of symbols in
             seconds. Almost every symbol has a recording time of less than
             $\SI{20}{\second}$ and a standard deviation of less than
             $\SI{500}{\second}$. The high standard deviation indicates that
             there are some recordings with extremely wrong timestamps. Some
             recordings have a huge gap between two subsequent control points.
             The points seemed to be still in order, but the data looked
             as if the system clock was changed while the symbol was drawn.}
\label{fig:time-mean-std-deviation}
\end{figure}

\begin{table}[ht]
    \centering
    \begin{tabular}{crr|crr} 
    \toprule
    Symbol            & Mean         & std deviation & Symbol            & Mean       & std deviation \\\midrule
    \verb+\boxdot+    & $\SI{2864}{\milli\second}$ & $31.76 \cdot 10^6$ & \verb+\nsubseteq+ & $\SI{1994}{\milli\second}$ & $26.86 \cdot 10^6$\\
    \verb+\subsetneq+ & $\SI{1199}{\milli\second}$ & $18.27 \cdot 10^6$ & \verb+\psi+  & $\SI{324}{\milli\second}$ & $7.04 \cdot 10^6$\\
    \bottomrule
    \end{tabular}
    \caption{Mean and standard deviation of the recording time in milliseconds
             of symbols that are not shown in \cref{fig:time-mean-std-deviation}.}
\label{table:time-mean-std-deviation}
\end{table}

\paragraph{Other problematic user actions are:}
\begin{problemenumo}
    \item\label{itm:filled-areas} \textit{Filling areas}: Filling areas
          produces a lot of points. But the order and the number of those points
          is arbitrary in contrast to many --- eventually all --- other strokes.
          See \cref{fig:filled-area} on \cpageref{fig:filled-area} as an example
          for a recording with a filled area.
    \item\label{itm:strengthened-strokes} \textit{Strengthened strokes}:
          Sometimes users want to \enquote{strengthen} strokes. As
          with~\cref{itm:filled-areas}, the number of those strengthening points
          might vary a lot even for a single user. See
          \cref{fig:strengthened-stroke} on \cpageref{fig:strengthened-stroke}
          as an example for a recording with a strengthened stroke.
\end{problemenumo}

\begin{figure}[h!]
    \centering
    \includegraphics*[width=3cm]{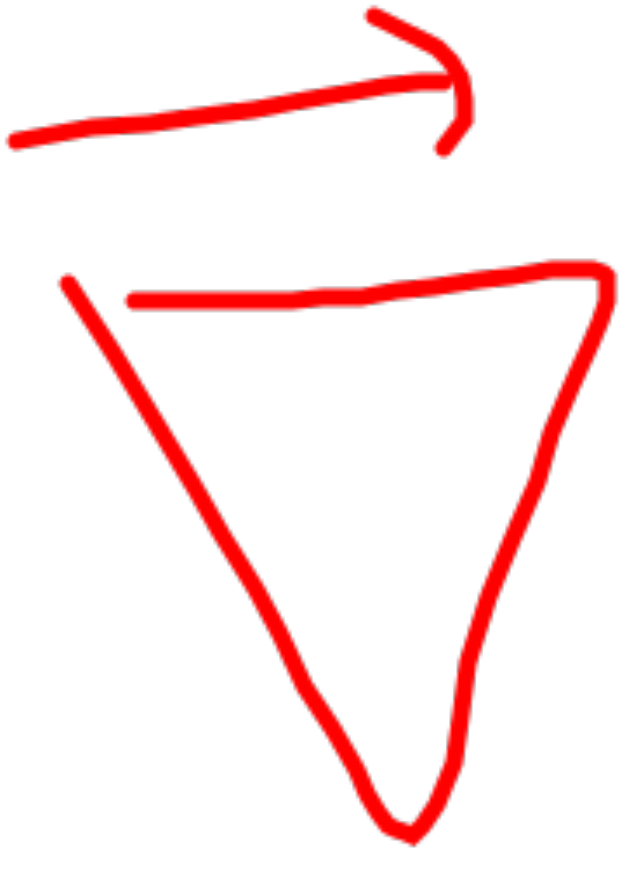}
    \caption{$\nabla$ written by the user \enquote{Marienkaefer}. It is an
             example where the user expects the system to recognize something
             different than the closest \LaTeX{} pendant
             $\overset{\rightarrow}{\nabla}$.}
\label{fig:nabla-marienkaefer}
\end{figure}

All recordings that suffered from problems
\crefrange{itm:h1-confusion}{itm:h4-manipulation} were excluded from the
evaluated dataset. For \cref{itm:h2-creativity}, the recording was additionally
marked as an image or as a member of the \enquote{trash} class. The trash
class was neither used for training nor for evaluation, but it could be used in
future to detect if a user wants to delete a recording he just drew.

Recordings that were multiple symbols (\cref{itm:h3-cherry-picking}) were
additionally annotated with the number of symbols for future complete formula
recognition.

\Crefrange{itm:d1-wild-points}{itm:d5-interrupted-stroke} are covered by
automatic methods which are explained in \cref{sec:preprocessing}.
\Cref{itm:d6-wrong-time} was ignored.

One reason why \cref{itm:h1-confusion} (symbol confusion) and
\cref{itm:h4-manipulation} (manipulation) are very difficult to note and to
resolve is that users might write something different when they use handwriting
compared to what they use in printed text. One example is the following: In
physics, it seems to be common to write $\overset{\rightarrow}{\nabla}$ in
handwritten text, but use $\nabla$ in \LaTeX{}. In that case, the classifier
should recognize
\cref{fig:nabla-marienkaefer} as $\nabla$, although the appearance is closer to
$\overset{\rightarrow}{\nabla}$.


\subsection{Data Cleansing}\label{subsection:Outlier-detection}

The data was collected by crowdsourcing. There were no restrictions and
everybody could enter data anonymously. In the case of Detexify, where over
\detexifyPercentage{} of the data comes from, this happened over 4~years.

This means a lot of the data is classified wrong.

In the case of the test set, all recordings were checked manually. But there
is too much data to manually check all recordings. So different techniques were
used to automatically find suspicious recordings.

The greedy time warping classifier, which is explained in
\cref{sec:system-a-greedy-matching}, was used to find recordings with
a high distance within all recordings a single symbol. The distance of every
recording to every other recording of the same symbol was measured. This means
when a symbol had $n$ recordings, there were $n \cdot (n-1)$ time warpings done.
Then the recordings were ordered descending by distance. They were reviewed
until at least 10 recordings in a row were classified correct.

The global features were used to find outliers. For every global feature in
\cref{sec:features}, the mean and the standard deviation of every symbol was calculated.
The symbols with highest standard deviation were examined. For those symbols, the
recordings were ordered descending and reviewed until at least 10 recordings in
a row were classified correct.

Neural network classifiers were trained and the errors they made were
examined for misclassified recordings.

All results in \cref{ch:Evaluation} were obtained after the data cleansing
steps.

\section{Preprocessing}\label{sec:preprocessing}
Preprocessing in symbol recognition is done to improve the quality and
expressive power of the data. It should make follow-up tasks like segmentation
and feature extraction easier, more effective or faster. It does so by resolving
errors in the input data, reducing duplicate information and removing irrelevant
information.

\subsection{Normalization: Scaling, Shifting and Resampling}
\textit{Scaling}\label{preprocessing:scale-and-shift} --- which is also called
\textit{size normalization} --- is done by many handwriting recognition systems,
but the way in which size normalization is done varies.\\
Single-symbol recognizers such as the one presented in~\cite{Kirsch} scale the
data points to fit into a unit square while keeping their aspect ratio. To do
so, the bounding box of the symbol is taken and everything is scaled according
to this bounding box. Afterwards, the points are shifted to the
$[0, 1] \times [0, 1]$ unit square. It was shown in~\cite{Huang09,Kirsch} that
this kind of preprocessing notably boosts classification accuracy.\\
\cite{Guyon91}~shifts the symbol to $[-1, 1] \times [-1, 1]$. That might be
better for the training of neural networks as it might lead to a mean feature
value of 0 (see \cref{sec:feature-standardization} for more information).\\
An algorithm that does scaling and shifting to $[-0.5, 0.5] \times [-0.5, 0.5]$
while keeping the aspect ratio is given in pseudocode on
\cpageref{alg:scale-and-shift}. Three implementation variants of the scale and
shift algorithm are explained and evaluated on
\cpageref{scale-and-shift-implementations}.

Everything that makes the recording artificially bigger makes scaling less
effective. That includes wild points (\cref{itm:d1-wild-points}) and hooks
(\cref{itm:d4-hook}). Algorithms that can deal with
those problems are described in \cref{sec:noise-reduction}.

Another method to normalize data is \textit{resampling}. This is called
\textit{stroke length normalization} in~\cite{Tappert90}.
\cite{Guyon91}~resampled characters and digits to 81 points each, where
different strokes were connected by \enquote{pen-up} segments. They resampled to
get points regularly spaced in arc length, not in time. \cite{Manke01}~also 
resampled the points to be equidistant in space, but they used a distance of
$\frac{\text{corpus height}}{13}$. They found an improvement of
$\SI{5}{\percent}$ with this preprocessing step.
\cite{ICASSP-94}~also resampled data to get points regularly spaced in arc
length, but they encoded speed as an extra feature. A simple resampling
algorithm that interpolates strokes linearly and spaces points equidistant in
time for a fixed number of points. \Cref{alg:resampling} on
\cpageref{alg:resampling} shows this simple resampling algorithm in pseudocode.
\clearpage

\subsection{Noise Reduction}\label{sec:noise-reduction}
\paragraph{The following list of noise reduction techniques was created by~\cite{Tappert90}
and is still up-to-date.}
\begin{itemize}
    \item \textbf{Dot reduction} reduces dots to single points. Sometimes
          multiple points get recorded although the user wanted to make only a
          single point, e.g.\ for one of the following symbols: $\cdot$, ., 
          $\dots$, $\vdots$, $\ddots$, i, $\therefore$, $\because$. This can be
          detected by calculating the maximum distance $d$ two points in a
          stroke have. If $d$ is smaller than a threshold, then it is a single
          point. In that case all points of the stroke get reduced to a single
          dot. This dot could be the center of mass of all points in the stroke.
          The algorithm can be found in pseudocode on
          \cpageref{alg:dot-reduction}.
    \item \textbf{Dehooking} is the removal of hooks (see \cref{itm:d4-hook})
          which the author did not want to write. Hooks appear sometimes at the
          beginning or the end of strokes. Examples can be seen in
          \cref{fig:hooks}. An algorithm for dehooking is described
          in~\cite{Huang09}.
    \item \textbf{Filtering} is the process of removing points by some criteria.
          Those criteria include:
          \begin{itemize}
              \item Duplicate points as applied in~\cite{Huang09,Guerfali93},
              \item Enforcing a minimal distance between consecutive
                    points~\cite{Tappert90}.
              \item Maximum velocity / acceleration~\cite{Division87}
              \item Enforcing a minimal change in direction~\cite{Tappert90}.
          \end{itemize}

          Occasionally occuring control points that were generated by device
          errors are one reason to apply a filtering preprocessing step. Those
          points are also called \textit{wild points}
          (\cref{itm:d1-wild-points}). Filtering wild
          points might be difficult for humans when the points could also be
          decorations as shown in \cref{fig:wild-points-or-decorations}.

          One way to detect wild points is by measuring the speed from the last
          point to the wild point. If that speed is too high, it can be assumed
          that it is a wild point.

        \begin{figure}[tb]
            \centering
            \subfloat[Raw data ID 149550]{
                \hspace*{3em}
                \includegraphics[height=0.3\linewidth, keepaspectratio]{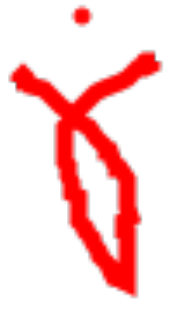}
                \hspace*{3em}
\label{fig:raw-data-id-149550}
            }%
            \subfloat[Raw data ID 138361]{
                \includegraphics[height=0.3\linewidth, keepaspectratio]{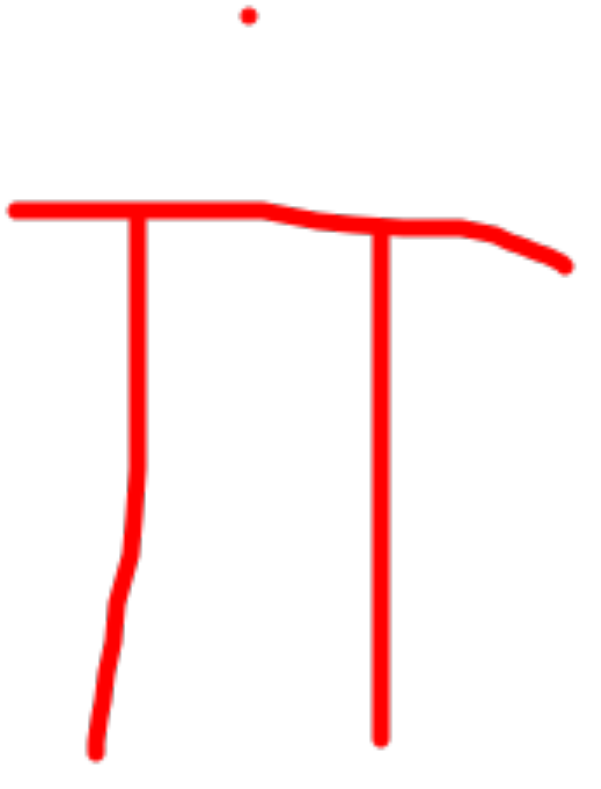}
\label{fig:raw-data-id-138361}
            }%
            \caption{Examples of recordings with a dot over the symbol. It is
                     not possible to tell if that is a wild point or a
                     decoration which was intended by the user.}
\label{fig:wild-points-or-decorations}
        \end{figure}

    \item\label{preprocessing:weighted-average-smoothing}\textbf{Smoothing} can
          be done in at least two ways. An approach that was used quite often is
          applying a weighted average~\cite{Groner66,Division87,Arakaw83}.
          \Cref{alg:weighted-average-smoothing} describes in pseudocode how
          weighted average smoothing can be implemented.

          It takes three weighting parameters
          $\theta_1, \theta_2, \theta_3 \in [0,1]$ and recalculates the point
          coordinates of every point $p_i$ except the first point $p_1$ and the
          last point $p_n$ like this:

          \[p_i' \gets \theta_1 \cdot p_{i-1} + \theta_2 \cdot p_{i} + \theta_3 \cdot p_{i+1}\]

\label{preprocessing:douglas-peucker-smoothing}
          Another way to do smoothing would be to reduce the number of points
          with the Douglas-Peucker algorithm to the most relevant ones and
          then interpolate those points. The Douglas-Peucker stroke
          simplification algorithm is usually used in cartography to simplify
          the shape of roads. The Douglas-Peucker algorithm works recursively to
          find a subset of control points of a stroke that is simpler and still
          similar to the original shape. The algorithm adds the first and the
          last point $p_1$ and $p_n$ of a stroke to the simplified set of points
          $S$. Then it searches the control point $p_i$ in between that has
          maximum distance from the \gls{line} $p_1 p_n$. If this distance is above a
          threshold $\varepsilon$, the point $p_i$ is added to $S$. Then
          the algorithm gets applied to $p_1 p_i$ and $p_i p_n$ recursively.
          Pseudocode of this algorithm is on \cpageref{alg:douglas-peucker}.
          It is described as \enquote{Algorithm 1} in~\cite{Visvalingam1990}
          with a different notation.
    \item\label{alg:stroke-connect-preprocessing}\textbf{Connecting strokes}
          should be done if \cref{itm:d5-interrupted-stroke} (see
          \cpageref{itm:d5-interrupted-stroke}) occurs. This can be detected by
          measuring the distance between the end of one stroke and the beginning
          of the next stroke. If this distance is below a threshold, then the
          strokes are connected.\\
          \cite{Guerfali93}~describes that such maliciously disconnected 
          components can get detected by observing angular continuity and the
          shortness of distance between two strokes. The distance between two
          consecutive strokes $(s_i, s_{i+1})$ is calculated by measuring the
          euclidean distance from the last point of $s_i$ to the first point of
          $s_{i+1}$. As this error seems just to split strokes, but not miss any
          control point, it might result in control points of subsequent strokes
          being very close. So one could also use only the distance and a
          distance threshold to determine if two strokes should be connected.
    \item \textbf{Deskewing} corrects character slant. Although this technique
          was applied by some authors~\cite{Bozinovic1989,Guerfali93,IWFHR94},
          it seems not to be applicable to the domain of mathematical
          handwriting, because on the one hand symbols might occur in variations
          with slant, like $\rightarrow$ and $\nearrow$. On the other hand it is
          questionable if slant is as consistent with symbols as it is with
          cursive handwriting.
\end{itemize}

\begin{figure}[ht]
    \centering
    \subfloat[Interrupted stroke]{
        \includegraphics[height=0.3\linewidth, keepaspectratio]{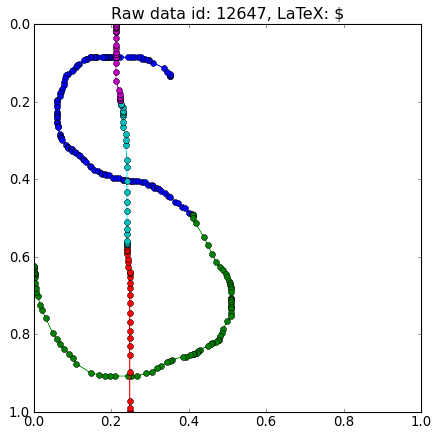}
\label{fig:interrupted-stroke}
    }%
    \subfloat[Filled area]{
        \includegraphics[height=0.3\linewidth, keepaspectratio]{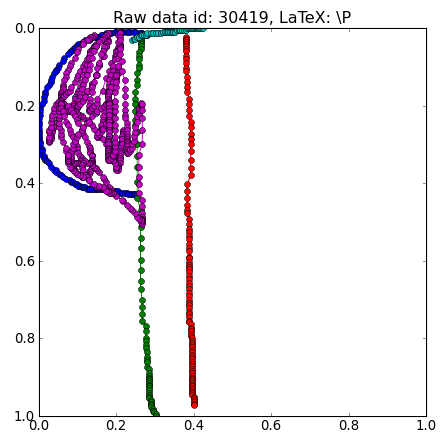}
\label{fig:filled-area}
    }

    \subfloat[Strengthened stroke]{
        \includegraphics[height=0.3\linewidth, keepaspectratio]{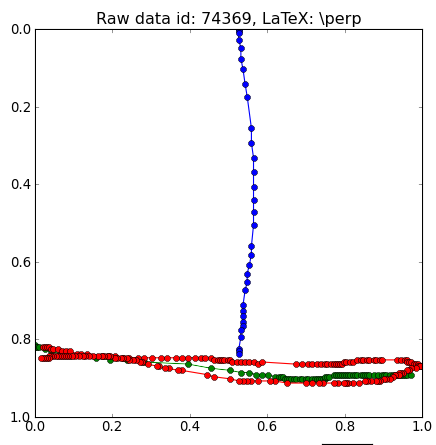}
\label{fig:strengthened-stroke}
    }%
    \subfloat[Multiple stroke drawing]{
        \includegraphics[height=0.3\linewidth, keepaspectratio]{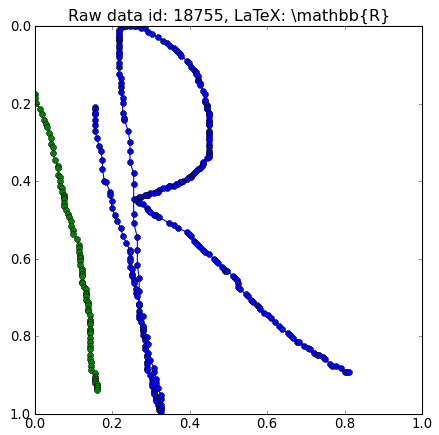}
\label{fig:multiple-stroke-drawing}
    }%
    \caption{Every image shows a recording after scaling and shifting by
             visualizing the stored points and connecting them with a straight
             line. Points and the lines between them are colored with the same
             color if they belong to the same stroke and otherwise with a
             different color.}
\label{fig:problematic-data}
\end{figure}
\clearpage

\subsection{Order of Preprocessing Steps}
\paragraph{There are multiple dependencies regarding the order in which the
mentioned preprocessing steps should be executed:}
\begin{itemize}
    \item Duplicate point removal is dot reduction with any minimum distance
          $> 0$.
    \item Dot reduction should be done before wild point filtering is done,
          because multiple points might get reduced to a single dot. Hence
          wild point detection might improve, because the reduced dot is
          isolated a little bit more.
    \item The scaling step depends on the size of the bounding box. As
          wild point removal and smoothing could change that size, those two
          algorithms should be applied before smoothing gets applied.
    \item Everything that changes the number of points should be done before
          resampling. That includes (wild) point filtering and smoothing.
\end{itemize}

Those dependencies and the preprocessing parameters are visualized in
\cref{fig:preprocessing-dependencies-and-parameters}.

\Cref{table:preprocessing-summation} lists all presented preprocessing
algorithms with the range of their parameters.

\begin{figure}[htb]
    \centering
    \includegraphics*[width=\textwidth]{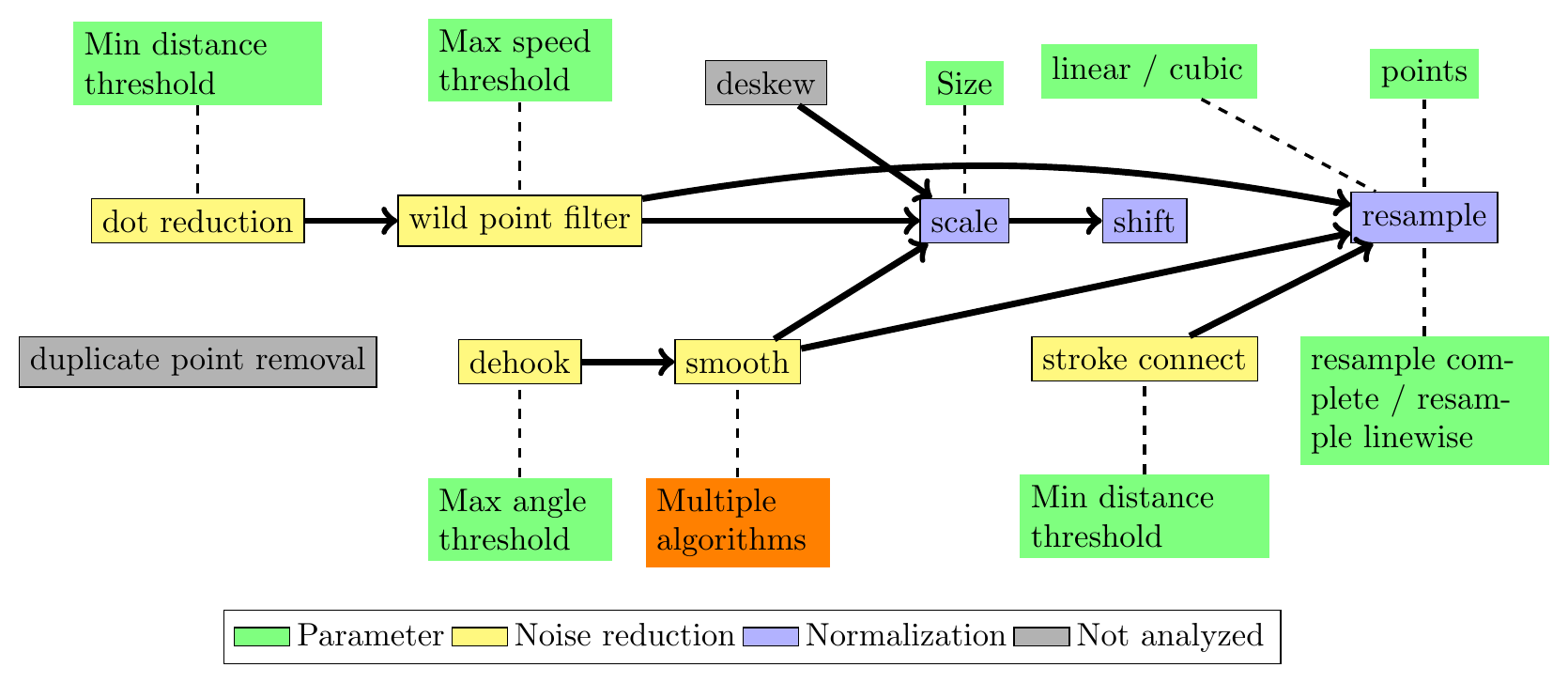}
    \caption{It makes sense to order the application of preprocessing algorithms
             --- if they are applied at all --- as shown in the diagram. The
             noise reduction algorithms are yellow, normalization algorithms
             are blue, green are parameter. Smoothing has too many variants
             and parameters to show them all in this diagram. Deskewing was not
             analyzed and is very likely not applicable in the domain of on-line
             handwritten recognition of single mathematical symbols. Duplicate
             points removal is only a special variant of dot reduction.}
\label{fig:preprocessing-dependencies-and-parameters}
\end{figure}

\clearpage


\section{Data Augmentation}\label{sec:Data-augmentation}

Obtaining a lot of original data can be difficult. Although projects like Amazon
Mechanical Turk might help, one could eventually still see the need of more
data. One way to get more data and to make the classifier invariant to some
transformations is by giving \enquote{virtual examples} that incorporate those
invariances \cite{Schoelkopf1996}. That means domain knowledge is used to
artificially generate more data from original data.

For on-line handwriting recognition, invariant transformations could be

\begin{itemize}
    \item Rotation by a maximum degree in the range of $(\SI{-22.5}{\degree},
          \SI{+22.5}{\degree})$ as symbols like $\rightarrow$ and
          $\nearrow$ are already transformations of $\SI{45}{\degree}$.
          The rotation center could be the center of mass (arithmetic mean of
          coordinates)
    \item Small random movements of single points independently from other
          points. However, this has to be used very carefully because of symbols
          like $\rightarrow$, $\leadsto$ and $\rightsquigarrow$ where those
          movements could easily lead to recordings that cannot be distinguished.
    \item Scaling with or without respect to the aspect ratio. This might also
          have side effects like $\pi$ and $\Pi$ or $\shortrightarrow$,
          $\rightarrow$ and $\longrightarrow$.
\end{itemize}

Other variations like scaling with respect to the aspect ratio or shifting do
only make sense when the preprocessing algorithm that removes those invariances
is not used. The use of data augmentation algorithms can break invariances
created by preprocessing steps. An example is that after applying a scaling
algorithm, one expects all recordings to have the same bounding box size.
However, after a recording was rotated that is no longer the case.

\section{Features}\label{sec:features}
A number of different features have been suggested for on-line handwriting
recognition. They can be grouped into local features and global features. Local
features apply to a given point on the drawing plane and sometimes even only to
point on the drawn curve whereas global features apply to a complete stroke or
even the complete recording.

\subsection{Local Features}
The following local features were used for on-line handwriting recognition.
However, most features were used as part of a bigger system without evaluating
the effect of the single feature.

\begin{itemize}
    \item \textbf{Coordinates} of the current point are used by~\cite{Guyon91}.
    \item \textbf{Speed} has been used by~\cite{ICASSP-94}, but
          \cite{Kosmala98,Kosmala11} suggest that speed is a bad feature, 
          because they think that speed is \enquote{highly inconsistent}.
    \item \textbf{Binary pen pressure} has been used
          by~\cite{Kosmala98,Kosmala11,ICASSP-94,Manke94,Guyon91}.
    \item \textbf{Direction} has been used by~\cite{Manke95,Huang06}.
The \textbf{direction} at the point $i$ can be described by the vector
$(\cos \theta(i), \sin \theta(i))$ as described in~\cite{Guyon91}:
\begin{align}
    \cos \theta(i) &= \frac{\Delta x^{(i)}}{\Delta s^{(i)}}\\
    \sin \theta(i) &= \frac{\Delta y^{(i)}}{\Delta s^{(i)}}
\end{align}
where
\begin{align}
    \Delta x (i) &= x^{(i+1)} - x^{(i-1)}\\
    \Delta y (i) &= y^{(i+1)} - x^{(i-1)}\\
    \Delta s (i) &= \sqrt{{(\Delta x(i))}^2 + {(\Delta y (i))}^2}
\end{align}
    \item \textbf{Curvature} has been used by~\cite{Groner66,Manke95,ICASSP-94,Guyon91}.
          It is calculated in~\cite{Guyon91} by the angle of two neighboring
          lines like this:
          \begin{align}
              \varphi(i)      &= \theta(i+1) - \theta(i-1)\\
              \cos \varphi(i) &=\cos \theta (i-1) \cdot \cos \theta (i+1)\\
                              &+\sin \theta (i-1) \cdot \sin \theta (i+1)\\
              \cos \varphi(i) &=\cos \theta (i-1) \cdot \cos \theta (i+1)\\
                              &-\sin \theta (i-1) \cdot \sin \theta (i+1)
          \end{align}
    \item\label{feature:bitmap-environment} \textbf{Bitmap-environment} has been
          used by~\cite{Manke94}. This feature is a $3 \times 3$ pixel
          environment around the current point. It allows the recognizer to
          determine points that cross or touch strokes. Adding this feature
          reduced the error by \SI{50}{\percent} compared to using only
          coordinates, the direction, curvature and speed.
    \item \textbf{Hat-Feature} has been used by~\cite{ICASSP-94,Manke00}.
\end{itemize}

\subsection{Global Features}
\begin{itemize}
    \item\label{feature:recurvature}\textbf{Re-curvature} is defined
          in~\cite{Huang06,Huang09} as the ratio between the height of a stroke
          and the distance between its start and end point. It is not clear
          if this distance was meant to be the euclidean distance or the
          distance on the stroke. Both variants were tried, but the distance
          on the stroke gives much better evaluation results. So it was chosen
          to use the feature
          \[\text{re-curvature}(stroke) = \frac{\text{height}(stroke)}{\text{length}(stroke)}\]
    \item \textbf{Center point} for every single stroke was used
          in~\cite{Huang06}. A center point of a stroke is the arithmetic mean
          of the coordinates.
    \item \textbf{Stroke length} was used in~\cite{Huang06}. It can be
          calculated by using the summed \gls{line segment} length after a
          linear interpolation step.
    \item \textbf{Number of strokes} was used in~\cite{Huang09}.
    \item \textbf{Sequence features}
    \begin{itemize}
        \item \textit{Pen-tip sequence}:~\cite{Kirsch} used the raw pen-tip
              sequence combined with \gls{DTW} variants to recognize
              mathematical symbols. Other authors like~\cite{Koschinski95} used
              pen-tip sequences, too, but made use of \glspl{HMM} or \glspl{ANN}
              to recognize symbols.
        \item \textit{Zone sequences} are used by~\cite{Brown1964,Hanaki80}. The
              idea is to recognize symbols by dividing the box in which the
              character is written into zones. By examining the position of the
              pen-tip a sequence of zones can be generated for a written
              symbol.
        \item \textit{Direction sequences} were used
              in~\cite{Impedovo1976,Powers1973}.
    \end{itemize}
    \item \textbf{Aspect ratio} of the bounding box of the recording.
\end{itemize}

There are other global features used for off-line handwriting recognition which
will not be examined. Examples are Pseudo-Zernike moments and Shadow Code features
which were used in~\cite{Khotanzad}.


\chapter{Domain Independent Classification Steps}

The previous chapter shows the used data as well as preprocessing steps and
features that can be found in on-line \gls{HWR}. This chapter introduces
some general methods that can be applied in any classification task of time
series data. At this point we have pairs  of feature vectors $x \in \mdr^n$ and
class labels $y$. The set of those pairs $(x, y)$ is split into three distinct
subsets: A training set, a validation set and a test set. The training set can
be used by a learning algorithm to adjust internal parameters. However, the
training algorithm could be able to adjust too much and create a recognizer that
works well on the training set but much worse on new examples. Hence the
validation set is used to detect when the algorithm suffers from overfitting.
The test set on the other hand only gets used when the training is finished and
the system can be evaluated.

Although a lot of learning algorithms like \glspl{GMM}, \glspl{SVM}, $k$-Nearest
Neighbors and even more can be applied for classification tasks,
only two are explained and evaluated: \Gls{GTW} and \glspl{MLP}. The
\gls{GTW} classifier is easy to implement and works reasonably well with only
a few training examples, but it is slow in evaluation. \Glspl{MLP} on the other
hand are harder to implement, take longer to train, but evaluate new data faster
and with higher recognition rates if enough data is available as showed in
\cref{ch:Evaluation}.


\section{Feature Enhancement}\label{sec:feature-enhancement}\label{sec:feature-standardization}

Feature enhancement algorithms can be used to make the already calculated
features more useful for training algorithms. The effect of those algorithms
depends on both, the data and the used learning algorithm.

An important subset of the feature enhancement algorithms are those that
reduce the dimensionality. \Gls{PCA} and \gls{LDA} are such algorithms.

One simple feature enhancement is \textit{feature standardization} sometimes
also called \textit{feature normalization}. For some learning algorithms it is
useful if the different features have a mean of $0$ and either a similar range
or a similar variance. Feature standardization gives this property.

Feature standardization is done by calculating the mean $\overline{x}$ of all
feature vectors $x \in T$ in the training set $T$. Then, before the training
gets applied and before every evaluation, the mean gets subtracted from every
feature vector $x_i$:

\[ x' \gets x_i - \overline{x} \]

This is called \textit{mean normalization}. In order to standardize features one
has to divide $x'$ by the range of values $\max(T) - \min(T)$ of the training
set.

If the feature is only divided by either the range or the variance it is called
\textit{feature scaling}.

\section{Greedy Time Warping}\label{sec:system-a-greedy-matching}
A web system for on-line handwritten symbol recognition was implemented and
is described in~\cite{Kirsch}. It uses an \gls{GTW} algorithm which is
similar to \Gls{DTW}.

The idea of the \gls{GTW} algorithm is to calculate how far the points between
two recordings $A$ and $B$ have to be moved to match each other. The algorithm
calculates a distance $d(A,B)$ of two recordings. This is done with help of the
squared euclidean distance $\delta$.

In the following $a_i$ denotes the $i$th point of the recording $A$ and
$b_i$ the $i$th point of the recording $B$. $\alpha_i$ denotes the number of
the point in the recording $A$ that was moved in step $i$ and $\beta_i$ denotes
the number of the point in recording $B$ that was moved in step $i$.

The distance $\delta(a_{\alpha_0=0}, b_{\beta_0=0})$ between the first
points of $A$ and $B$ is calculated. Then the minimum of
$\delta(a_{\alpha_{i+1}=\alpha_i+1}, b_{\beta_{i+1}=\beta_i})$,
$\delta(a_{\alpha_{i+1}=\alpha_i+1}, b_{\beta_{i+1}=\beta_i+1})$ and
$\delta(a_{\alpha_{i+1}=\alpha_i}, b_{\beta_{i+1}=\beta_i+1})$ is added to the already
calculated distance. If $\alpha_i + 1$ does not exist because $\alpha_i$ is
already the number of points in $A$ then only the last distance is taken.
Similar, if $\beta_i + 1$ does not exist because $\beta_i$ is already the
number of points in $B$ then only the first distance is taken.

Pseudocode is on \cpageref{alg:greedy-matching}.
\section{The Perceptron Algorithm}\label{sec:perceptron}

The idea of developing an algorithm that has similar capabilities as the brain
probably began in 1943 when Warren~McCulloch and Walter~Pitts described the
binary threshold unit in~\cite{McCulloch1943}. This work was later continued by
Frank Rosenblatt who invented the perceptron algorithm in
1958~\cite{Rosenblatt58}. The perceptron is a function

\begin{align*}
    p_{w, \varphi}:&\mathbb{R}^n \rightarrow \mathbb{R} & p_{w, \varphi}(x) :&= \varphi(w^T \cdot x)\\
    \varphi    :&\mathbb{R}^{\hphantom{n}} \rightarrow \mathbb{R} &
    \varphi(x) :&= \begin{cases} 1 &\text{if } x > 0\\
                                 0 &\text{otherwise}
                  \end{cases}
\end{align*}

This function, or rather the visualization of it as shown in
\cref{fig:artificial-neuron}, is also called an \textit{artificial neuron}.
Artificial neurons are inspired by biological neurons such as the one
illustrated in \cref{fig:biological-neuron}. In biological neurons, signals are
sent within the cell by charged particles, so called \textit{ions}. But before a
biological neuron sends a signal, a threshold charge has to be reached at the
axon hillock. This threshold charge is called \textit{action potential}. The
action potential can be reached by multiple factors, but the one which is most
interesting are charges send by other neurons. The closer other axon terminals
are to the axon hillock, the more their signal contributes to reaching the
action potential. If the stimulated neuron has reached the action potential, it
sends a signal.

\begin{figure}[bt]
    \centering
    \subfloat[Biological Neuron]{
        \includegraphics*[width=0.48\linewidth, keepaspectratio]{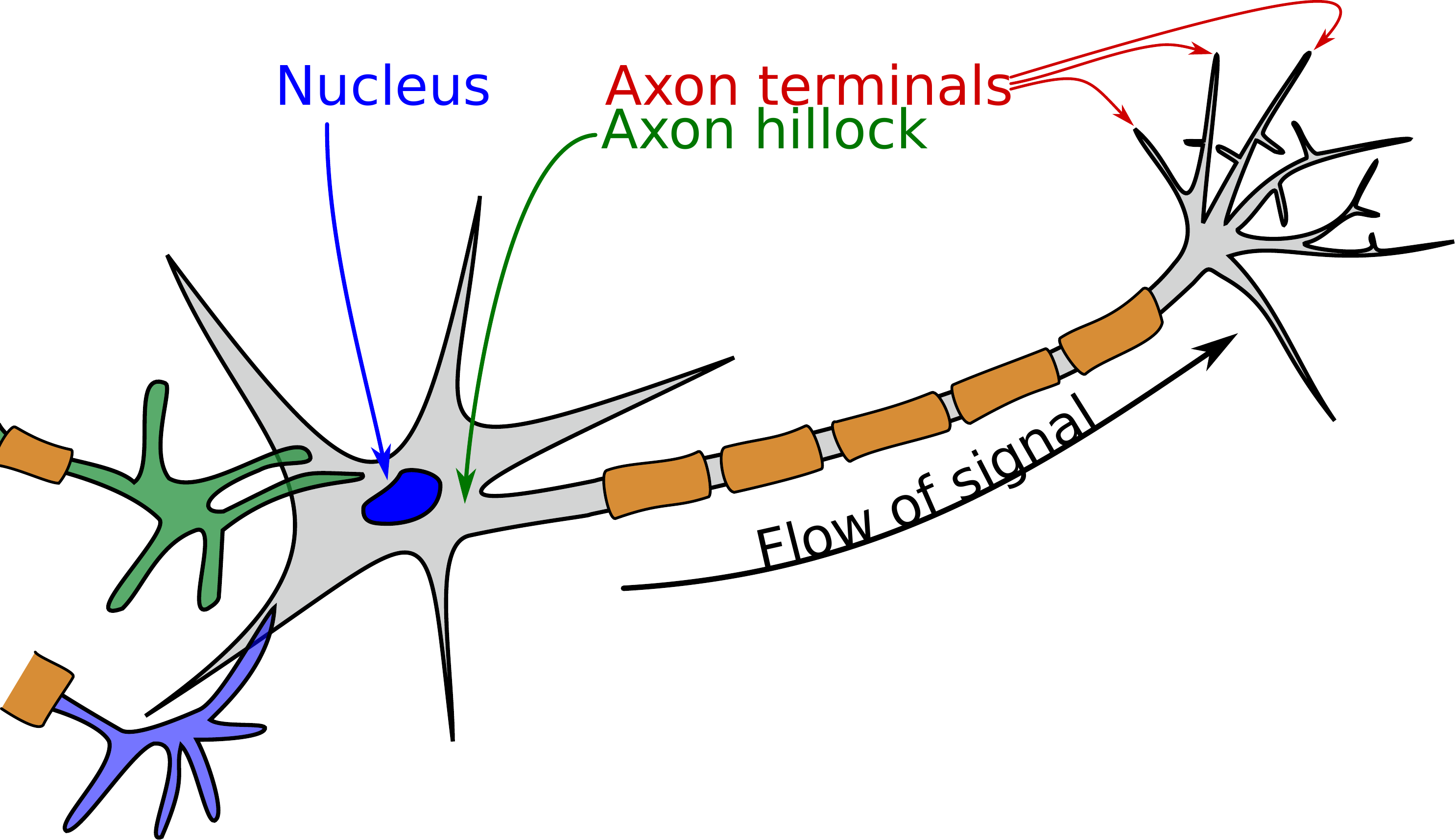}
\label{fig:biological-neuron}
    }%
    \subfloat[Artificial Neuron]{
        \includegraphics*[width=0.48\textwidth]{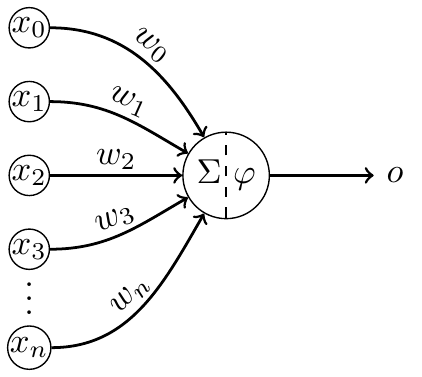}
\label{fig:artificial-neuron}
    }%
\label{fig:artificial-and-biological-neuron}
    \caption{Both neurons receive weighted input signals, apply a function to
             that sum and send an output signal.}
\end{figure}

Artificial neurons are similar as they receive input signals and give an output
signal. Those input signals get weighted and summed up. Then an activation
function $\varphi$ is applied to the weighted sum. However, there are important
differences, too. Artificial neurons use different activation functions. In most
applications, artificial neurons use a differentiable function which sends a
continuous signal whereas a biological neuron encodes the information by the
frequency it sends a signal. Biological neurons send signals asynchronously, but
PCs work synchronously. More details can be found in \cite[p. 1001--1026]{Lodish08}
and \cite[p. 1047--1061]{Campbell08}.

An application of the perceptron is a  binary classifier where
the parameters $w \in \mathbb{R}^n$ have to be learned. In the context of
supervised learning there are already $m$ training examples of input vectors
$x_i \in \mathbb{R}^n, i \in \Set{1, \dots, m}$ together with the desired output
$y \in \Set{0, 1}$ given. The output is called the \textit{class} and
$x_i^{(j)}$ is the $j$th \textit{feature} of the $i$th training example.

When such a training set $T = \Set{(x_i, y_i) | i \in \Set{1, \dots, m}}$ is
given, we want to find a choice for $w$ that is best for that set according to
an differentiable error function $E: \mathbb{R}^n \rightarrow \mathbb{R}_0^+$.
The error function $E$ can be modified to represent not only the error on the
training set, but also additional targets. Therefore it is also called loss
function, objective function or cost function.
\cite[p.89--92]{Mitchell97} describes in detail how the perceptron 
learns its weight parameters $w$.

We want to find the minimum of that function $E$. One way to find the minimum of a
function is by gradient descent. That means one starts at a random point $w$,
calculates the gradient at this point and \enquote{goes} in the direction of the
gradient, that means the weights are adjusted. This is commonly expressed as
\[w \gets w + \Delta w\]
and hence this learning method is called \textit{delta rule}. In this case
$w$ and $\Delta w$ are vectors where the single vector components are
\[\Delta w^{(j)} = - \eta \frac{\partial E}{\partial w^{(j)}}\]
where $\eta \in \mdr_{>0}$ is called the \textit{learning rate}. The training
algorithm will overshoot the minimum if it is too big, but when it is too small,
the training algorithm will make progress very slow.

A common way to visualize gradient descent is to imagine the error surface. It
is a surface in the $\mathbb{R}^{n+1}$ where $n$ dimensions are the possible
choices of the parameter $w \in \mathbb{R}^n$ and the last dimension is the
error $E(w) \in \mathbb{R}$. The form of that surface depends on the training
examples and the error function. As the error function uses the output of the
perceptron, it depends on the activation function. It follows that the
activation function $\varphi$ has to be differentiable. Hence the sign function
is not a good choice. A common choice for $\varphi$ is the
\textit{sigmoid function}:

\[\sigmoid(x) := \frac{1}{1+e^{-x}}\]

The perceptron classifier is able to make use of an arbitrary number of features
to distinguish two classes.

However, in the case of symbol classification there are more than two classes.
One way to solve this is by applying the \textit{one-vs.-rest} strategy. That
means for every class there is one classifier that tests if the recording
belongs its class. When a recording should get classified, the output of
every single neuron gets calculated. Then the softmax function gets applied
to the vector of outputs $o$ of those neurons.

\[\softmax: \mathbb{R}^n \rightarrow {[0, 1]}^n \qquad {\softmax(o)}^{(i)} := \frac{e^{o^{(i)}}}{\sum_{j=1}^{n} e^{o^{(j)}}}\]

The softmax function makes sure that every single value of the result is in
$[0,1]$ and that the sum of all values is exactly $1$. Furthermore, the order
of the values in that vector remains the same. One could say that
the softmax function transforms a vector of scores to a vector of probabilities.

Another mayor drawback of a single layer perceptron is the fact that it can only
classify data which is linearly separable in the feature space. The feature
space is usually an $\mathbb{R}^n$, where $n$ is the number of features. Every
recording of the training data is a point in that space. While the obtained
data might usually be in the $\mathbb{R}^2$ or $\mathbb{R}^3$, the features
might give relationships between this information. By a clever choice of
features one can make data that was not linear separable in the obtained space
separable in the feature space. In fact, one can make every training set
linearly separable by adding a new feature per training example that gives the
distance to that training example. But that would be a lot of features and
the resulting model would very likely suffer from overfitting. As the number
of training examples might be very high, even dimensionality reduction algorithms
like \gls{PCA} and \gls{LDA} could be difficult to apply.

For this reason it is desirable that the neural network is able to learn features by itself.
This can be achieved by using multiple layers, where every layer computes
a new set of features.

\section{Multilayer Perceptron}

A \gls{MLP} is organized in layers of artificial neurons. Every artificial neuron
is a function $p_{w,\varphi}$ with different weight vectors $w$ per artificial
neuron. Every layer has exactly one activation function $\varphi$, but
the activation functions of different layers may be different. Every layer
is fully connected with its predecessor and its successor.

The number of layers is in principle not limited and the number of neurons is not
limited either. However, the number of parameters between a layer $i$ with $n_i$
neurons and a layer $j$ with $n_j$ neurons is $n_i \cdot n_j$. That means for
subsequent layers with many neurons the number of parameters that have to be
learned gets very big.


\subsection{Notation}

A notation that is almost identical to the one in~\cite{Mitchell97} was chosen:

\begin{itemize}
    \item $\layernumber \in \mathbb{N}$ is the number of layers of the
          \gls{MLP}.
    \item $n_j \in \mathbb{N}$ is the number of neurons in layer
          $1 \leq j \leq \layernumber$.
    \item $(x_i, y_i) \in \mdr^{n_1} \times \mdr^{n_\layernumber}$ is a single
          training example of the training set $T$.
    \item $v^{(i)}$ is the $i$th element of a vector $v$.
    \item $x_{j,i} \in \mathbb{R}$ is the $i$th input to the $j$ neuron.
    \item $w_{j,i} \in \mathbb{R}$ is the weight from neuron $i$ to neuron $j$.
    \item $\net_j := \sum_{i \in \text{input neurons}} w_{j,i} x_{j,i}$ is
          the activation of neuron $j$, that means the value that the activation
          function is applied to.
    \item $o_j(x) := \varphi_j(\net_j(x)) = \varphi_j(\sum_{i \in \text{input neurons}} w_{j,i} x_{j,i})$
          is the output of neuron $j$ after the \gls{MLP} got $x$ as input
          feature. If $j$ was not in the last layer, there is at least one
          $i$ such that $o_j = x_{i,j}$ (note the order).
    \item $outputs$ is the set of all neurons in the last layer (the output
          layer).
    \item $D(j)$ is the \textit{Downstream}, that means the set of all
          neurons that have neuron $j$ as a direct input. That means the
          downstream of $j$ includes all neurons of the layer that is nearer to
          the output layer directly after the layer in which the neuron $j$ is.
\end{itemize}

\Cref{fig:perceptron-notation} visualizes the notation.

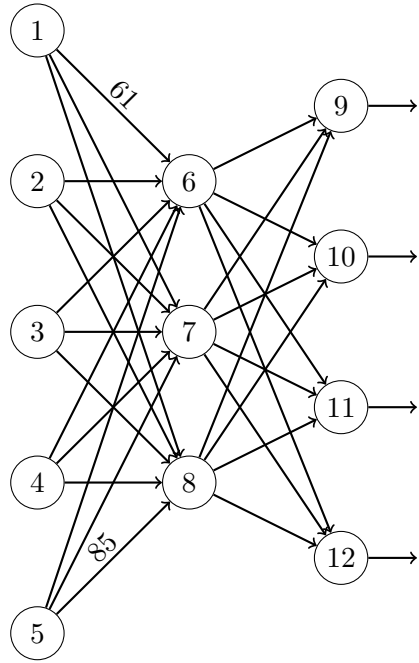
\begin{figure}[h!]
    \centering


\tikzstyle{neuron}=[draw,circle,minimum size=20pt,inner sep=0pt]

\tikzstyle{stateTransition}=[->, thick]

\begin{tikzpicture}[scale=2]
    \node (h11)[neuron] at (0,4) {$1$};
    \node (h12)[neuron] at (0,3) {$2$};
    \node (h13)[neuron] at (0,2) {$3$};
    \node (h14)[neuron] at (0,1) {$4$};
    \node (h15)[neuron] at (0,0) {$5$};
    \node (h21)[neuron] at (1, 3) {$6$};
    \node (h22)[neuron] at (1, 2) {$7$};
    \node (h23)[neuron] at (1, 1) {$8$};
    \node (h31)[neuron] at (2,3.5) {$9$};
    \node (h32)[neuron] at (2,2.5) {$10$};
    \node (h33)[neuron] at (2,1.5) {$11$};
    \node (h34)[neuron] at (2,0.5) {$12$};

    \draw[stateTransition] (h11) -- (h21) node [midway,above=-0.06cm,sloped] {$61$};
    \draw[stateTransition] (h12) -- (h21) node [midway,above=-0.06cm,sloped] {};
    \draw[stateTransition] (h13) -- (h21) node [midway,above=-0.06cm,sloped] {};
    \draw[stateTransition] (h14) -- (h21) node [midway,above=-0.06cm,sloped] {};
    \draw[stateTransition] (h15) -- (h21) node [midway,above=-0.06cm,sloped] {};

    \draw[stateTransition] (h11) -- (h22) node [midway,above=-0.06cm,sloped] {};
    \draw[stateTransition] (h12) -- (h22) node [midway,above=-0.06cm,sloped] {};
    \draw[stateTransition] (h13) -- (h22) node [midway,above=-0.06cm,sloped] {};
    \draw[stateTransition] (h14) -- (h22) node [midway,above=-0.06cm,sloped] {};
    \draw[stateTransition] (h15) -- (h22) node [midway,above=-0.06cm,sloped] {};

    \draw[stateTransition] (h11) -- (h23) node [midway,above=-0.06cm,sloped] {};
    \draw[stateTransition] (h12) -- (h23) node [midway,above=-0.06cm,sloped] {};
    \draw[stateTransition] (h13) -- (h23) node [midway,above=-0.06cm,sloped] {};
    \draw[stateTransition] (h14) -- (h23) node [midway,above=-0.06cm,sloped] {};
    \draw[stateTransition] (h15) -- (h23) node [midway,above=-0.06cm,sloped] {$85$};

    \draw[stateTransition] (h21) -- (h31) node [midway,above=-0.06cm,sloped] {};
    \draw[stateTransition] (h22) -- (h31) node [midway,above=-0.1cm,sloped] {};
    \draw[stateTransition] (h23) -- (h31) node [midway,above=-0.06cm,sloped] {};

    \draw[stateTransition] (h21) -- (h32) node [midway,above=-0.06cm,sloped] {};
    \draw[stateTransition] (h22) -- (h32) node [midway,above=-0.1cm,sloped] {};
    \draw[stateTransition] (h23) -- (h32) node [midway,above=-0.06cm,sloped] {};

    \draw[stateTransition] (h21) -- (h33) node [midway,above=-0.06cm,sloped] {};
    \draw[stateTransition] (h22) -- (h33) node [midway,above=-0.1cm,sloped] {};
    \draw[stateTransition] (h23) -- (h33) node [midway,above=-0.06cm,sloped] {};

    \draw[stateTransition] (h21) -- (h34) node [midway,above=-0.06cm,sloped] {};
    \draw[stateTransition] (h22) -- (h34) node [midway,above=-0.1cm,sloped] {};
    \draw[stateTransition] (h23) -- (h34) node [midway,above=-0.06cm,sloped] {};

    \draw[stateTransition] (h31) -- (2.5,3.5) node [midway,above=-0.1cm] {};
    \draw[stateTransition] (h32) -- (2.5,2.5) node [midway,above=-0.1cm] {};
    \draw[stateTransition] (h33) -- (2.5,1.5) node [midway,above=-0.1cm] {};
    \draw[stateTransition] (h34) -- (2.5,0.5) node [midway,above=-0.1cm] {};

\end{tikzpicture}
    \caption{Visualization of the used notation. Every neuron has an index
             (1--12). Only two weights were labeled:
             $w_{6,1} = 61$ and $w_{8,5} = 85$. The downstream of neuron 6 is
             $D(6) = \Set{9, 10, 11, 12} = D(7) = D(8)$. If this was a 3 layer
             perceptron where the neurons 9, 10, 11 and 12 are the output layer,
             then $outputs = \Set{9, 10, 11, 12}$.}
\label{fig:perceptron-notation}
\end{figure}

\subsection{Activation Functions}
The activation function of artificial neurons have to be differentiable and
their derivative has to be non-zero so that the gradient descent learning
algorithm can be applied. At least one layer should also be non-linear, because
linear combinations of linear functions are again linear functions. So if all
activation functions of a \gls{MLP} were linear, the complete \gls{MLP} would
only represent a linear function. This means the neural network could be
reduced to a \gls{MLP} without any hidden layer.

The last layer in classification tasks is often the $\softmax$ function. For
all other layers it is often the $\sigmoid$ function and sometimes also the
hyperbolic tangent $\tanh$. The advantage of $\tanh$ over the $\softmax$
function is that it converges faster when the absolute value of the argument is
big.

\Cref{fig:activation-functions} shows the activation functions $\sigmoid$,
$\tanh$ and the sign function.

\begin{figure}[ht]
    \begin{tikzpicture}[scale=1.0]
        \begin{axis}[
            legend pos=north west,
            axis x line=middle,
            axis y line=middle,
            grid = major,
            width=16cm,
            height=8cm,
            grid style={dashed, gray!30},
            xmin=-5,     
            xmax= 5,    
            ymin=-1,     
            ymax= 1,   
            xlabel=$x$,
            ylabel=$y$,
            tick align=outside,
            enlargelimits=true]
          \addplot[green!50!black, ultra thick] coordinates {(-5,-1) (0,-1) (0, 1) (5, 1)};
          \addplot[domain=-5:5, red, ultra thick,samples=500, dashed] {1/(1+exp(-x))};
          \addplot[domain=-5:5, blue, ultra thick,samples=500, dotted] {tanh(x)};
          \addlegendentry{sign function}
          \addlegendentry{$\sigmoid$}
          \addlegendentry{$\tanh$}
        \end{axis}
    \end{tikzpicture}
    \caption{A plot of the sign function, the sigmoid function and the
             hyperbolic tangent. All three functions can be used as activation
             functions in artificial neurons.}
\label{fig:activation-functions}
\end{figure}
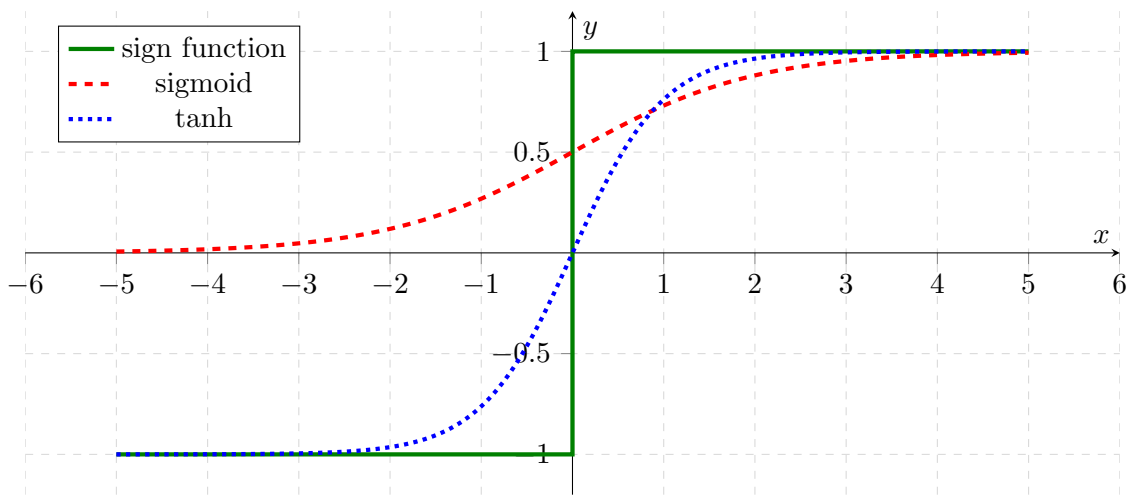

\subsection{Evaluation}
The evaluation of a neural network is very similar to the evaluation of a single
perceptron. For every perceptron of the first layer, the weights are multiplied
with the input. Those values are added and then the activation function gets
applied. This is repeated until the output of the first layer completely
calculated. Then exactly the same process is repeated with every following
layer.

However, this evaluation can also be expressed with matrix multiplications. The
input vector $x_I \in \mathbb{R}^{n_i - 1}$ gets extended by one value to the
vector $x \in \mathbb{R}^{n_i}$. This value is $1$ and represents the bias. Then
the vector $x$ is multiplied by weight matrix
$W \in \mathbb{R}^{n_i \times n_j}$ resulting in a vector $a$ which is also
called \textit{activation}:
\[x^T \cdot W = a^T \in \mathbb{R}^{n_j}\]
After that, all activation functions get applied point-wise to the activation
vector $a$ to get the output vector $o$ with the output of every
neuron of that layer.

The advantage of this matrix-wise expression is that some programs can
automatically parallelize this multiplication and that \glspl{GPU} can compute
those matrix multiplications directly.

\subsection{Supervised Training with Gradient Descent}\label{sec:training}
The gradient descent algorithm is a supervised algorithm for training
\glspl{MLP}. Just like the perceptron algorithm in \cref{sec:perceptron} it
needs an error function which can be minimized. \Gls{CE} is a possible choice
for \glspl{MLP} with a softmax output layer:

\begin{align*}
    E_{x_i}: &\mdr^{n_1 \times n_2} \times \mdr^{n_2 \times n_3} \times \cdots \times \mdr^{n_{\layernumber-1} \times n_\layernumber} \rightarrow \mdr_{\geq 0}\\
    E_{x_i}(W) :&= -\sum_{k=1}^{n_\layernumber} \left ( {y_i^{(k)} \log (o(x_i)^{(k)}) + (1-{y_i^{(k)}}) \log(1-o(x_i)^{(k)})} \right )\\
    E_B: &\mdr^{n_1 \times n_2} \times \mdr^{n_2 \times n_3} \times \cdots \times \mdr^{n_{\layernumber-1} \times n_\layernumber} \rightarrow \mdr_{\geq 0}\\
    E_B(W) &= \sum_{x_i \in B} E_{x_i}(W)
\end{align*}

where $E_{x_i}$ is the error for a single training example and $E_B$ with
$\emptyset \neq B \subseteq T$ is called a \textit{mini-batch}. Different
choices of $B$ lead to different training modes as explained in
\cref{subsec:batch-minibatch-stochastic-gradient-descent}.

There are other error functions like the \gls{CFM} or
\gls{MSE}~\cite{Hampshire1989}. However, in the following describes only the
training with the \gls{CE} function.

The error function $E_B$ is to be minimized. The gradient descent algorithm with
batch gradient descent converges to a local minimum if the learning rate is
decreased while applying gradient descent multiple times.

As the error is the sum of non-negative values, we get a lower error by
minimizing the error for every single training example if the learning rate
$\eta$ is low enough. However, it should be noted that those minimizations are
not independent. This means the global error could increase with single
stochastic gradient descent and single mini-batch gradient descent steps,
although the learning rate is low.

The training algorithm is

\begin{algorithm}[H]
    \begin{algorithmic}
        \Function{train}{$T$, $W$}
            \For{$\text{epoch}\gets 1$; $\text{epoch} \leq 1000$; $\text{epoch}\gets \text{epoch}+1$}
                \ForAll{$(x, t_x) \in T$}
                    \ForAll{weights $w_{j,i}$}
                        \State$\displaystyle w_{j,i} \gets w_{j,i} - \eta \frac{\partial E_{\Set{x}}}{\partial w_{j,i}} (W)$
                    \EndFor%
                \EndFor%
            \EndFor%
        \EndFunction%
    \end{algorithmic}
\caption{Stochastic Gradient Descent}
\label{alg:gradient-descent}
\end{algorithm}

where the number of epochs could be adjusted or changed to another stopping
criterion like a threshold for the change in validation error or the value of the
cost function.

Computing the partial derivatives
$\frac{\partial E_B}{\partial w_{j,i}}$ is not a trivial task, but
it is explained in detail in \cite{Mitchell97}.

\paragraph{Finally, the weight update rule can be formulated as}

\begin{align}
    w_{j,i} &\gets w_{j,i} + \Delta w_{j,i}\\
    \Leftrightarrow w_{j,i} &\gets w_{j,i} + \eta \delta_j x_{j,i}
\end{align}

where $\delta_j$ is a term that depends on the layer and is recursively
defined. For \gls{CE} as an error function, an output layer that makes use
of the softmax activation function and sigmoid activation functions in all
hidden layers it is

\begin{align*}
    \delta_j = \begin{cases}
        \frac{1}{|B|} \sum_{x_i \in B} \left (y_i^{(j)} - o_j(x_i) \right) &\text{if } j \in outputs\\
        \frac{1}{|B|} \sum_{x_i \in B} \left (o_j(x_i) (1-o_j(x_i)) \sum_{k \in D(j)} \delta_k w_{k,j} \right) &\text{otherwise}\\
    \end{cases}
\end{align*}

The $\delta_j$ get calculated layer-wise, starting from the output layer. This
is the reason why this learning algorithm is also called the
\textit{backpropagation algorithm}, although it is only a special case of
gradient descent. The signal gets propagated through the network, the output
is generated and then the error is propagated back.

\subsection{Batch, Mini-Batch and Stochastic Gradient Descent}\label{subsec:batch-minibatch-stochastic-gradient-descent}
Neural Networks can be trained in three different training modes. The
\textit{stochastic gradient descent} takes one training example and adjusts the
weights. Another training mode is \textit{mini-batch gradient descent} where a
chunk of a fixed size $b$, the size of the mini-batch, is used to calculate the
gradient and to adjust the weights. A third training mode is \textit{batch
gradient descent} where all training examples are used to calculate the
adjustment of weights. A common choice for the mini-batch size is $b=256$.
However, for $b=1$ it is stochastic gradient descent and for $b=|T|$ it is batch
gradient descent. The advantage of stochastic gradient descent is that the
weights are updated faster, compared to batch gradient descent. The advantage of
batch gradient descent is that weight updates are more meaningful.
Mini-batch gradient descent can be faster than stochastic gradient descent,
because the weights are updated less often.

\subsection{Momentum}\label{subsec:momentum}

One problem of simple gradient descent is the choice of the learning rate.
Depending on how much the error changes between different epochs, one might
choose a higher learning rate or lower it. A learning parameter called
\textit{momentum} tries to implement such an automatic adjustment of the error.

If one imagines the error surface in the parameter space, one can imagine the
current weight as a ball. The ball begins to roll down the error surface.
If it does not change the direction much and keeps rolling down, it speeds up.
If the direction changes or if the weight increases, the momentum decreases.
It also keeps the ball going in the direction that worked before. So in case of
an error surface that has a plateau, the momentum helps to get away from that plateau.

The momentum $\alpha \in [0, 1]$ changes the weight update to

\[\Delta w_{j,i(\text{epoch}_i)} = \eta \delta_i x_{j,i} + \alpha \Delta w_{j,i}(\text{epoch}_{i-1})\]

as described in~\cite{Mitchell97,bishop2007}.

\subsection{Newbob Training}\label{subsec:newbob-training}

Newbob training is an adaptive training that is described in~\cite{newbob-icis}.
It starts with a learning rate $\eta$ and trains until the error on the validation set
decreases by less than $\theta_1 = \SI{0.5}{\percent}$. When that happens, the learning
rate is multiplied with a decay parameter.
$\text{\texttt{newbob-decay}}=0.5$ is chosen in~\cite{newbob-icis}. The training is
stopped when the error drops by less than
$\theta_2 = \SI{0.5}{\percent}$ after the threshold $\theta_1$ was hit in the
training step before. Those two thresholds can be adjusted, of course.

\subsection{Denoising Auto-encoder}

An \textit{auto-encoder} is a neural network that is trained to restore its
input. This means the number of input neurons is equal to the number of output
neurons. The weights are an \textit{encoding} of the input that allows restoring
the input. As the neural network finds the encoding by itself, it is called
auto-encoder. If the hidden layer is smaller than the input layer, it can be
used for dimensionality reduction~\cite{Hinton1989}. If only one hidden layer with
linear activation functions is used, then the hidden layer contains the
principal components after training~\cite{Duda2001}.

Denoising auto-encoders are a variant introduced in~\cite{Vincent2008} that
is more robust to partial corruption of the input features. It is trained to
get robust by adding noise to the input features.

There are multiple ways how noise can be added. Gaussian noise and
randomly masking elements with zero are two possibilities. \cite{Deeplearning-Denoising-AE}
describes how such a denoising auto-encoder with masking noise can be
implemented. The \texttt{corruption} is the probability of a feature being
masked.

\subsection{Pretraining}\label{subsec:pretraining}
When a neural network gets more layers, the number of weights can decrease even
if the total number of neurons increases. For example, a \gls{MLP} with a
$500:500:500$ topology has
\[500^2 + 500^2 =\num{500000}\]
weights, but a
\gls{MLP} with a $500:100:500:500$ topology has
\[500\cdot 100 + 100 \cdot 500 + 500^2 = \num{350000}\]
weights.

However, the more weights a \gls{MLP} gets, the more random initializations are
done for this \gls{MLP}. This might lead to high variations in classification
performance for the same training queue, but different weight initializations.
One possible way to deal with this problem is to apply pretraining. This means
that the layers are trained before the layers get stacked to form the
resulting model. This means at first, the first hidden layer gets trained. Then the
first two layers get trained, etc.

Pretraining can be done supervised, semi-supervised or unsupervised.
A supervised training algorithms needs labels for all training examples,
an unsupervised does not use any labels and a semi-supervised needs labels for
some examples, but not for all.

Denoising auto-encoders are an example for unsupervised pretraining. \Gls{SLP}
is to train first a \gls{MLP} with one hidden layer, then discard the output
layer, add the second hidden layer and a new output layer and train again.

\subsection{Regularization}

Regularization is a group of methods that help to prevent overfitting, that
means the problem that a model performs much worse on the test set than on the
training set. The idea of regularization in \gls{MLP} is that sparse weights or
weights with a low absolute value tend not to cause overfitting and are
therefore preferred. This can be encoded in the training algorithm by modifying
the cost function such that higher weights correspond with a higher cost when
compared to lower weights that have a similar error on the training set.

Two common regularizations are $L_1$ and $L_2$ regularization. $L_1$
regularization adds the absolute value of the weights to the error function and
$L_2$ regularization adds the squared parameters to the error
function~\cite{Ng2004}.


\chapter{Implementation}\label{ch:Implementation}

When this bachelor's thesis was written, there was no publicly available data
set for on-line handwritten mathematical symbols. In order to get the necessary
data to conduct experiments, the website \url{write-math.com} was created in
preparation for this bachelor's thesis as a free-time project. The code for the
website is available at \url{https://github.com/MartinThoma/write-math}. While
data was gathered, Daniel Kirsch was contacted and asked for the data recorded
by \url{detexify.kirelabs.org}. After some months, he published the data. A link
to the data as well as a description of the data format is available at
\href{http://martin-thoma.com/write-math}{martin-thoma.com/write-math}.

The following sections describe four different projects that were important for
this bachelor's thesis:

\begin{itemize}
    \item \texttt{write-math}: The website that was created to collect recordings
    \item \texttt{hwrt}: The toolkit to view recordings and make experiments
    \item \texttt{hwr-experiments}: The files that define the experiments
    \item Neural Network Training: An internal project for creation, evaluation
          and training of neural networks.
\end{itemize}

\section{write-math.com}
The website \url{http://write-math.com} was created to get data. It is a
combination of PHP, MySQL, JavaScript, CSS and HTML\@. It makes use of the 
front-end framework Bootstrap and the template engine Twig. The source is
available at \url{https://github.com/MartinThoma/write-math}.

The website allows the users to classify recordings (see
\cref{fig:website-classify} on \cpageref{fig:website-classify}). As soon as the
user has drawn the symbol, he clicks on submit and gets redirected to a
classification page. He sees the recording, get a link to a page where he
can try out preprocessing methods and see the symbols that classifiers
suggested. Every user has the possibility to add his own classifier that others
are also able to use. Every time a new recordings gets submitted, the
website contacts every known classifier by sending a JSON string via
POST-request. The website expects every classifier to respond by serving a JSON
string that contains a list of at most 10 dictionaries which map symbol
identifiers (integers) to probabilities. This could look like

\begin{verbatim}
    [{"31":0.88842893496419},
     { "1":0.10999419040225},
     {"36":0.001499575497246},
     {"40":7.7299136313199e-5}]
\end{verbatim}

The list must be ordered descending by probability. \Cref{fig:website-workflow}
visualizes this workflow.

Currently, only System A is online.

\begin{figure}[htb]
    \centering
    \includegraphics*[width=0.7\textwidth,keepaspectratio]{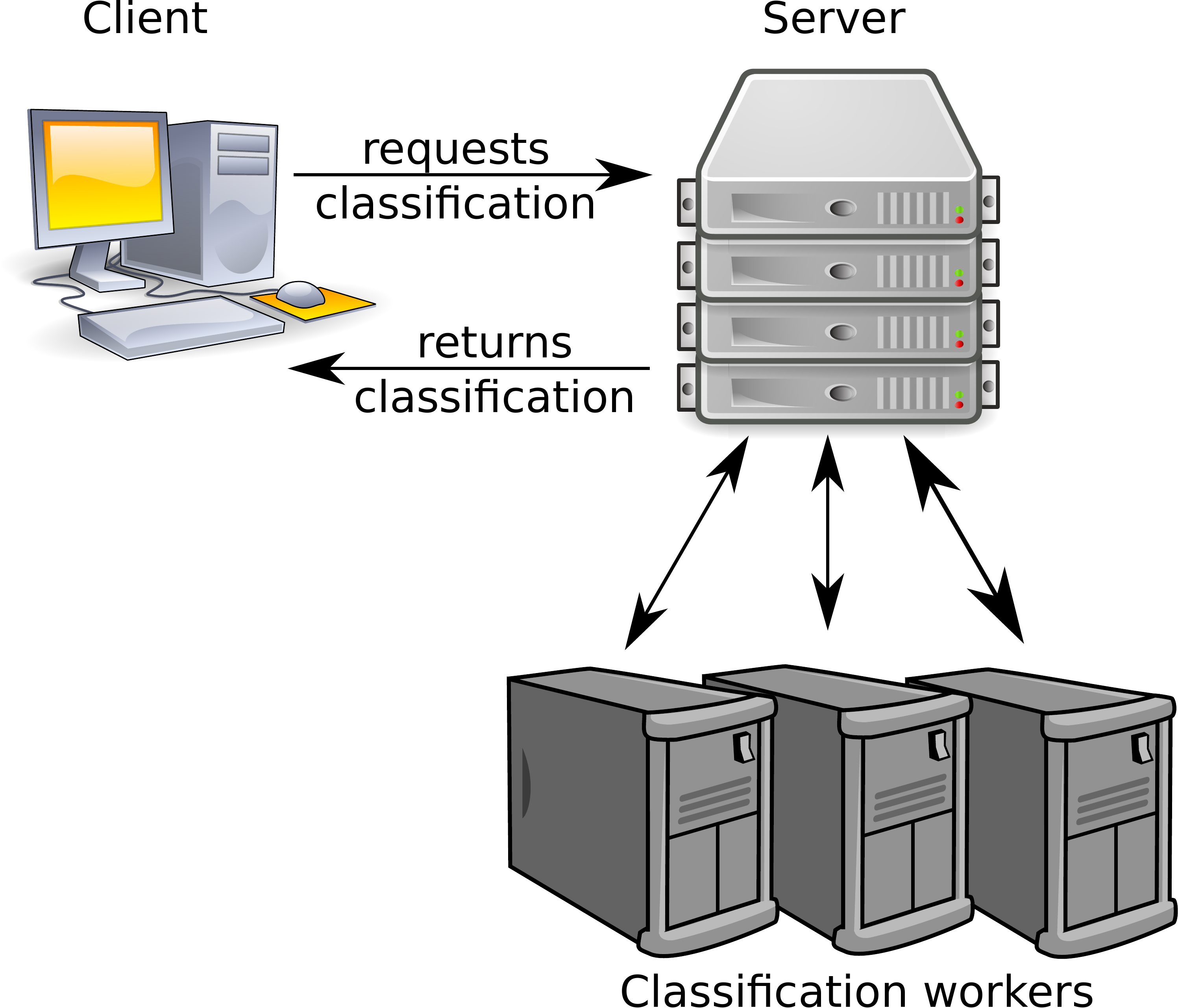}
    \caption{The workflow of a single classification is the following:
             \inlinelist{%
              \protect\item The user writes a symbol. This symbol gets recorded by the
                    users browser via JavaScript and send to the server.
              \protect\item The server stores the recording and contacts all
                    classification workers.
              \protect\item Each classification workers sends a list with up to 10
                    symbols and their probability back to the server.
              \protect\item The server stores those suggestions and shows the user all
                    results.}\\
              The image of a desktop computer on the top left is from
              \url{https://commons.wikimedia.org/wiki/File:Computer-aj_aj_ashton_01.svg}
              and was created by an unknown artist,
              the server image on the top right is from
              \url{https://commons.wikimedia.org/wiki/File:Server-multiple.svg} and was created by RRZEicons and
              the images that was used three times for classification workers is from
              \url{https://commons.wikimedia.org/wiki/File:Server_by_mimooh.svg} and was created by Mimooh.
            }
\label{fig:website-workflow}
\end{figure}

\section{Handwriting Recognition Toolkit}
A toolset was created for the analyzation, preprocessing and feature
calculation of on-line handwritten data. This toolset was bundled in a Python
module called \texttt{hwrt}. It is freely available over the \gls{PyPI} and can
be installed with \texttt{pip install hwrt}. It contains algorithms for
preprocessing, feature selection and data augmentation as well as tools to
download the latest data, view and analyze the data.

The following preprocessing algorithms were implemented. They all work on
exactly one recording. They were described in \cref{sec:preprocessing}.

\begin{itemize}
    \item \texttt{RemoveDuplicateTime}: If a recording has two points with the
          same timestamp, than the second point is discarded. This is
          useful for a couple of algorithms that don't expect two points at
          the same time.
    \item \texttt{RemoveDots}: Remove all strokes that have only a single point
          (a dot) from the recording, except if the whole recording consists of
          dots only.
    \item \texttt{ScaleAndShift}: Scale a recording so that it fits into a unit
          square. This keeps the aspect ratio. Then the recording is shifted.
          The default way is to shift it so that the recording is in
          $[0, 1] \times [0, 1]$. However, it can also be used to be centered
          within $[-1, 1] \times [-1, 1]$ around the origin $(0, 0)$ by setting
          \texttt{center=True} (for the smaller dimension) and
          \verb+center_other=True+ (for the bigger dimension).
    \item \texttt{SpaceEvenly}: Space the points evenly in time over the complete
          recording. The parameter \texttt{number} defines how many points
          should the recording should get in total. All strokes get connected
          by lines. All points on the strokes get a \verb+pen_down=True+
          feature and all points between strokes get a \verb+pen_down=False+
          feature.
    \item \texttt{SpaceEvenlyPerStroke}: Space the points evenly for every
          single stroke separately. The parameter \texttt{number} defines how many
          points are used per stroke and the parameter \texttt{kind} defines which
          kind of interpolation is used. Possible values include \texttt{cubic},
          \texttt{quadratic}, \texttt{linear}, \texttt{nearest}. This part of the
          implementation relies on \texttt{scipy.interpolate.interp1d}.
    \item \texttt{DouglasPeucker}: Apply the Douglas-Peucker stroke
          simplification algorithm separately to each stroke of the recording.
          The algorithm has a threshold parameter \texttt{epsilon} that indicates
          how much the stroke is simplified. The smaller the parameter, the
          closer the resulting strokes are to the original.
    \item \texttt{StrokeConnect}: Detect if strokes were probably accidentally
          disconnected. If that is the case, connect them. This is detected by
          the threshold parameter \verb+minimum_distance+. If the distance
          between the end point of a stroke and the first point of the next
          stroke is below the minimum distance, the strokes are connected.
    \item \texttt{DotReduction}: Reduce strokes where the maximum distance
          between points is below a \texttt{threshold} to a single dot.
    \item \texttt{WildPointFilter}: Find wild points and remove them. The
          threshold means speed in pixels / ms.
    \item \texttt{WeightedAverageSmoothing}: Smooth every stroke by a weighted
          average. This algorithm takes a list \texttt{theta} of 3 numbers that
          are the weights used for smoothing.
\end{itemize}

The following data augmentation algorithms were implemented. They were
described in \cref{sec:Data-augmentation}.

\begin{itemize}
    \item \texttt{Multiply}: Copy the data $n$ times.
    \item \texttt{Rotate}: Adds rotational variants of the recording. It has three
          parameters: \texttt{min}, \texttt{max} and \texttt{num}. The
          algorithm adds \texttt{num} rotated variants of the recording to
          the dataset.
\end{itemize}

The following features were implemented. They were described in
\cref{sec:features}.

\begin{itemize}
    \item \texttt{ConstantPointCoordinates}: Take the first
          \verb+points_per_stroke=20+ points coordinates of the first
          \texttt{strokes=4} strokes as features. This leads to
          $2 \cdot \text{points\_per\_stroke} \cdot \text{strokes}$ features.

          If \texttt{points} is set to $0$, the first \verb+points_per_stroke+
          point coordinates and the \verb+pen_down+ feature is used. This leads
          to $3 \cdot \text{points\_per\_stroke}$ features.

          If there are not enough points or strokes, the feature gets filled
          with \verb+fill_empty_with=0+.
    \item \texttt{FirstNPoints}: Similar to the
          \texttt{ConstantPointCoordinates} feature, this feature takes the first
          \texttt{n=81} point coordinates. It also has the
          \verb+fill_empty_with=0+ to make sure that the dimension of this
          feature is always the same.
    \item \texttt{StrokeCount}: The number of used strokes can be a powerful
          feature. \Cref{fig:stroke-count-mean-std-deviation}
          (\cpageref{fig:stroke-count-mean-std-deviation}) gives an impression
          how good this feature can separate some symbols.
    \item \texttt{Bitmap}: $n \times n$ grayscale bitmap or the recording, where
          \texttt{n} is a parameter. A human can recognize most recordings with
          $n=32$ and still many with $n=18$.
    \item \texttt{Ink}: Ink as a 1-dimensional feature. It gives a numeric value
          for the amount of ink this would eventually have consumed.
    \item \texttt{AspectRatio}: Aspect ratio
          ($\frac{\text{width}+0.01}{\text{height}+0.01}$) of a recording as a
          1-dimensional feature.
    \item \texttt{Width}: Width of a recording as a 1-dimensional feature.\\
          Note that this is the current width. So if the recording was scaled,
          this will not be the original width.
    \item \texttt{Height}: Height of a recording as a 1-dimensional feature.\\
          Note that this is the current height. So if the recording was scaled,
          this will not be the original height.
    \item \texttt{Time}: The time in milliseconds it took to create the recording.
          This is a 1-dimensional feature.
    \item \texttt{CenterOfMass}: Center of mass of a recording as a 2-dimensional
          feature.
    \item \texttt{StrokeCenter}: Get the stroke center of mass coordinates for
          the first \texttt{stroke=4} strokes. The dimension of this feature
          is $2 \cdot \text{stroke}$.
    \item \texttt{StrokeIntersections}: Count the number of intersections which
          strokes in the recording have with each other in form of a symmetrical
          matrix for the first \texttt{stroke=4} strokes.
          The feature dimension is
          $\Call{round}{\frac{\text{strokes}^2}{2}} + \frac{\text{strokes}}{2}$,
          because the symmetrical part is discarded.
    \item \texttt{ReCurvature}: Re-curvature is a 1-dimensional, stroke-global
          feature for a recording. It is the ratio
          $\frac{\text{height}(stroke)}{\text{length}(stroke)}$.
\end{itemize}

\section{Experiments}
All experiments are saved as configuration files on
\url{https://github.com/MartinThoma/hwr-experiments}. The \gls{HWRT} is able
to use those configuration files and regenerate the models automatically. The
structure of the configuration files is explained in
\cref{appendix:hwrt-handbook}.

\section{Neural Network Implementation}
The training and testing of neural networks with \texttt{hwrt} needs an
executable \texttt{nntoolkit} that supports the following usages:

\begin{verbatim}
    $ nntoolkit run --batch-size 1 -f%0.4f <test_file> < <model>
\end{verbatim}
has to output the evaluation result in standard output as a list of floats
separated by newlines \verb+\n+. The evaluation result might either be the
index of the neuron with highest activation or the list of probabilities
of each class separated by spaces.

\begin{verbatim}
    $ nntoolkit make mlp <topology>
\end{verbatim}
has to print the model in standard output.

The \texttt{hwrt} toolset is independent of the way the training command is
formatted as the training command gets inserted directly into the configuration
file \texttt{info.yml} of the model.

In order to implement such a neural network executable one can use Theano, cuDNN
(\url{https://developer.nvidia.com/cuDNN}) or Caffe
(\url{http://caffe.berkeleyvision.org/}).
\url{http://www.deeplearning.net/tutorial/} contains example code for multilayer
perceptrons written with Theano (Python).


\chapter{Evaluation}\label{ch:Evaluation}
The following experiments and their results show how the previously described
algorithms perform and how they influence the classification error on the test
set. The training set has $\num{\trainingsetsize}$ recordings, the validation
set has $\num{\validtionsetsize}$ recordings and the test set has
$\num{\testsetsize}$ recordings. \totalClassesAnalyzed{} symbols were
tested. Those symbols are listed in
\crefrange{table:symbols-used-for-evaluation-0}{table:symbols-used-for-evaluation-8}.

All changes that are described in the following were done with $4$ systems.
All of those $4$ systems use a simple preprocessing queue: Scaling with
respect to the aspect ratio to fit into a unit square, shifting to
$[-1, 1] \times [-1, 1]$ and linear resampling. The first $4$ strokes of a
recording were used for features, all other strokes were discarded. For each
stroke, $20$ points coordinates that were spread equidistant in time were taken
as features. If a recording had less then $4$ strokes, the feature got $0$ as a
value. Hence the trained neural networks gets
$4 \text{ strokes } \cdot 20 \frac{\text{points}}{\text{stroke}} \cdot 2 \frac{\text{features}}{\text{point}} = 160$
input features which equals the number of input neurons.

System $B_i$ has $i$ hidden layers with $500$ neurons per hidden layer.
Mini-batch training with a batch size of $256$, a learning rate of $\eta = 0.1$
and a momentum of $\alpha = 0.1$ was used. Every system $B_i$ has a softmax
layer at the end. Neither regularization nor pretraining were used. As different
topologies might severely influence the classification results of \glspl{MLP},
one baseline system was chosen for each of the 4 tested topologies.

\Cref{table:baseline-systems} shows three types of errors for four
different\label{merged-error-introduction}
\glspl{MLP}: \textit{TOP1}, \textit{TOP3} and \textit{MER}. TOP $n$ is the
standard classification error which tests if the \gls{reference} class was
within the $n$ \glspl{hypothesis} with highest probability. The \gls{MER} error
(short for \textit{merged classes}) accepts the symbols in
\cref{table:difficult-symbols} as being equivalent. MER first gets the TOP3
hypotheses, extends this set $M$ by all equivalent symbols and then checks if
the reference class is within $M$.

\begin{table}[H]
    \centering
    \begin{tabular}{clrrr}
    \toprule
    \multirow{2}{*}{System}  & \multirow{2}{*}{Topology} & \multicolumn{3}{c}{Classification error}\\
    \cmidrule(l){3-5}
          &                         & TOP1                   & TOP3                  & MER \\\midrule
    $B_1$ & 160:500:369             & $\SI{23.34}{\percent}$ & $\SI{6.80}{\percent}$ & $\SI{6.64}{\percent}$ \\
    $B_2$ & 160:500:500:369         & \underline{$\SI{21.51}{\percent}$} & $\SI{5.75}{\percent}$ & $\SI{5.67}{\percent}$ \\
    $B_3$ & 160:500:500:500:369     & $\SI{21.93}{\percent}$ & \underline{$\SI{5.74}{\percent}$} & \underline{$\SI{5.64}{\percent}$} \\
    $B_4$ & 160:500:500:500:500:369 & $\SI{23.88}{\percent}$ & $\SI{6.12}{\percent}$ & $\SI{6.04}{\percent}$ \\
    \bottomrule
    \end{tabular}
    \caption{Evaluation of the baseline systems $B_1$--$B_4$ with three
             different classification error measures. All errors were measured
             on the test set.}
\label{table:baseline-systems}
\end{table}

\section{Influence of Random Weight Initialization}\label{sec:random-weight-initialization}
The neural networks in all experiments got initialized with a small random
weight

\[w \sim U(-4 \cdot \sqrt{\frac{6}{n_j + n_{j+1}}}, 4 \cdot \sqrt{\frac{6}{n_j + n_{j+1}}}) \text { where } w \text{ is a weight between layer } j \text{ and layer } (j+1)\]

as suggested on \cite{deeplearningweights}. The random initialization is done to
break symmetry.

This might lead to different error rates for the same models just because the
initialization was different.

In order to get an impression of the magnitude of the influence on the different
topologies and error rates the baseline models were trained 5 times with
random initializations.
\Cref{table:baseline-systems-random-initializations-summary}
shows a summary of the results and
\cref{table:baseline-systems-random-initializations} shows the raw data. The
more hidden layers were used, the more have the results varied.

\begin{table}[h]
    \centering
    \begin{tabular}{crrr|rrr|rrr} 
    \toprule
    \multirow{3}{*}{System}  & \multicolumn{9}{c}{Classification error}\\
    \cmidrule(l){2-10}
          & \multicolumn{3}{c}{TOP1}   & \multicolumn{3}{c}{TOP3}& \multicolumn{3}{c}{MER} \\
          & min     & max     & range  & min    & max    & range & min    & max    & range  \\\midrule
    $B_1$ & $\SI{23.08}{\percent}$ & $\SI{23.44}{\percent}$ & $\SI{0.36}{\percent}$ & $\SI{6.67}{\percent}$ & $\SI{6.80}{\percent}$ & $\SI{0.13}{\percent}$ & $\SI{6.54}{\percent}$ & $\SI{6.64}{\percent}$ & $\SI{0.10}{\percent}$ \\
    $B_2$ & \underline{$\SI{21.45}{\percent}$} & \underline{$\SI{21.83}{\percent}$} & $\SI{0.38}{\percent}$ & $\SI{5.68}{\percent}$ & \underline{$\SI{5.75}{\percent}$} & $\SI{0.07}{\percent}$ & $\SI{5.60}{\percent}$ & \underline{$\SI{5.68}{\percent}$} & $\SI{0.08}{\percent}$ \\
    $B_3$ & $\SI{21.54}{\percent}$ & $\SI{22.28}{\percent}$ & $\SI{0.74}{\percent}$ & \underline{$\SI{5.50}{\percent}$} & $\SI{5.82}{\percent}$ & $\SI{0.32}{\percent}$ & \underline{$\SI{5.41}{\percent}$} & $\SI{5.75}{\percent}$ & $\SI{0.34}{\percent}$ \\
    $B_4$ & $\SI{23.19}{\percent}$ & $\SI{24.84}{\percent}$ & $\SI{1.65}{\percent}$ & $\SI{5.98}{\percent}$ & $\SI{6.44}{\percent}$ & $\SI{0.46}{\percent}$ & $\SI{5.83}{\percent}$ & $\SI{6.21}{\percent}$ & $\SI{0.38}{\percent}$ \\
    \bottomrule
    \end{tabular}
    \caption{The systems $B_1$ -- $B_4$ were randomly initialized, trained
             and evaluated 5~times to estimate the influence of random weight
             initialization.}
\label{table:baseline-systems-random-initializations-summary}
\end{table}

\section{Preprocessing Algorithms}
The preprocessing algorithms can be
split in two groups as shown in \cref{sec:preprocessing}: Normalization and
noise reduction algorithms.

Both, normalization and noise reduction algorithms, can be analyzed for
computational costs and effect on the test classification error. Additionally,
noise reduction algorithms can be analyzed for effectiveness in terms of false
positives or false negatives. However, in the following they were only analyzed
for their effect on the three error measures TOP1, TOP3 and MER\@. 

\subsection{Scale and Shift}\label{scale-and-shift-implementations}
There are several ways to implement the scale and shift algorithm. Especially
how one deals with dots or straight lines ($x_{\max}-x_{\min} = 0$ or
$y_{\max} - y_{\min} = 0$) makes a difference.

The following transformation is done with each point:

\begin{align*}
    p['x'] &\gets (p['x'] - x_{\min}) \cdot factor - add_x\\
    p['y'] &\gets (p['y'] - y_{\min}) \cdot factor - add_y\\
    p['t'] &\gets (p['t'] - t_{\min})
\end{align*}

where $x_{\min}$, $y_{\min}$ and $t_{\min}$ are the minimal values among all
points of a single recording, $factor \in \mathbb{R}^+$ is positive scaling
constant and $add_x, add_y \in \mathbb{R}_0^+$ are non-negative shifting
constants.

\textbf{Implementation~1} is the implementation that was used for all other
evaluations. It does not shift the bigger dimension, but centers the smaller
dimension of the bounding box within the $[-1, 1] \times [-1, 1]$ unit square.

A recording with a bounding box of the dimension $1 \times 0.8$ would be within
${[-0.4, 0.4]} \times {[0.0, 1.0]}$ after implementation~1 shifting.

\paragraph{The following lines show how such a shifting could be implemented:}
\begin{algorithmic}
    \State$width \gets x_{\max} - x_{\min}$
    \State$height \gets y_{\max} - y_{\min}$
    \State$factor_x, factor_y \gets 1, 1$
    \If{$width \neq 0$}
        \State$factor_x \gets \frac{1}{width}$
    \EndIf%
    \If{$height \neq 0$}
        \State$factor_y \gets \frac{1}{height}$
    \EndIf%
    \State$factor \gets \Call{min}{factor_x, factor_y}$
    \State$add_x, add_y \gets 0, 0$
    \LineComment{Only the smaller dimension ($x$ or $y$) gets centered}
    \State$add \gets - \frac{\Call{min}{width, height} \cdot factor}{2}$

    \If{$factor == factor_x$}
        \State$add_y \gets add$
    \Else%
        \State$add_x \gets add$
    \EndIf%
\end{algorithmic}

\textbf{Implementation~2} is the same as implementation~1, but with $add_x = 0$
and $add_y = 0$. So no centering was done. After that, the recording is in the
$[0,1] \times [0,1]$ unit square, aligned to $(0, 0)$.

A recording with a bounding box of the dimension $1 \times 0.8$ would be within
${[0.0, 0.8]} \times {[0.0, 1.0]}$ after implementation~2 shifting.

\textbf{Implementation~3} is the same as implementation~1, but with the bigger
dimension being shifted by $-0.5$. So in implementation~1, only one dimension
gets centered around $(0,0)$. In implementation~3, both dimensions get centered
around $(0, 0)$.

\begin{algorithmic}
    \If{$factor == factor_x$}
        \State$add_y \gets add$
        \State$add_x \gets -0.5$
    \Else%
        \State$add_x \gets add$
        \State$add_y \gets -0.5$
    \EndIf%
\end{algorithmic}

A recording with a bounding box of the dimension $1 \times 0.8$ would be within
${[-0.4, 0.4]} \times {[-0.5, 0.5]}$ after implementation~3 shifting.

Those three implementations of the scale and shift algorithm were tested with all
the neural networks $B_1$--$B_4$. The results in
\cref{table:scale-and-shift-variations} show that system $B_4$ was most
sensitive for changes in this implementation. Implementation~2 performed best or
was at lest not more than $\SI{0.04}{\percent}$ worse than implementation~1 for
$B_1$--$B_3$. However, implementation~2 was by far the worst for $B_4$. The
experiment was executed four times with different weight initializations for
$B_{4,I2}$ and all evaluations were at least $\SI{3}{\percent}$ worse in TOP1
error than $B_{4}$.

\subsection{Wild Point Filter}
Wild points are strokes which consist of a single point which the user did
not want to draw. Wild points are likely to be caused by hardware errors (see
\cref{itm:d1-wild-points},
\cpageref{itm:d1-wild-points}).

The dataset contained \recordingsWithDots{} recordings with dots, excluding all
recordings of the symbols i, j, \verb+\cdot+, \verb+\div+,
\verb+\because+ and \verb+\therefore+. However, removing those dots changed
the bounding box size of only \recordingsWithDotsSizechange{} of all recordings.

As the proposed wild point detection relies only on the speed of single points
of a stroke it was analyzed in which range those points are. The mean speed was
$\SI{0.35}{\pixel\per\milli\second}$ with a standard deviation of
$\SI{0.65}{\pixel\per\milli\second}$. \Cref{fig:instroke-point-speed} shows the
distribution of the speed between control points in a histogram. After that, the
wild point filter with a threshold $\theta=\SI{3}{\pixel\per\milli\second}$
and $\theta=\SI{6}{\pixel\per\milli\second}$ and were tested.
The results are listed in \cref{table:instroke-point-speed}. The models
$B_3$ and $B_4$ improved by both applications, whereas the models $B_1$ and
$B_2$ did not improve.

\begin{figure}[h!]
    \centering

\newcommand\clipright[1][white]{
  \fill[#1](current axis.south east)rectangle(current axis.north-|current axis.outer east);
  \pgfresetboundingbox%
  \useasboundingbox(current axis.outer south west)rectangle([xshift=.5ex]current axis.outer north-|current axis.east);
}

\definecolor{mycolor}{rgb}{0.02,0.4,0.7}

\begin{tikzpicture}
    \begin{axis}[
        ymajorgrids,
        xmajorgrids,
        grid style={white,thick},
        axis on top,
        /tikz/ybar interval,
        tick align=outside,
        ymin=0,
        axis line style={draw opacity=0},
        tick style={draw=none},
        enlarge x limits=false,
        height=7cm,
        title style={font=\Large},
        xlabel={Speed between two subsequent points of a stroke in $\si{\pixel\per\milli\second}$},
        ylabel={Number of point pairs},
        ytick={ 0,2000000,4000000,6000000,8000000,10000000,12000000,14000000,16000000,18000000 },
        scaled ticks=false,
        yticklabels={ 0, 2M, 4M, 6M, 8M, 10M, 12M, 14M, 16M, 18M },
        xticklabels={ $0.0$, $0.5$, $1.0$, $1.5$, $2.0$, $2.5$, $3.0$, $3.5$, $4.0$, $4.5$, $\infty$ },
        width=\textwidth,
        xtick=data,
        label style={font=\large},
        ticklabel style={
            inner sep=1pt,
            font=\small
        },
        nodes near coords,
        every node near coord/.append style={
            fill=white,
            anchor=mid west,
            shift={(3pt,4pt)},
            inner sep=0,
            font=\footnotesize,
            rotate=45},
            ]
    \addplot[mycolor!80!white, fill=mycolor, draw=none] coordinates { (0, 19638952) (1, 3382079) (2, 1089716) (3, 288877) (4, 154611) (5, 53197) (6, 37141) (7, 15664) (8, 14529) (9, 5046) (10, 25799)  };
    \end{axis}
    \clipright
\end{tikzpicture}
    \caption{The speed between two subsequent control points of the same stroke
             is used can be used for point filtering preprocessing steps. Points
             with high speeds could be caused by errors in the hardware. The
             plot shows that the majority of all point pairs have a speed of
             less than $\SI{0.5}{\pixel\per\milli\second}$. Less than
             $\SI{0.3}{\percent}$ of all point pairs have a speed of more than
             $\SI{3}{\pixel\per\milli\second}$.}
\label{fig:instroke-point-speed}
\end{figure}
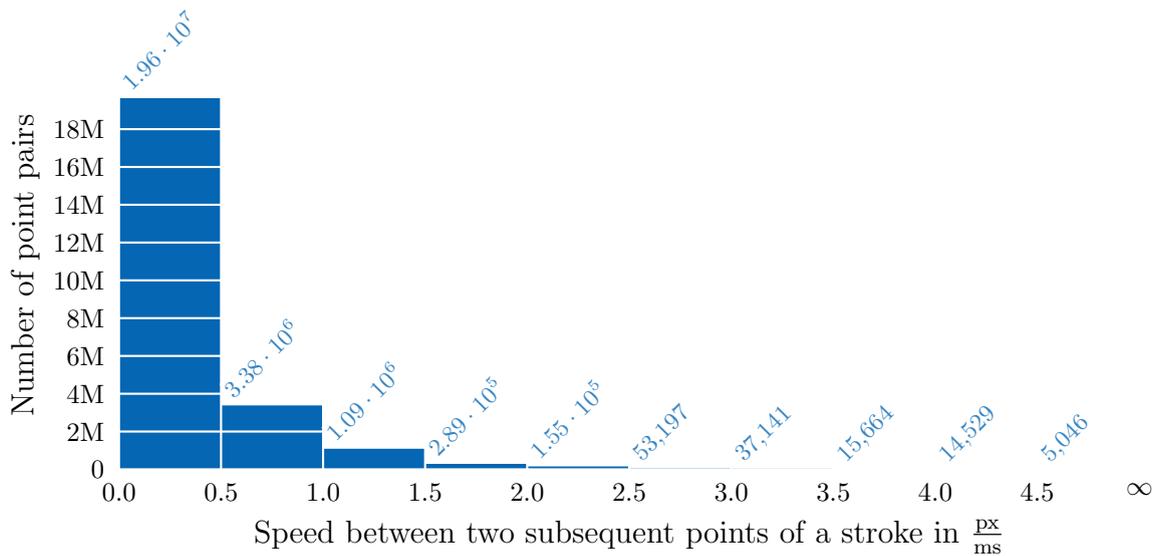

\begin{table}[h]
    \centering
    \begin{tabular}{lrrrrrrr}
    \toprule
    \multirow{2}{*}{System} & \multicolumn{6}{c}{Classification error}\\
    \cmidrule(l){2-7}
                                                       & TOP1                   & change                 & TOP3                   & change                 & MER                 & change \\\midrule
    $B_{1,\theta_w = 3\si{\pixel\per\milli\second}}$   & $\SI{23.55}{\percent}$ & $\SI{+0.21}{\percent}$ &  $\SI{6.85}{\percent}$ & $\SI{+0.05}{\percent}$ &  $\SI{6.70}{\percent}$ & $\SI{+0.06}{\percent}$ \\
    $B_{2,\theta_w = 3\si{\pixel\per\milli\second}}$   & $\SI{21.73}{\percent}$ & $\SI{+0.22}{\percent}$ &  $\SI{5.67}{\percent}$ & $\SI{-0.08}{\percent}$ &  $\SI{5.58}{\percent}$ & $\SI{-0.09}{\percent}$ \\
    $B_{3,\theta_w = 3\si{\pixel\per\milli\second}}$   & \underline{$\SI{21.25}{\percent}$} & $\SI{-0.68}{\percent}$ &  \underline{$\SI{5.66}{\percent}$} & $\SI{-0.08}{\percent}$ & \underline{$\SI{5.55}{\percent}$} & $\SI{-0.09}{\percent}$ \\
    $B_{4,\theta_w = 3\si{\pixel\per\milli\second}}$   & $\SI{21.91}{\percent}$ & $\SI{-1.97}{\percent}$ &  $\SI{5.77}{\percent}$ & $\SI{-0.35}{\percent}$ &  $\SI{5.65}{\percent}$ & $\SI{-0.39}{\percent}$ \\\midrule
    $B_{1,\theta_w = 6\si{\pixel\per\milli\second}}$   & $\SI{23.30}{\percent}$ & $\SI{-0.04}{\percent}$ &  $\SI{6.94}{\percent}$ & $\SI{+0.14}{\percent}$ &  $\SI{6.80}{\percent}$ & $\SI{+0.16}{\percent}$ \\
    $B_{2,\theta_w = 6\si{\pixel\per\milli\second}}$   & $\SI{21.80}{\percent}$ & $\SI{+0.29}{\percent}$ &  $\SI{5.77}{\percent}$ & $\SI{+0.02}{\percent}$ &  $\SI{5.65}{\percent}$ & $\SI{-0.02}{\percent}$ \\
    $B_{3,\theta_w = 6\si{\pixel\per\milli\second}}$   & $\SI{22.30}{\percent}$ & $\SI{-0.63}{\percent}$ &  $\SI{5.79}{\percent}$ & $\SI{+0.05}{\percent}$ &  $\SI{5.58}{\percent}$ & $\SI{-0.06}{\percent}$ \\
    $B_{4,\theta_w = 6\si{\pixel\per\milli\second}}$   & $\SI{22.98}{\percent}$ & $\SI{-0.90}{\percent}$ &  $\SI{6.10}{\percent}$ & $\SI{-0.02}{\percent}$ &  $\SI{5.99}{\percent}$ & $\SI{-0.05}{\percent}$ \\
    \bottomrule
    \end{tabular}
    \caption{The wild point filtering algorithm was added to the baseline
             systems $B_1$--$B_4$ and the absolute change to the baseline model
             of the same topology was calculated. The threshold $\theta_w$,
             which is measured in $\si{\pixel\per\milli\second}$, denotes which
             points get filtered. All points that are faster than the threshold
             get filtered. This means a threshold
             $\theta = \SI{0}{\pixel\per\milli\second}$ results in the same as
             if no wildpoint filter had been applied. The table shows that the
             effect of wild point filtering is less than the effect of random
             weight initialization for the models $B_1$--$B_3$, but improves
             model $B_4$.}
\label{table:instroke-point-speed}
\end{table}

\subsection{Stroke Connect}\label{subsec:stroke-connect}
In order to solve \cref{itm:d5-interrupted-stroke} (interrupted strokes, see
\cpageref{itm:d5-interrupted-stroke}) the stroke connect algorithm was
introduced on \cpageref{alg:stroke-connect-preprocessing}. The idea is that for
a pair of consecutively drawn strokes $s_{i}, s_{i+1}$ the last point $s_i$ is
close to the first point of $s_{i+1}$ if a stroke was accidentally split
into two strokes.

\Cref{fig:histogram-distance-between-strokes} shows the distance between
consecutively drawn stroke pairs. $\SI{59}{\percent}$ of all stroke pair
distances are between $\SI{30}{\pixel}$ and $\SI{150}{\pixel}$. Hence the stroke
connect algorithm was tried with $\SI{5}{\pixel}$, $\SI{10}{\pixel}$ and
$\SI{20}{\pixel}$.
\Cref{table:stroke-connect-evaluation} shows the results of this algorithm.
All models improved much with a threshold of $\theta = \SI{10}{\pixel}$ with
all error measures, except $B_4$ with the TOP3 error measure.

\begin{figure}[h!]
    \centering
    \newcommand\clipright[1][white]{
  \fill[#1](current axis.south east)rectangle(current axis.north-|current axis.outer east);
  \pgfresetboundingbox%
  \useasboundingbox(current axis.outer south west)rectangle([xshift=.5ex]current axis.outer north-|current axis.east);
}

\definecolor{mycolor}{rgb}{0.02,0.4,0.7}

\begin{tikzpicture}
    \begin{axis}[
        ymajorgrids,
        xmajorgrids,
        grid style={white,thick},
        axis on top,
        /tikz/ybar interval,
        tick align=outside,
        ymin=0,
        axis line style={draw opacity=0},
        tick style={draw=none},
        enlarge x limits=false,
        height=7cm,
        title style={font=\Large},
        xlabel={Distance between subsequently drawn strokes in $\si{\pixel}$},
        ylabel={Number of stroke pairs},
        ytick={ 0,4000,8000,12000,16000,20000 },
        scaled ticks=false,
        yticklabels={ 0, 4K, 8K, 12K, 16K, 20K },
        xticklabels={ $0.0$, $30$, $60$, $90$, $120$, $150$, $180$, $210$, $240$, $270$, $\infty$ },
        width=\textwidth,
        xtick=data,
        label style={font=\large},
        ticklabel style={
            inner sep=1pt,
            font=\small
        },
        nodes near coords,
        every node near coord/.append style={
            fill=white,
            anchor=mid west,
            shift={(3pt,4pt)},
            inner sep=0,
            font=\footnotesize,
            rotate=45},
            ]
    \addplot[mycolor!80!white, fill=mycolor, draw=none] coordinates { (0, 15629) (1, 20643) (2, 23400) (3, 22757) (4, 20616) (5, 16659) (6, 11988) (7, 7960) (8, 4853) (9, 2690) (10, 830)  };
    \end{axis}
    \clipright
\end{tikzpicture}
    \caption{The distance between two subsequently drawn strokes
             $(s_{i}, s_{i+1})$ in pixels is calculated by measuring the euclidean
             distance between the last point of $s_{i}$ and the first point of
             $s_{i+1}$. Less than $\SI{11}{\percent}$ of those distances are
             below $\SI{30}{\pixel}$. It is assumed that accidentally
             interrupted strokes (see \cpageref{itm:d5-interrupted-stroke})
             are rarely happening. The stroke connect algorithm is therefore
             evaluated with thresholds of less than $\SI{30}{\pixel}$.}
\label{fig:histogram-distance-between-strokes}
\end{figure}
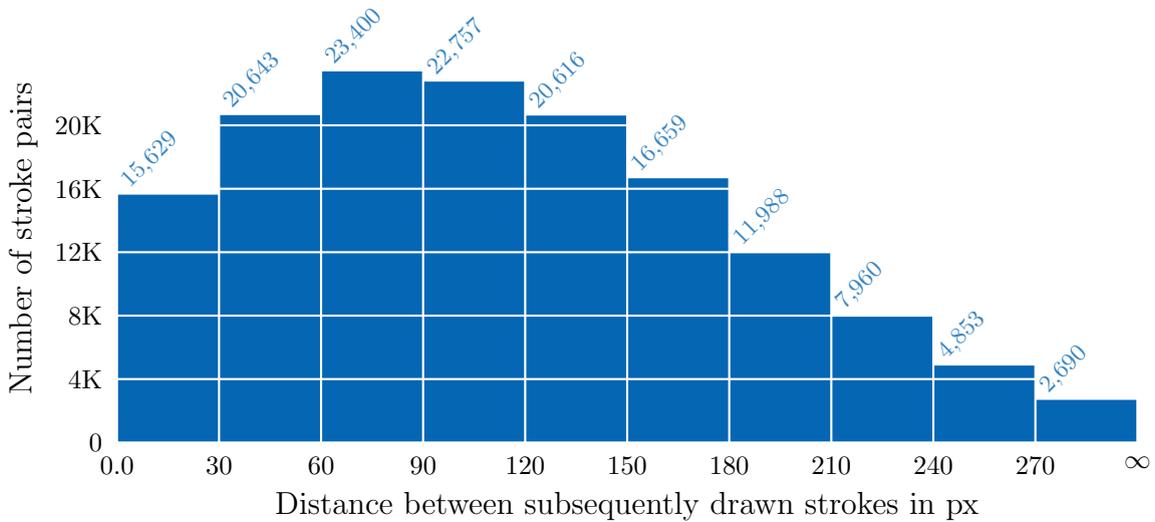

\clearpage

\subsection{Weighted Average Smoothing}

Weighted average smoothing was described in
\cref{preprocessing:weighted-average-smoothing} on
\cpageref{preprocessing:weighted-average-smoothing}. It takes consecutive
points, weights the $x$, $y$ and $time$ values independently and calculates a
new average point. Three points were used to calculate the new average point
with weights $w_1 = [\frac{1}{6}, \frac{4}{6},
\frac{1}{6}]$ and $w_2 = [\frac{1}{3}, \frac{1}{3}, \frac{1}{3}]$. The results
are shown in
\cref{table:weighted-average-smoothing}. The results with $w_1$ did not change
enough to make a meaningful statement about the influence of this algorithm, but
$w_2$ had a positive effect on $B_1$ -- $B_3$. 

\begin{table}[h]
    \centering
    \begin{tabular}{llrrrrrr}
    \toprule
    \multirow{2}{*}{System}          & \multirow{2}{*}{Weights} & \multicolumn{6}{c}{Classification error}\\
    \cmidrule(l){3-8}
               &                                         & TOP1                   & change                 & TOP3                   & change                 & MER                 & change  \\\midrule
    $B_{1,WAS}$& $\frac{1}{6}, \frac{4}{6}, \frac{1}{6}$ & $\SI{23.33}{\percent}$ & $\SI{-0.01}{\percent}$ &  $\SI{6.68}{\percent}$ & $\SI{-0.12}{\percent}$ &  $\SI{6.57}{\percent}$ & $\SI{-0.07}{\percent}$ \\
    $B_{2,WAS}$& $\frac{1}{6}, \frac{4}{6}, \frac{1}{6}$ & $\SI{21.73}{\percent}$ & $\SI{+0.22}{\percent}$ &  $\SI{5.87}{\percent}$ & $\SI{+0.12}{\percent}$ &  $\SI{5.75}{\percent}$ & $\SI{+0.08}{\percent}$ \\
    $B_{3,WAS}$& $\frac{1}{6}, \frac{4}{6}, \frac{1}{6}$ & $\SI{21.77}{\percent}$ & $\SI{-0.16}{\percent}$ &  \underline{$\SI{5.57}{\percent}$} & $\SI{-0.17}{\percent}$ & \underline{$\SI{5.52}{\percent}$} & $\SI{-0.12}{\percent}$ \\
    $B_{4,WAS}$& $\frac{1}{6}, \frac{4}{6}, \frac{1}{6}$ & $\SI{24.17}{\percent}$ & $\SI{+0.29}{\percent}$ &  $\SI{6.47}{\percent}$ & $\SI{+0.35}{\percent}$ &  $\SI{6.21}{\percent}$ & $\SI{+0.17}{\percent}$ \\\midrule
    $B_{1,WAS}$& $\frac{1}{3}, \frac{1}{3}, \frac{1}{3}$ & $\SI{23.26}{\percent}$ & $\SI{-0.08}{\percent}$ &  $\SI{6.55}{\percent}$ & $\SI{-0.25}{\percent}$ &  $\SI{6.41}{\percent}$ & $\SI{-0.23}{\percent}$ \\
    $B_{2,WAS}$& $\frac{1}{3}, \frac{1}{3}, \frac{1}{3}$ & $\SI{21.67}{\percent}$ & $\SI{-0.16}{\percent}$ &  $\SI{5.69}{\percent}$ & $\SI{-0.06}{\percent}$ &  $\SI{5.60}{\percent}$ & $\SI{-0.07}{\percent}$ \\
    $B_{3,WAS}$& $\frac{1}{3}, \frac{1}{3}, \frac{1}{3}$ & \underline{$\SI{21.44}{\percent}$} & $\SI{-0.49}{\percent}$ &  $\SI{5.67}{\percent}$ & $\SI{-0.07}{\percent}$ &  $\SI{5.58}{\percent}$ & $\SI{-0.06}{\percent}$ \\
    $B_{4,WAS}$& $\frac{1}{3}, \frac{1}{3}, \frac{1}{3}$ & $\SI{24.24}{\percent}$ & $\SI{+0.36}{\percent}$ &  $\SI{6.26}{\percent}$ & $\SI{+0.14}{\percent}$ &  $\SI{5.95}{\percent}$ & $\SI{-0.09}{\percent}$ \\
    \bottomrule
    \end{tabular}
    \caption{The baseline models $B_1$--$B_4$ were tested with additionally
             weighted average smoothing (WAS) being applied. The smoothing was applied
             before every other preprocessing step.}
\label{table:weighted-average-smoothing}
\end{table}

\subsection{Douglas-Peucker Smoothing}\label{subsec:douglas-peucker-smoothing-evaluation}

The Douglas-Peucker algorithm, which is described on
\cpageref{preprocessing:douglas-peucker-smoothing}, can be used to find control
points that are more relevant for the overall shape of a recording. After that,
an interpolation can be done. If the interpolation is a cubic spline
interpolation, this makes the recording smooth.

The Douglas-Peucker algorithm was applied with a threshold of
$\varepsilon = 0.05$, $\varepsilon = 0.1$ and $\varepsilon = 0.2$ after scaling
and shifting, but before the interpolation. The interpolation was done linearly
and with cubic splines in two experiments. The recording was scaled and shifted
again after the interpolation because the bounding box might have changed.

The result of the application of the Douglas-Peucker smoothing with $\varepsilon
> 0.05$ was a high rise of all classification error measures for all models.
This means that the simplification process removes some relevant information and
does not --- as it was expected --- remove only noise. For $\varepsilon = 0.05$
with linear interpolation some models improved for some error measures, but the
changes were small. It could be an effect of random weight initialization.
However, cubic spline interpolation made all systems perform much worse.

The lower the value of $\varepsilon$, the less does the recording change after
this preprocessing step. As it was applied after scaling the recording such that
the biggest dimension of the recording (width or height) is $1$, a value of
$\varepsilon = 0.05$ means that a point has to move at least $\SI{5}{percent}$
of the biggest dimension.

\Cref{table:douglas-peucker-smoothing-evaluation} shows the evaluation results.
\clearpage

\section{Data Augmentation}

Data augmentation can be used to make the model invariant to transformations.
However, this idea seems not to work well in the domain of on-line handwritten
mathematical symbols. It was tried to triple the data by adding a rotated
version that is rotated 3 degrees to the left and another one that is rotated
3 degrees to the right around the center of mass. This data augmentation
made all classifiers for most error measures perform worse than before as
\cref{table:data-augmentation-evaluation} shows.

Data augmentation was also used in \cref{subsec:newbob-evaluation} on
\cpageref{subsec:newbob-evaluation} combined with newbob training.

\begin{table}[htb]
    \centering
    \begin{tabular}{lrrrrrrr}
    \toprule
    \multirow{2}{*}{System}  & \multicolumn{6}{c}{Classification error}\\
    \cmidrule(l){2-7}
                               & TOP1                   & change                  & TOP3                   & change                 & MER                & change \\\midrule
    $B_{1,min=-3,max=3,num=3}$ & $\SI{25.43}{\percent}$ & $\SI{ +2.09}{\percent}$ &  $\SI{6.44}{\percent}$ & $\SI{-0.36}{\percent}$ & $\SI{6.28}{\percent}$ & $\SI{-0.36}{\percent}$ \\
    $B_{2,min=-3,max=3,num=3}$ & \underline{$\SI{23.58}{\percent}$} & $\SI{ +2.07}{\percent}$ &  \underline{$\SI{5.78}{\percent}$} & $\SI{+0.03}{\percent}$ & \underline{$\SI{5.57}{\percent}$} & $\SI{-0.10}{\percent}$ \\
    $B_{3,min=-3,max=3,num=3}$ & $\SI{28.90}{\percent}$ & $\SI{ +6.97}{\percent}$ &  $\SI{8.39}{\percent}$ & $\SI{+2.65}{\percent}$ & $\SI{7.54}{\percent}$ & $\SI{+1.90}{\percent}$ \\
    $B_{4,min=-3,max=3,num=3}$ & $\SI{39.71}{\percent}$ & $\SI{+15.83}{\percent}$ & $\SI{18.12}{\percent}$ &$\SI{+12.00}{\percent}$ &$\SI{15.20}{\percent}$ & $\SI{+9.16}{\percent}$\\
    \bottomrule
    \end{tabular}
    \caption{Evaluation of the baseline models that used a training set $T'$ which
             was three times bigger than the normal training set $T$. $T'$ was
             created from $T$ by adding two rotational variants for each original
             recording. Those two rotational variants were rotated
             by $\SI{-3}{\degree}$ and by $\SI{+3}{\degree}$ around the center
             of mass.}
\label{table:data-augmentation-evaluation}
\end{table}

\section{Features}\label{sec:features-evaluation}

A single dimension of a feature $F$ of a given symbol $S$ could be modeled by a
random variable. For a random variable $X_S$ that is normally distributed and
has a mean of $\mu$ for the feature $F$ and a standard deviation of $\sigma$ one
writes:

\[X_{S, F} \sim \mathcal{N}(\mu, \sigma^2)\]

In the interval $(\mu_S-2\sigma_S, \mu+2\sigma)$ is about $\SI{99.7}{\percent}$
of the data. That means if those intervals are disjunct for two given symbols,
the symbols can be separated well by the feature $F$. This knowledge can be used
to calculate the mean $\mu$ and the standard deviation $\sigma$ of every symbol
for a given feature. The symbol can then be plotted in a mean-standard deviation
scatter plot at the coordinates $(\mu, \sigma)$. Ideally, the intra-symbol
standard deviation would be low, the inter-symbol standard deviation would be
high and the means of the symbols would be well separated from each other.

For example, in \cref{fig:re-curvature-mean-std-deviation} one can see that
the symbol $-$ at $(0.04,0.03)$ can be distinguished from many other
symbol only by using the re-curvature feature for the first stroke. In contrast,
the $\top$ symbol cannot be distinguished from any other symbol by this feature.

Five features are evaluated in the following. The mean and variance of the first
dimension of those features was plotted to give the reader an impression of
how well they separate symbols and which symbols cannot be separated by those
single features. Additionally, the baseline systems were extended by those
features to measure their influence on the three error measures.

\subsection{Re-curvature}
The re-curvature feature is a feature for single strokes. It was defined on \cpageref{feature:recurvature} as

\[\recurvature(stroke) := \frac{\height(stroke)}{\text{length}(stroke)}\]

As both, the height and the distance are measured in the same unit,
the feature is a dimensionless quantity.

In order to get a constant feature dimension it is required to define on how
many strokes this feature should get applied to. If a recording has less
strokes, the feature is defined to have the value $0$.

The results of this feature, applied to the first four strokes, are shown
in \cref{table:re-curvature-evaluation}. This feature improved classification
for all models and all error measures a lot.

\Cref{fig:re-curvature-mean-std-deviation} shows the mean and standard deviation
of the re-curvature feature for the first stroke of every recording.

\begin{figure}[H]
    \centering
    \includegraphics*[width=\textwidth]{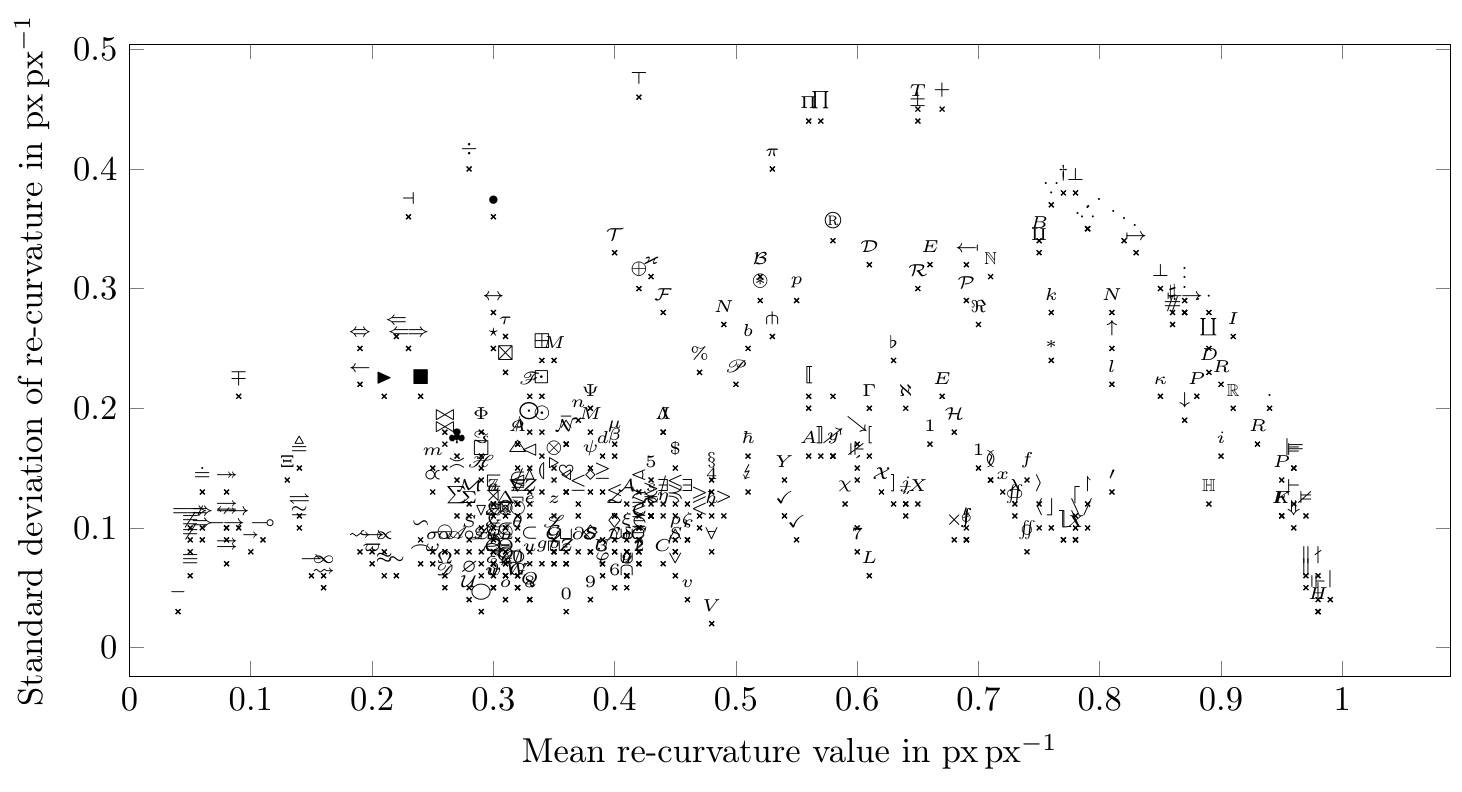}
    \caption{Mean and standard deviation of the re-curvature feature of the first
             stroke
             every symbol. The $-$ (0.04,0.03) and the $\rightsquigarrow$
             (0.16,0.05) are well-separated, but this feature is not able to
             distinguish $\pm$ (0.65,0.44) from either of them.}
\label{fig:re-curvature-mean-std-deviation}
\end{figure}

\begin{table}[h]
    \centering
    \begin{tabular}{lrrrrrrr}
    \toprule
    \multirow{2}{*}{System}  & \multicolumn{6}{c}{Classification error}\\
    \cmidrule(l){2-7}
                & TOP1                   & change                 & TOP3                   & change                 & MER                & change \\\midrule
    $B_{1,rec}$ & $\SI{22.14}{\percent}$ & $\SI{-1.20}{\percent}$ &  $\SI{5.96}{\percent}$ & $\SI{-0.84}{\percent}$ & $\SI{5.86}{\percent}$ & $\SI{-0.78}{\percent}$ \\
    $B_{2,rec}$ & \underline{$\SI{20.65}{\percent}$} & $\SI{-0.86}{\percent}$ & \underline{$\SI{5.19}{\percent}$} & $\SI{-0.56}{\percent}$ & \underline{$\SI{5.10}{\percent}$} & $\SI{-0.57}{\percent}$ \\
    $B_{3,rec}$ & $\SI{20.84}{\percent}$ & $\SI{-1.09}{\percent}$ &  $\SI{5.30}{\percent}$ & $\SI{-0.44}{\percent}$ & $\SI{5.22}{\percent}$ & $\SI{-0.42}{\percent}$ \\
    $B_{4,rec}$ & $\SI{23.25}{\percent}$ & $\SI{-0.63}{\percent}$ &  $\SI{5.90}{\percent}$ & $\SI{-0.22}{\percent}$ & $\SI{5.65}{\percent}$ & $\SI{-0.39}{\percent}$\\
    \bottomrule
    \end{tabular}
    \caption{Evaluation of baseline systems $B_1$ -- $B_4$ with an additional 
             re-curvature feature (rec) for each of the 4 strokes. All error
             measures improved notably.}
\label{table:re-curvature-evaluation}
\end{table}
\clearpage

\subsection{Stroke Center Point}

The stroke center point is a 2-dimensional feature. It calculates the center of
mass of a stroke by calculating the arithmetic mean of its coordinates. The
feature was added to all four baseline systems
$B_1$ -- $B_4$. As those systems had four strokes, the feature was applied 
for four strokes resulting in 8 new features.

\Cref{table:stroke-center-point-evaluation} shows the results
of this experiment. The results changed by less than the range of random
weight initialization which indicates that this feature is useless.

\begin{table}[h]
    \centering
    \begin{tabular}{lrrrrrrr}
    \toprule
    \multirow{2}{*}{System}  & \multicolumn{6}{c}{Classification error}\\
    \cmidrule(l){2-7}
               & TOP1                   & change                 & TOP3                   & change                 & MER                & change \\\midrule
    $B_{1,cp}$ & $\SI{23.18}{\percent}$ & $\SI{-0.16}{\percent}$ &  $\SI{6.53}{\percent}$ & $\SI{-0.27}{\percent}$ & $\SI{6.39}{\percent}$ & $\SI{-0.25}{\percent}$ \\
    $B_{2,cp}$ & $\SI{21.74}{\percent}$ & $\SI{+0.23}{\percent}$ &  $\SI{5.85}{\percent}$ & $\SI{+0.10}{\percent}$ & $\SI{5.74}{\percent}$ & $\SI{+0.07}{\percent}$ \\
    $B_{3,cp}$ & \underline{$\SI{21.19}{\percent}$} & $\SI{-0.74}{\percent}$ & \underline{$\SI{5.63}{\percent}$} & $\SI{-0.11}{\percent}$ & \underline{$\SI{5.55}{\percent}$} & $\SI{-0.09}{\percent}$ \\
    $B_{4,cp}$ & $\SI{23.94}{\percent}$ & $\SI{+0.06}{\percent}$ &  $\SI{6.24}{\percent}$ & $\SI{+0.12}{\percent}$ & $\SI{6.08}{\percent}$ & $\SI{+0.04}{\percent}$\\
    \bottomrule
    \end{tabular}
    \caption{Evaluation of baseline systems $B_1$ -- $B_4$ with additional 
             stroke center point features (cp) for 4 strokes. The results
             indicate that the feature is useless.}
\label{table:stroke-center-point-evaluation}
\end{table}

\subsection{Ink}
The ink feature measures how long each stroke is. This feature improved all
models except for $B_4$ as one can see in the results listed in
\cref{table:ink-feature-evaluation}. Although the experiment was executed
multiple times for $B_4$, all evaluations showed that the ink feature made
$B_4$ perform worse.

The mean-standard deviation scatterplot is shown in
\cref{fig:ink-mean-std-deviation}

\begin{table}[h]
    \centering
    \begin{tabular}{lrrrrrrr}
    \toprule
    \multirow{2}{*}{System}  & \multicolumn{6}{c}{Classification error}\\
    \cmidrule(l){2-7}
              & TOP1                   & change                 & TOP3                   & change                 & MER                & change \\\midrule
    $B_{1,i}$ & $\SI{22.22}{\percent}$ & $\SI{-1.12}{\percent}$ &  $\SI{5.88}{\percent}$ & $\SI{-0.92}{\percent}$ & $\SI{5.77}{\percent}$ & $\SI{-0.87}{\percent}$ \\
    $B_{2,i}$ & \underline{$\SI{20.91}{\percent}$} & $\SI{-0.60}{\percent}$ & \underline{$\SI{5.20}{\percent}$} & $\SI{-0.55}{\percent}$ & \underline{$\SI{5.11}{\percent}$} & $\SI{-0.53}{\percent}$ \\
    $B_{3,i}$ & $\SI{21.33}{\percent}$ & $\SI{-0.60}{\percent}$ &  $\SI{5.28}{\percent}$ & $\SI{-0.46}{\percent}$ & $\SI{5.21}{\percent}$ & $\SI{-0.43}{\percent}$ \\
    $B_{4,i}$ & $\SI{27.15}{\percent}$ & $\SI{+3.27}{\percent}$ &  $\SI{6.60}{\percent}$ & $\SI{+0.48}{\percent}$ & $\SI{6.31}{\percent}$ & $\SI{+0.27}{\percent}$\\
    \bottomrule
    \end{tabular}
    \caption{Evaluation of baseline systems $B_1$ -- $B_4$ with additional 
             ink feature (i). The small systems $B_1$ -- $B_3$ benefit from
             this feature, but the bigger model $B_4$ performs worse.}
\label{table:ink-feature-evaluation}
\end{table}
\clearpage

\subsection{Stroke Count}
The number of strokes is a strong single feature, because most people tend to
use the same number of strokes for a given symbol. There are some symbols where
people make variations, but those variations are only if two strokes are
connected or not. For example, The letter \enquote{E} is by some people drawn as
4 strokes, by others as an \enquote{L} with two more strokes.
\Cref{fig:stroke-count-mean-std-deviation} shows the mean-standard deviation
scatterplot of the number of strokes for each of the evaluated symbols excluding
a few listed in \cref{table:stroke-count-cut-out}.

The feature improves recognition rates for the models $B_1$--$B_3$, but makes
$B_4$ perform worse.

\begin{figure}[H]
    \centering
    \includegraphics*[width=\textwidth]{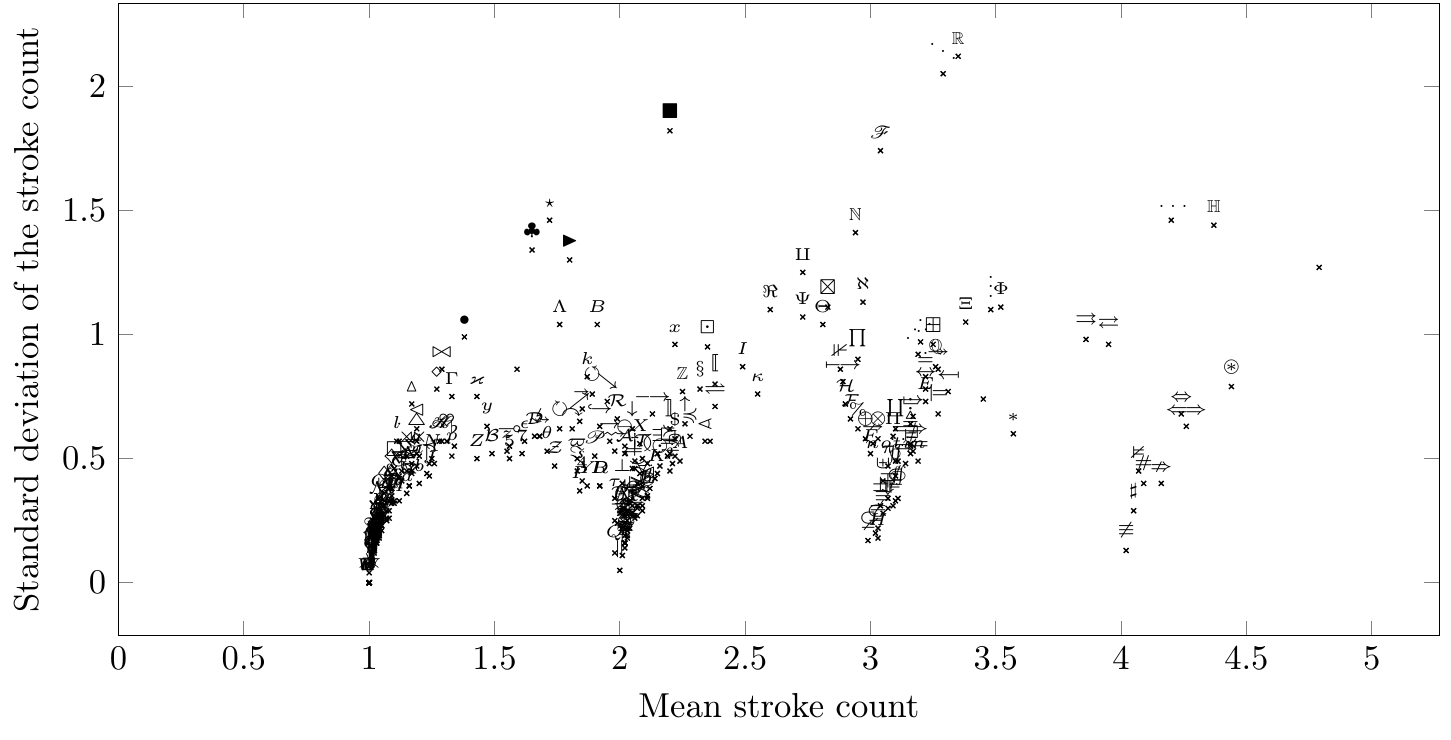}
    \caption{Mean-standard deviation scatterplot of the stroke count feature
             as it was introduced on \cpageref{sec:features-evaluation}.}
\label{fig:stroke-count-mean-std-deviation}
\end{figure}

\begin{table}[h]
    \centering
    \begin{tabular}{lrrrrrrr}
    \toprule
    \multirow{2}{*}{System}  & \multicolumn{6}{c}{Classification error}\\
               & TOP1                   & change                 & TOP3                   & change                 & MER                & change \\\midrule
    $B_{1,sc}$ & $\SI{23.28}{\percent}$ & $\SI{-0.06}{\percent}$ &  $\SI{6.71}{\percent}$ & $\SI{-0.09}{\percent}$ & $\SI{6.55}{\percent}$ & $\SI{-0.09}{\percent}$ \\
    $B_{2,sc}$ & $\SI{21.75}{\percent}$ & $\SI{-0.24}{\percent}$ &  $\SI{5.46}{\percent}$ & $\SI{-0.29}{\percent}$ & \underline{$\SI{5.38}{\percent}$} & $\SI{-0.29}{\percent}$ \\
    $B_{3,sc}$ & \underline{$\SI{21.42}{\percent}$} & $\SI{-0.51}{\percent}$ & \underline{$\SI{5.45}{\percent}$} & $\SI{-0.29}{\percent}$ & $\SI{5.39}{\percent}$ & $\SI{-0.25}{\percent}$ \\
    $B_{4,sc}$ & $\SI{25.28}{\percent}$ & $\SI{+1.40}{\percent}$ &  $\SI{6.87}{\percent}$ & $\SI{+0.75}{\percent}$ & $\SI{6.50}{\percent}$ & $\SI{+0.46}{\percent}$\\
    \bottomrule
    \end{tabular}
    \caption{Evaluation of baseline systems $B_1$ -- $B_4$ with additional 
             stroke count feature (sc). $B_{2,sc}$ and $B_{3,sc}$ performed
             slightly better, $B_{1,sc}$ barely changed, but
             $B_{4,sc}$ performed much worse than before.}
\label{table:stroke-count-evaluation}
\end{table}

\subsection{Aspect Ratio}
The aspect ratio is a 1~dimensional feature of a recording that is calculated
by calculating the ratio
\[\aspectratio(recording) = \frac{\widthimage(recording)}{\heightimage(recording)}\]

However, as the width (and the height) get calculated by subtracting the minimum
$x$ ($y$) value from the maximum $x$ ($y$) value, it can be $0$. In order to
avoid zero division errors $+0.01$ was added to both values, width and height.
This could be seen as the thickness of a stroke and could have an impact on
the value of this feature.

One dimension --- either width or height --- has the value $1$ as all baseline
systems use the scale and shift algorithm (except if it was only a point).

\Cref{fig:aspect-ratio-mean-std-deviation} on \cpageref{fig:aspect-ratio-mean-std-deviation}
shows a scatterplot of the mean and the standard deviation of this feature and
\cref{table:aspect-cut-out} on \cpageref{table:aspect-cut-out} lists all
symbols that were not used in the figure.

The evaluation results showed that the models $B_1$ and $B_3$ improved with this 
feature by about $\SI{0.2}{\percent}$ MER error, but $B_4$ got worse. The
error rate of model $B_{2,ar}$ barely changed.

\begin{table}[h]
    \centering
    \begin{tabular}{lrrrrrrr}
    \toprule
    \multirow{2}{*}{System}  & \multicolumn{6}{c}{Classification error}\\
    \cmidrule(l){2-7}
               & TOP1                   & change                 & TOP3                   & change                 & MER                & change \\\midrule
    $B_{1,ar}$ & $\SI{23.07}{\percent}$ & $\SI{-0.27}{\percent}$ &  $\SI{6.55}{\percent}$ & $\SI{-0.25}{\percent}$ & $\SI{6.45}{\percent}$ & $\SI{-0.19}{\percent}$ \\
    $B_{2,ar}$ & \underline{$\SI{21.45}{\percent}$} & $\SI{-0.06}{\percent}$ &  $\SI{5.67}{\percent}$ & $\SI{-0.08}{\percent}$ & $\SI{5.60}{\percent}$ & $\SI{-0.07}{\percent}$ \\
    $B_{3,ar}$ & $\SI{21.49}{\percent}$ & $\SI{-0.44}{\percent}$ & \underline{$\SI{5.36}{\percent}$} & $\SI{-0.38}{\percent}$ & \underline{$\SI{5.28}{\percent}$} & $\SI{-0.36}{\percent}$ \\
    $B_{4,ar}$ & $\SI{25.01}{\percent}$ & $\SI{+1.13}{\percent}$ &  $\SI{6.45}{\percent}$ & $\SI{+0.33}{\percent}$ & $\SI{6.10}{\percent}$ & $\SI{+0.06}{\percent}$\\
    \bottomrule
    \end{tabular}
    \caption{Evaluation of baseline systems $B_1$ -- $B_4$ with additional 
             aspect ratio feature (ar). The small systems
             $B_{1,ar}$--$B_{3,ar}$ improved, but $B_{4,ar}$ got worse.}
\label{table:aspect-ratio-feature-evaluation}
\end{table}

\section{System A: Greedy Time Warping}\label{system-a-evaluation} 

System A used only scaling and shifting as resampling. After that, greedy time
warping was applied (a variant of dynamic time warping that is faster, but
not optimal) to calculate the distance of two recordings.

A new recording was classified getting the minimal distance to any known
recording. The label of the recording with minimal distance was used as a
classification result.

This system needed about $\SI{22}{\second}$ in average
on Intel Pentium P6200 processor to classify a single recording, although the
amount of recordings per symbol was limited to 50 at maximum. So it was much
too slow.

The classification error was $\SI{85.46}{\percent}$. The TOP10 classification
error was $\SI{65.78}{\percent}$. So this system was clearly much worse than
the \glspl{MLP}.

Note that those results are much worse than what was achieved in
\cite{Kirsch} on a similar dataset. This is probably the reason, because
Kirsch did apply more preprocessing steps and tweaked the time warping approach.
However, even his results with time warping were much worse than what can be
done with \glspl{MLP}.

\section{System B: Multilayer Perceptrons} 

The tested systems were already described in \cref{table:baseline-systems}.
However, there are many parameters that might influence how fast a \gls{MLP} can
learn. The effect of changes to some of those parameters were tested and are
described in the following.

\subsection{Baseline Testing}
\Cref{fig:training-and-test-error-for-different-topologies}, a plot of the
validation and test error over the epochs shows that --- except for the system
$B_4$ --- the biggest drops in error are done until the \nth{400} epoch. After that,
there is almost no change. Only system $B_4$ seems to be able to improve
after that. Training $B_4$ for $\num{10000}$ epochs led to an TOP1
error of $\SI{21.00}{\percent}$ which is $\SI{2.88}{\percent}$ better than the
result with only $\num{1000}$ epochs of training. Model $B_3$ was also able to
improve, but not that much. The exact results for all models and errors are in
\cref{table:10000-epochs}.

\begin{table}[htb]
    \centering
    \begin{tabular}{lrrrrrrr}
    \toprule
    \multirow{2}{*}{System}  & \multicolumn{6}{c}{Classification error}\\
    \cmidrule(l){2-7}
                         & TOP1                   & change                 & TOP3                   & change                 & MER                & change \\\midrule
    $B_{1,epochs=\num{10000}}$ & $\SI{21.31}{\percent}$ & $\SI{-2.03}{\percent}$ &  $\SI{5.81}{\percent}$ & $\SI{-0.99}{\percent}$ & $\SI{5.68}{\percent}$ & $\SI{-0.96}{\percent}$ \\
    $B_{2,epochs=\num{10000}}$ & $\SI{21.47}{\percent}$ & $\SI{-0.04}{\percent}$ &  $\SI{5.84}{\percent}$ & $\SI{+0.09}{\percent}$ & $\SI{5.68}{\percent}$ & $\SI{+0.01}{\percent}$ \\
    $B_{3,epochs=\num{10000}}$ & $\SI{21.16}{\percent}$ & $\SI{-0.77}{\percent}$ & \underline{$\SI{5.44}{\percent}$} & $\SI{-0.30}{\percent}$ & \underline{$\SI{5.34}{\percent}$} & $\SI{-0.30}{\percent}$ \\
    $B_{4,epochs=\num{10000}}$ & \underline{$\SI{21.00}{\percent}$} & $\SI{-2.88}{\percent}$ &  $\SI{5.74}{\percent}$ & $\SI{-0.38}{\percent}$ & $\SI{5.64}{\percent}$ & $\SI{-0.40}{\percent}$\\
    \bottomrule
    \end{tabular}
    \caption{Evaluation of baseline systems $B_1$ -- $B_4$ after $\num{10000}$ 
             epochs of training. The models were tested after every epoch.
             There was no model in between that performed much better. So
             overfitting is not a problem in that case.}
\label{table:10000-epochs}
\end{table}

\Cref{fig:learning-curve} shows a learning curve for model $B_1$. Every point in
that plot was generated by artificially reducing the training set to a maximum
number (from 1 to 150) of training examples per symbol and training for only 300
epochs. As the training and the test error were plotted, one can see how more
data might affect the error rate. Although there are symbols with only $50$
recordings, it seems not to make a difference if one had $150$ or more
recordings per symbol for a \gls{MLP} with only one hidden layer. The TOP1 error
on the training set for 6 training examples per symbol is already at
$\SI{29}{\percent}$. This indicates that more or better features could improve
the model, whereas more training examples will not help to get below
$\SI{29}{\percent}$.

\begin{figure}[htb]
    \centering
    \includegraphics*[width=\textwidth]{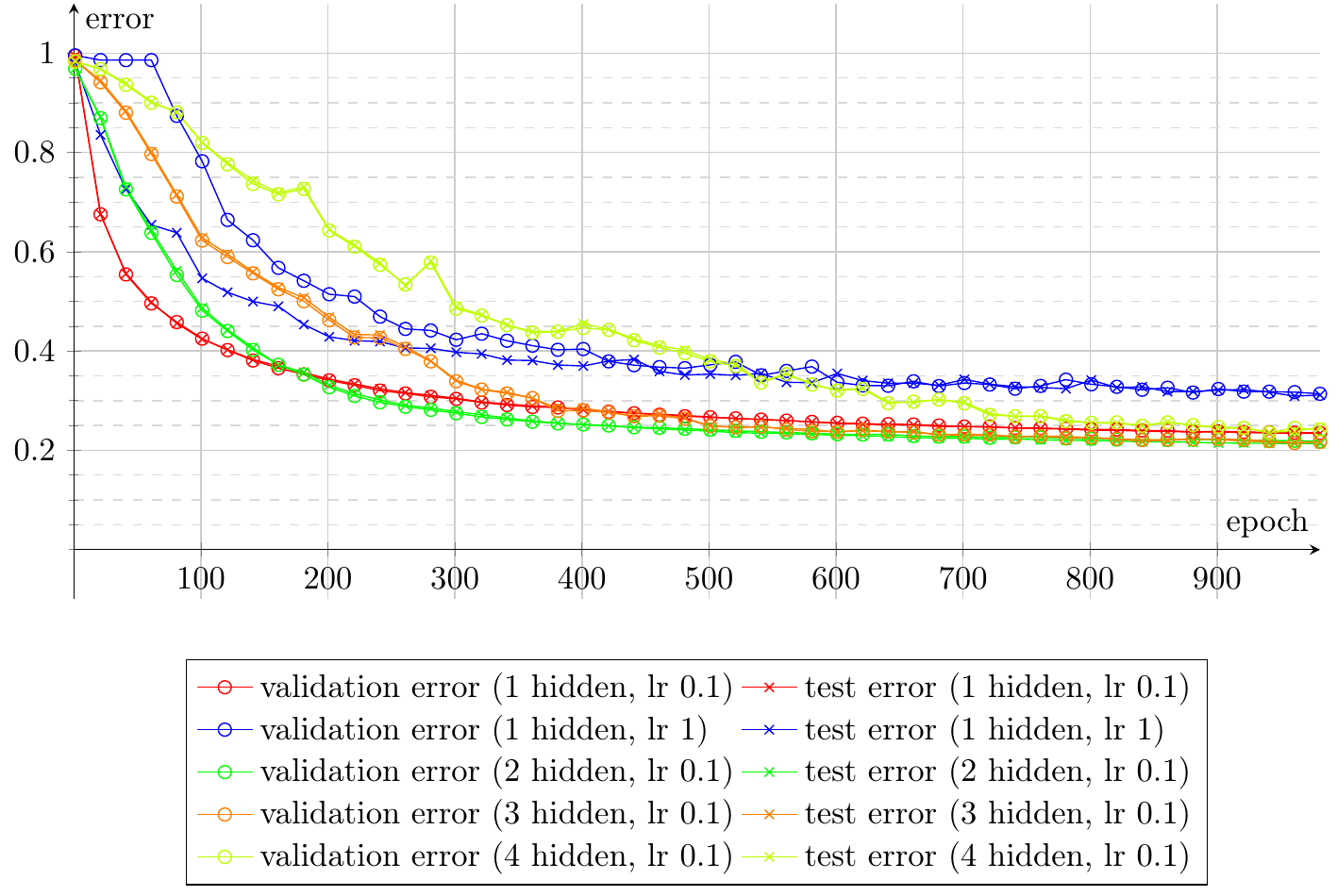}
    \caption{Training- and test error by number of trained epochs for different
             topologies. The curves all show almost no difference in
             validation and test error. The error for all curves with learning
             rates of $\eta = 0.1$ converge to similar values, although
             the error of $B_4$ still drops while all other models do not
             perform better with more training. A learning rate of
             $\eta=1$ is much worse than $\eta=0.1$.}
\label{fig:training-and-test-error-for-different-topologies}
\end{figure}

\begin{figure}[htb]
    \centering
    \includegraphics*[width=\textwidth]{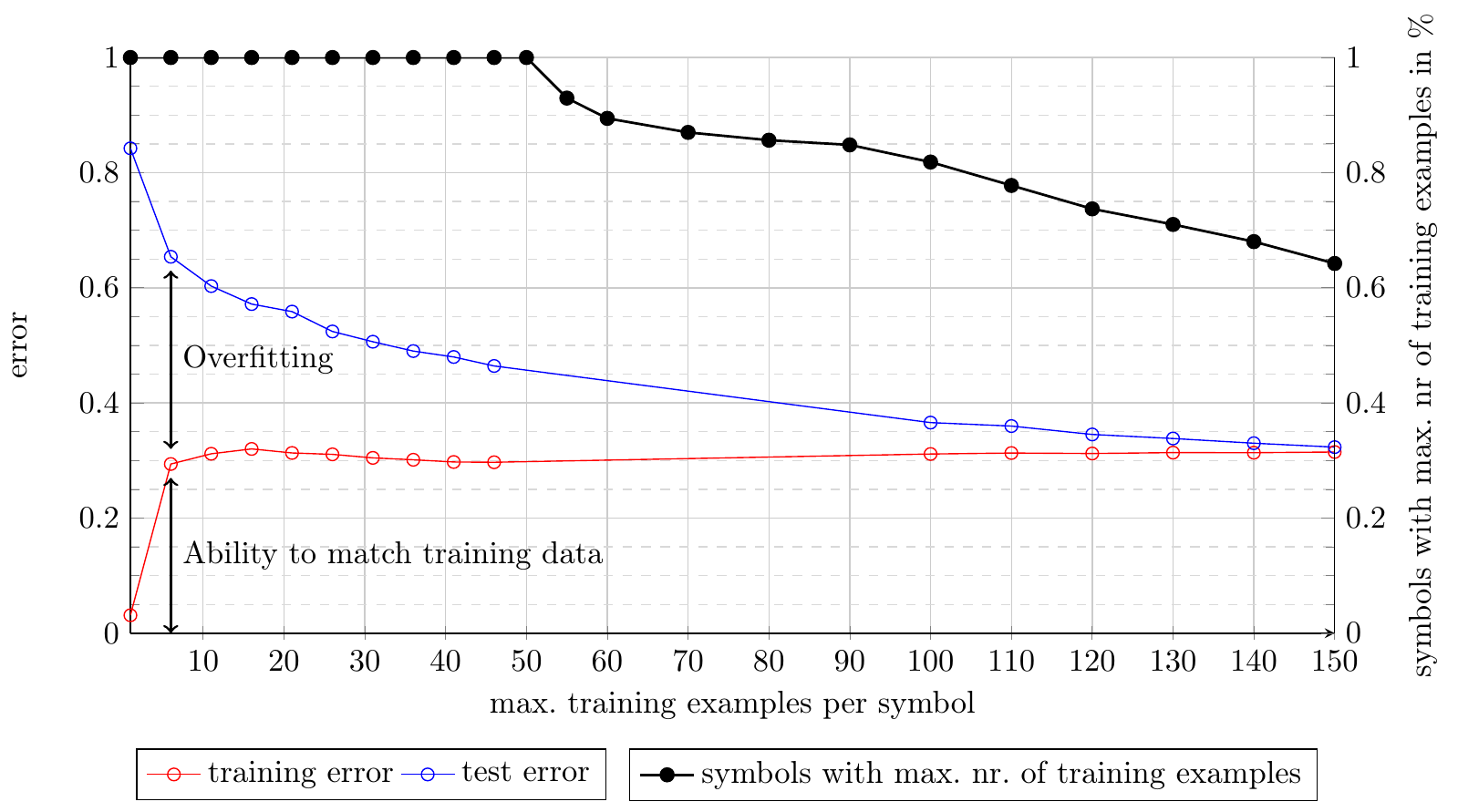}
    \caption{This plot shows the learning curve for a 160:500:369 \gls{MLP}.
             The $x$ axis shows the number of training examples per symbol, the
             $y$ axis shows the error for the colored lines and the percentage
             of symbols that had at least the maximum number of training
             examples. The training and the testing error is plotted as well as
             the percentage of symbols with the maximum number of training
             examples.}
\label{fig:learning-curve}
\end{figure}

\subsection{Execution Time}
The neural network training was executed on a Nvidia GeForce GTX Titan Black.
It took about 12~minutes to train $B_1$ and about 25~minutes to train $B_4$.

The training of $B_1$ executed on a Intel P6200 CPU was aborted after
$\SI{31}{\hour}$. This means the execution time was reduced by GPU training to
about $\SI{1.6}{\percent}$ of the time it took before.

\subsection{Learning Rate}\label{subsec:learning-rate}
The choice of the learning rate in mini batch training determines how wide the
steps in gradient descent are. Bigger steps lead to a faster improvement at
the beginning, but at the end the algorithm might jump back and forth and
not be able to improve. \Cref{table:learning-rate-choices}
shows the results after 1000 epochs of mini batch training with different
choices for the learning rate. Tested were learning rates of $0.05$, $0.1$,
$0.2$ and $1.0$. For all models, a learning rate of $0.1$ was the best choice.
\clearpage

\subsection{Momentum}
The momentum $\alpha$ in mini-batch training is, just like the learning rate,
important for the speed of improvements and eventually also for the final
result. It was explained in \cref{subsec:momentum}.

\begin{table}[htb]
    \centering
    \begin{tabular}{lrrrrrr}
    \toprule
    \multirow{2}{*}{System}  & \multicolumn{6}{c}{Classification error}\\
    \cmidrule(l){2-7}
                           & TOP1                   & change                 & TOP3                  & change                 & MER                 & change \\\midrule
    $B_{1,\alpha = 0.1}$   & $\SI{23.34}{\percent}$ &                        & $\SI{6.80}{\percent}$ &                        & $\SI{6.64}{\percent}$ & \\
    $B_{1,\alpha = 0.9}$   & $\SI{22.51}{\percent}$ & $\SI{-0.83}{\percent}$ & $\SI{6.88}{\percent}$ & $\SI{+0.08}{\percent}$ & $\SI{6.74}{\percent}$ & $\SI{+0.10}{\percent}$\\\midrule
    $B_{2,\alpha = 0.1}$   & $\SI{21.51}{\percent}$ &                        & $\SI{5.75}{\percent}$ &                        & $\SI{5.67}{\percent}$ & \\
    $B_{2,\alpha = 0.9}$   & $\SI{21.17}{\percent}$ & $\SI{-0.34}{\percent}$ & $\SI{5.96}{\percent}$ & $\SI{+0.21}{\percent}$ & $\SI{5.85}{\percent}$ & $\SI{+0.18}{\percent}$\\\midrule
    $B_{3,\alpha = 0.1}$   & $\SI{21.93}{\percent}$ &                        & $\SI{5.74}{\percent}$ &                        & $\SI{5.64}{\percent}$ & \\
    $B_{3,\alpha = 0.9}$   & \underline{$\SI{20.57}{\percent}$} & $\SI{-1.36}{\percent}$ & \underline{$\SI{5.59}{\percent}$} & $\SI{-0.15}{\percent}$ & \underline{$\SI{5.53}{\percent}$} & $\SI{-0.11}{\percent}$\\\midrule
    $B_{4,\alpha = 0.1}$   & $\SI{23.88}{\percent}$ &                        & $\SI{6.12}{\percent}$ &                        & $\SI{6.04}{\percent}$ & \\
    $B_{4,\alpha = 0.9}$   & $\SI{21.39}{\percent}$ & $\SI{-2.49}{\percent}$ & $\SI{6.00}{\percent}$ & $\SI{-0.12}{\percent}$ & $\SI{5.91}{\percent}$ & $\SI{-0.13}{\percent}$\\
    \bottomrule
    \end{tabular}
    \caption{Evaluation results of the systems $B_1$ -- $B_4$ with adjusted
             momentums $\alpha$. The column \enquote{change} was left blank,
             because the baseline systems $B_1$ -- $B_4$ use a momentum of
             $\alpha=0.1$.}
\label{table:momentum-rate-choices}
\end{table}

\subsection{Pretraining}\label{subsec:pretraining-evaluation}
Pretraining is a technique used to improve the training of deep neural networks.
Two pretraining algorithms are described in \cref{subsec:pretraining} on
\cpageref{subsec:pretraining}: \Gls{SLP} and
denoising auto-encoders.

\Cref{fig:training-and-test-error-for-different-topologies-pretraining} shows
the evolution of validation and test errors over 1000 epochs with supervised
layer-wise pretraining and without pretraining. It clearly shows that this
kind of pretraining improves the classification performance. Detailed
results are listed in \cref{table:pretraining-copy-before}.

\begin{figure}[htb]
    \centering
    \includegraphics*[width=\textwidth]{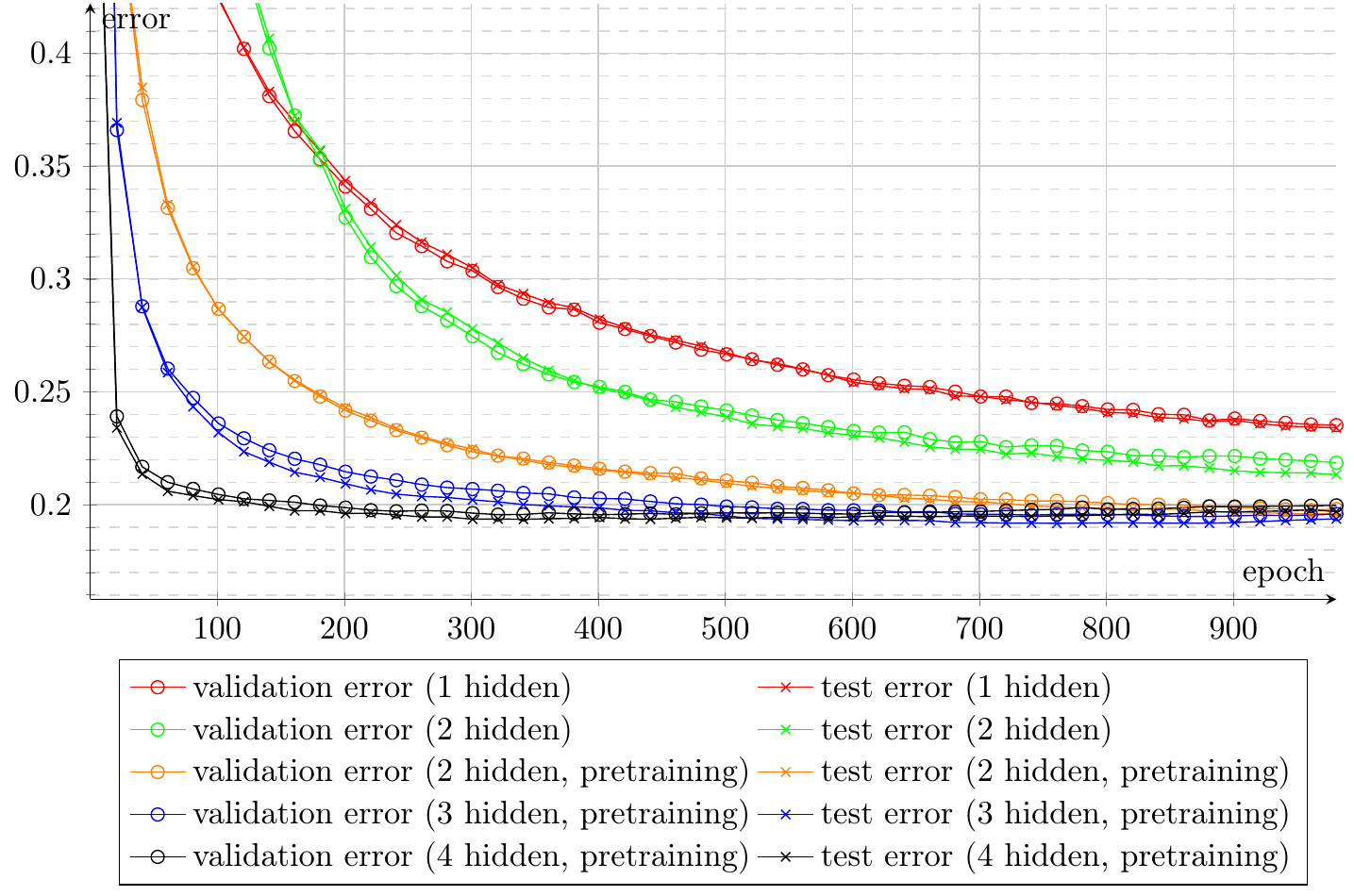}
    \caption{Training- and test error by number of trained epochs for different
             topologies with \gls{SLP}. The plot shows
             that all pretrained systems performed much better than the systems
             without pretraining. All plotted systems did not improve
             with more epochs of training.}
\label{fig:training-and-test-error-for-different-topologies-pretraining}
\end{figure}

\begin{table}[thb]
    \centering
    \begin{tabular}{lrrrrrr}
    \toprule
    \multirow{2}{*}{System}& \multicolumn{6}{c}{Classification error}\\
    \cmidrule(l){2-7}
              & TOP1                   & change                 & TOP3                  & change                 & MER                & change \\\midrule
    $B_1$     & $\SI{23.34}{\percent}$ &                        & $\SI{6.80}{\percent}$ &                        & $\SI{6.64}{\percent}$ & \\
    $B_{2,SLP}$ & $\SI{19.89}{\percent}$ & $\SI{-1.62}{\percent}$ & $\SI{4.76}{\percent}$ & $\SI{-0.99}{\percent}$ & $\SI{4.68}{\percent}$ & $\SI{-0.99}{\percent}$\\
    $B_{3,SLP}$ & \underline{$\SI{19.43}{\percent}$} & $\SI{-2.50}{\percent}$ & \underline{$\SI{4.64}{\percent}$} & $\SI{-1.10}{\percent}$ & \underline{$\SI{4.54}{\percent}$} & $\SI{-1.10}{\percent}$\\
    $B_{4,SLP}$ & $\SI{19.63}{\percent}$ & $\SI{-4.25}{\percent}$ & $\SI{4.66}{\percent}$ & $\SI{-1.46}{\percent}$ & $\SI{4.55}{\percent}$ & $\SI{-1.49}{\percent}$\\
    \bottomrule
    \end{tabular}
    \caption{Systems with \gls{SLP} compared to
             pure gradient descent. The \gls{SLP} systems performed notably
             better. Although the pretrained systems $B_{i,SLP}$ got $1000$
             epochs of training for each layer whereas the systems $B_{i}$ only
             got $1000$ epochs of training in total, it is important to note
             that the systems $B_1$--$B_3$ were not able to improve with more
             training epochs.\\
             Denoising auto-encoders (da) on the other hand made the system much
             worse.}
\label{table:pretraining-copy-before}
\end{table}

Pretraining with denoising auto-encoder lead to the much worse results listed in
\cref{table:pretraining-denoising-auto-encoder}. The first layer used a $\tanh$
activation function. Every layer was trained for $1000$ epochs and the
\gls{MSE} loss function. A learning-rate of $\eta = 0.001$, a corruption of
$0.3$ and a $L_2$ regularization of $\lambda = 10^{-4}$ were chosen. This
pretraining setup made all systems with all error measures perform much worse.

\begin{table}[tb]
    \centering
    \begin{tabular}{lrrrrrr}
    \toprule
    \multirow{2}{*}{System}  & \multicolumn{6}{c}{Classification error}\\
    \cmidrule(l){2-7}
              & TOP1                   & change                 & TOP3                  & change                 & MER                   & change \\\midrule
    $B_{1,p}$ & $\SI{23.75}{\percent}$ & $\SI{+0.41}{\percent}$ & $\SI{7.19}{\percent}$ & $\SI{+0.39}{\percent}$ & $\SI{6.98}{\percent}$ & $\SI{+0.34}{\percent}$ \\
    $B_{2,p}$ & \underline{$\SI{22.76}{\percent}$} & $\SI{+1.25}{\percent}$ & $\SI{6.38}{\percent}$ & $\SI{+0.63}{\percent}$ & $\SI{6.28}{\percent}$ & $\SI{+0.61}{\percent}$ \\
    $B_{3,p}$ & $\SI{23.10}{\percent}$ & $\SI{+1.17}{\percent}$ & \underline{$\SI{6.14}{\percent}$} & $\SI{+0.40}{\percent}$ & \underline{$\SI{6.04}{\percent}$} & $\SI{+0.40}{\percent}$ \\
    $B_{4,p}$ & $\SI{25.59}{\percent}$ & $\SI{+1.71}{\percent}$ & $\SI{6.99}{\percent}$ & $\SI{+0.87}{\percent}$ & $\SI{6.88}{\percent}$ & $\SI{+0.84}{\percent}$ \\
    \bottomrule
    \end{tabular}
    \caption{Systems with denoising auto-encoder pretraining compared to pure
             gradient descent. The pretrained systems clearly performed worse.}
\label{table:pretraining-denoising-auto-encoder}
\end{table}

\subsection{Newbob Training}\label{subsec:newbob-evaluation}
Newbob is a training mode that adjusts the training rate according to the
improvement over the last epoch. It was explained on \cpageref{subsec:newbob-training}.
\Cref{fig:newbob-training-and-test-error-for-different-topologies-and-trainings}
shows the result of the training and test error with newbob training.
It was used with $\theta_1 = \SI{0.5}{\percent}$ and $\theta_2 = \SI{0.1}{\percent}$.

\Cref{table:newbob-evaluation} shows results for different choices of parameters
for newbob training. Although the training finished fast (most of the time
after about $80$~epochs, but always before the \nth{400} epoch), no result of
newbob training was better than the baseline system.

\begin{figure}[htb]
    \centering
    \includegraphics*[width=\textwidth]{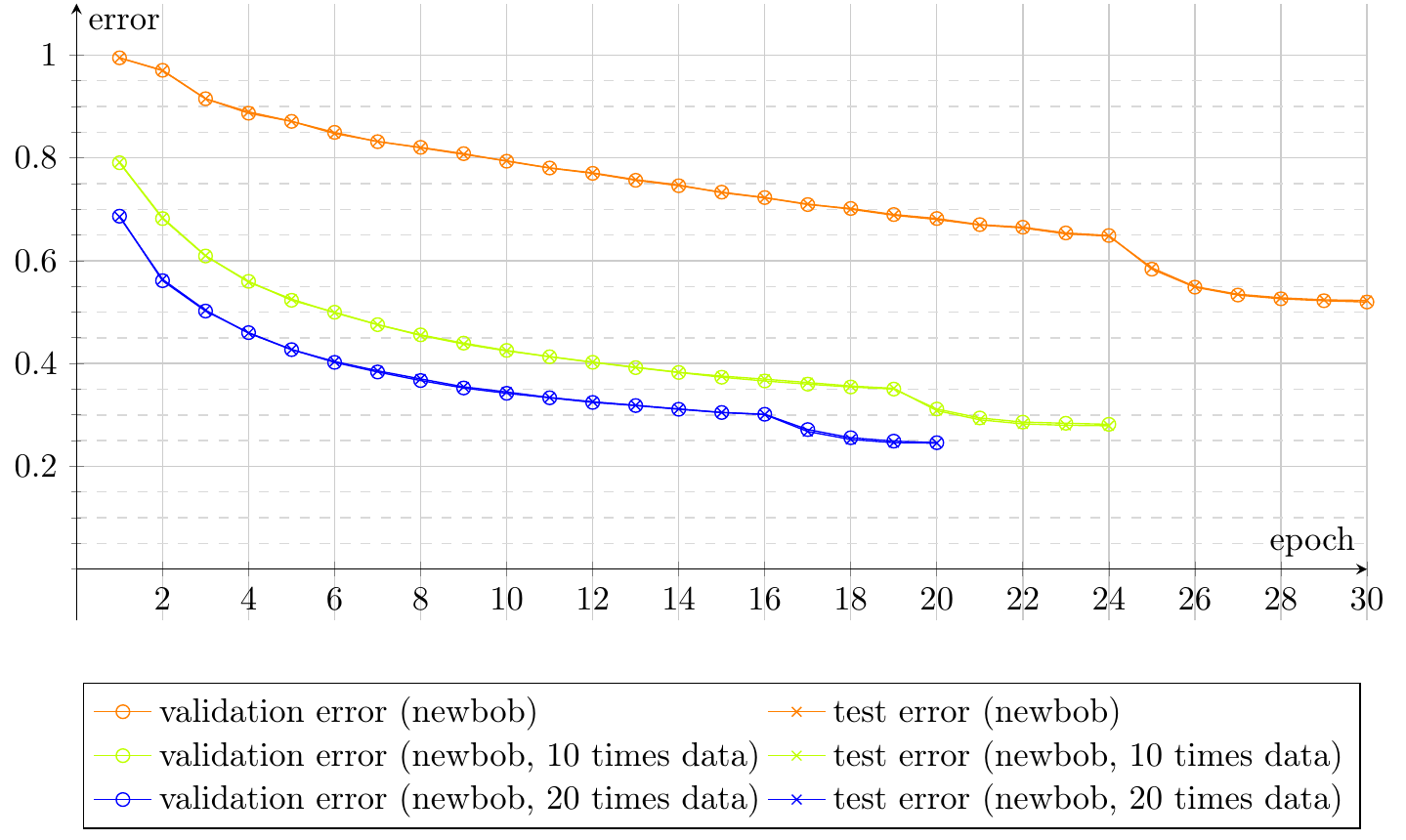}
    \caption{Training- and test error by number of trained epochs for different
             topologies and training types.}
\label{fig:newbob-training-and-test-error-for-different-topologies-and-trainings}
\end{figure}

\begin{table}[htb]
    \centering
    \begin{tabular}{lrrrrrr}
    \toprule
    \multirow{2}{*}{System}                        & \multicolumn{6}{c}{Classification error}\\
    \cmidrule(l){2-7}
                                                   & TOP1                   & change                 &  TOP3                  & change                 & MER                   & change \\\midrule
    $B_{1,\alpha=0.1,dm=20,d=0.5,\theta=0.001}$    & $\SI{34.86}{\percent}$ &$\SI{+11.52}{\percent}$ & $\SI{16.02}{\percent}$ & $\SI{+9.22}{\percent}$ & $\SI{15.69}{\percent}$ & $\SI{+9.05}{\percent}$ \\
    $B_{1,\alpha=1.0,dm=20,d=0.95,\theta=0.00001}$ & $\SI{30.60}{\percent}$ & $\SI{+7.26}{\percent}$ & $\SI{12.36}{\percent}$ & $\SI{+5.56}{\percent}$ & $\SI{12.09}{\percent}$ & $\SI{+5.45}{\percent}$ \\
    $B_{1,\alpha=0.1,dm=20,d=0.50,\theta=0.00001}$ & $\SI{28.24}{\percent}$ & $\SI{+4.90}{\percent}$ & $\SI{10.92}{\percent}$ & $\SI{+4.12}{\percent}$ & $\SI{10.68}{\percent}$ & $\SI{+4.04}{\percent}$ \\
    $B_{1,\alpha=0.1,dm=20,d=0.75,\theta=0.00001}$ & $\SI{27.70}{\percent}$ & $\SI{+4.36}{\percent}$ & $\SI{10.67}{\percent}$ & $\SI{+3.87}{\percent}$ & $\SI{10.41}{\percent}$ & $\SI{+3.77}{\percent}$ \\
    $B_{1,\alpha=0.1,dm=20,d=0.95,\theta=0.00001}$ & \underline{$\SI{26.77}{\percent}$} & $\SI{+3.43}{\percent}$ & \underline{$\SI{ 9.59}{\percent}$} & $\SI{+2.79}{\percent}$ & \underline{$\SI{ 9.42}{\percent}$} & $\SI{+2.78}{\percent}$ \\
    $B_{1,\alpha=0.1, dm=1,d=0.95,\theta=0.00001}$ & $\SI{26.80}{\percent}$ & $\SI{+3.46}{\percent}$ & $\SI{10.21}{\percent}$ & $\SI{+3.41}{\percent}$ & $\SI{10.00}{\percent}$ & $\SI{ 3.36}{\percent}$ \\\midrule
    $B_{2,\alpha=0.1,dm=20,d=0.95,\theta=0.00001}$ & $\SI{48.08}{\percent}$ & $\SI{26.57}{\percent}$ & $\SI{27.78}{\percent}$ & $\SI{22.03}{\percent}$ & $\SI{27.31}{\percent}$ & $\SI{21.64}{\percent}$ \\
    $B_{3,\alpha=0.1,dm=20,d=0.95,\theta=0.00001}$ & $\SI{98.12}{\percent}$ & $\SI{76.19}{\percent}$ & $\SI{97.06}{\percent}$ & $\SI{91.32}{\percent}$ & $\SI{97.03}{\percent}$ & $\SI{91.39}{\percent}$ \\
    $B_{4,\alpha=0.1,dm=20,d=0.95,\theta=0.00001}$ & $\SI{98.60}{\percent}$ & $\SI{74.72}{\percent}$ & $\SI{96.76}{\percent}$ & $\SI{90.64}{\percent}$ & $\SI{96.72}{\percent}$ & $\SI{90.68}{\percent}$ \\
    \bottomrule
    \end{tabular}
    \caption{Evaluation results of Newbob trainings with different learning
             rates $\alpha$, data augmentations $dm$ (by copying data),
             newbob weight decays $d$ and newbob thresholds $\theta$.}
\label{table:newbob-evaluation}
\end{table}
\clearpage

\section{Optimized Recognizer}\label{sec:complex-recognizer}
All preprocessing steps and features that were useful were combined to
create a recognizer that should perform best.

All models were much better than everything that was tried before. The results
of this experiment show that single-symbol recognition with
\totalClassesAnalyzed{} classes and usual touch devices and the mouse can be
done with a TOP1 error rate of $\SI{18.56}{\percent}$, a TOP3 error of
$\SI{4.11}{\percent}$ and a MER error rate of $\SI{4.01}{\percent}$. This was
achieved by a \gls{MLP} with a $167:500:500:\totalClassesAnalyzed{}$ topology.

It used an algorithm to connect strokes of which the ends were less than
$\SI{10}{\pixel}$ away, scaled each recording to a unit square and shifted this
unit square to $(0,0)$. After that, a linear resampling step was applied to the
first 4 strokes to resample them to 20 points each. All other strokes were
discarded.

The 167 features were

\begin{itemize}
     \item the first 4 strokes with 20 points per stroke resulting in 160
           features,
     \item the re-curvature for the first 4 strokes,
     \item the ink,
     \item the number of strokes and
     \item the aspect ratio
\end{itemize}

\Gls{SLP} was applied with $\num{1000}$ epochs per layer, a
learning rate of $\eta=0.1$ and a momentum of $\alpha=0.1$. After that, the
complete model was trained again for $1000$ epochs with standard mini-batch
gradient descent.

After the models $B_{1,c}$ -- $B_{4,c}$ were trained the first $1000$ epochs,
they were trained again for $1000$ epochs with a learning rate of $\eta = 0.05$.
\Cref{table:complex-recognizer-systems-evaluation} shows that
this improved the classifiers again.

System $B_{2,c}'$ had an error rate of $\SI{14.91}{\percent}$ if only the first
symbol and its equivalence class were accepted as correct.

\begin{table}[htb]
    \centering
    \begin{tabular}{lrrrrrr}
    \toprule
    \multirow{2}{*}{System}  & \multicolumn{6}{c}{Classification error}\\
    \cmidrule(l){2-7}
              & TOP1                   & change                 & TOP3                  & change                 & MER                   & change \\\midrule
    $B_{1,c}$ & $\SI{20.96}{\percent}$ & $\SI{-2.38}{\percent}$ & $\SI{5.24}{\percent}$ & $\SI{-1.56}{\percent}$ & $\SI{5.13}{\percent}$ & $\SI{-1.51}{\percent}$ \\
    $B_{2,c}$ & $\SI{18.26}{\percent}$ & $\SI{-3.25}{\percent}$ & $\SI{4.07}{\percent}$ & $\SI{-1.68}{\percent}$ & \underline{$\SI{3.98}{\percent}$} & $\SI{-1.69}{\percent}$ \\
    $B_{3,c}$ & \underline{$\SI{18.19}{\percent}$} & $\SI{-3.74}{\percent}$ & \underline{$\SI{4.06}{\percent}$} & $\SI{-1.68}{\percent}$ & $\SI{3.99}{\percent}$ & $\SI{-1.65}{\percent}$ \\
    $B_{4,c}$ & $\SI{18.57}{\percent}$ & $\SI{-5.31}{\percent}$ & $\SI{4.25}{\percent}$ & $\SI{-1.87}{\percent}$ & $\SI{4.18}{\percent}$ & $\SI{-1.86}{\percent}$ \\\midrule
    $B_{1,c}'$ & $\SI{19.33}{\percent}$ & $\SI{-1.63}{\percent}$ & $\SI{4.78}{\percent}$ & $\SI{-0.46}{\percent}$ & $\SI{4.67}{\percent}$ & $\SI{-0.46}{\percent}$ \\
    $B_{2,c}'$ & \underline{$\SI{17.52}{\percent}$} & $\SI{-0.74}{\percent}$ & \underline{$\SI{4.04}{\percent}$} & $\SI{-0.03}{\percent}$ & \underline{$\SI{3.96}{\percent}$} & $\SI{-0.02}{\percent}$ \\
    $B_{3,c}'$ & $\SI{17.65}{\percent}$ & $\SI{-0.54}{\percent}$ & $\SI{4.07}{\percent}$ & $\SI{+0.01}{\percent}$ & $\SI{4.00}{\percent}$ & $\SI{+0.01}{\percent}$ \\
    $B_{4,c}'$ & $\SI{17.82}{\percent}$ & $\SI{-0.75}{\percent}$ & $\SI{4.26}{\percent}$ & $\SI{+0.01}{\percent}$ & $\SI{4.20}{\percent}$ & $\SI{+0.02}{\percent}$ \\
    \bottomrule
    \end{tabular}
    \caption{Error rates of the optimized recognizer systems. The systems
             $B_{i,c}'$ were trained another $1000$ epochs with a learning rate
             of $\eta=0.05$. The value of the column \enquote{change} of the
             systems $B_{i,c}'$ is relative to $B_{i,c}$.}
\label{table:complex-recognizer-systems-evaluation}
\end{table}
\clearpage

\section{User Interviews}
Four people who used the recognizer on the website
\href{http://write-math.com}{write-math.com} with a Samsung Galaxy Note 10.1 (a
10.1 inch tablet with a stylus), with a smartphone and with a PC and a mouse
were asked for feedback about the input devices. They were asked which device
they prefer, why they prefer it, and if they could imagine entering
multi-symbol formulas on those devices.

User A preferred this tablet with the stylus over using the mouse or a smartphone.
She uses this tablet often and knows how to use the stylus for various
applications. While using the website, she noted that the recognition
works better if the symbol is written in a larger size. After noting that, she
entered all symbols in a bigger size. She could imagine using the tablet to
enter multi-symbol formulas.

User B used this tablet before, but did not use the stylus for writing before.
He could enter symbols without problems with the tablet, but preferred using the
index finger for writing instead of the styles. The reason is that he is used to
lay the heel of the hand on the surface on which he is writing, which did not
work well with the tablet. He preferred using a Nexus~4 smartphone to enter
symbols. He could not imagine entering complex formulas with a tablet or the
computer.

User C never used a touch device before. She preferred the tablet, but she had
problems with the stylus as the tablet did not allow to lay the heel of the hand
on the surface. This is the reason why she used the tablet with her fingers. She
could also imagine to use a recognition system with multi-symbol formulas on a
tablet, but not by using the mouse.

Like user C, user D never used a touch device before. He liked the stylus,
although it took him a few minutes to get used to not placing the heel of the
hand on the tablet. He also tried to enter symbols with a trackball, but
that didn't work for him. According to user~D, it is not possible to draw
straight lines with a trackball which makes it unusable for writing symbols.
He could also imagine to write multi-symbol formulas with a tablet.

\section{Evaluation Summary}

Five different classification systems were evaluated: A \gls{GTW} classifier
(see \cpageref{system-a-evaluation}) and four \glspl{MLP}. The \gls{GTW}
classifier performed much worse than all \gls{MLP} systems. For this reason,
it was only tested in one variant.

The four \glspl{MLP} $B_i, i \in \Set{1,2,3,4}$ had $i$ hidden layers with
$500$ neurons per layer. The baseline systems used scaling into a unit square,
shifting to $(0,0)$, 160~features that were coordinates of the first 20~points
of the first 4~strokes. The baseline systems were trained with mini-batch
gradient descent for 1000~epochs with a learning rate of $\eta=0.1$ and
a momentum of $\alpha=0.1$.

For many recordings, there are at least two possibilities which symbols humans
would recognize without context. This is the reason why the TOP1 error is
less meaningful for single-symbol recognition. Hence two other error measures
were calculated for every experiment: The TOP3 error and the MER error.
The MER error of the baseline system $B_2$ is $\SI{5.67}{\percent}$.

The \glspl{MLP} were tested with five~preprocessing algorithms, one~data
augmentation algorithm, five~features and five~variants for training in
16~separate experiments.

The effects of the preprocessing algorithms were often similar for the systems
$B_2$ and $B_3$. The system $B_4$ was very sensitive to changes. A possible
reason is the bigger number of weights, the higher order of internally computed
features and the fact that the system was still able to improve after $1000$
epochs of training. The system $B_4$ could be the best system if it was trained
long enough as \cref{table:10000-epochs} on \cpageref{table:10000-epochs} shows.
This means all results of the system $B_4$ should be taken with caution and
eventually be evaluated again with $\num{10000}$ training epochs. System $B_1$
on the other hand performs much worse than system $B_2$. The storage size and
the comparably small amount of necessary computing power to train $B_1$ are
the only reasons to use this system.

The optimized system $B_{2,c}'$ was the best evaluated system with a TOP1
error rate of $\SI{17.52}{\percent}$, a TOP3 error rate of $\SI{4.04}{\percent}$,
and a MER error rate of $\SI{3.96}{\percent}$. This was achieved by extending
the system $B_2$ by one algorithm or feature at a time, evaluating the
extended system and combining all changes into $B_{2,c}'$ which improved the
recognition rate.

The most important change was \gls{SLP} (see
\cpageref{subsec:pretraining-evaluation}). It improved the MER error by
$\SI{0.99}{\percent}$. Adding the re-curvature feature improved the systems MER
error by $\SI{0.57}{\percent}$, the ink feature improved it by
$\SI{0.53}{\percent}$, the stroke count feature by $\SI{0.29}{\percent}$ and the
aspect ratio feature by $\SI{0.07}{\percent}$. The stroke connect preprocessing
algorithm improved the MER error by $\SI{0.33}{\percent}$, but all other
preprocessing steps had either no effect that was bigger than random weight
initialization or even made the classifiers worse. Douglas-Peucker smoothing is
a preprocessing step that was not mentioned before in the literature for on-line
\gls{HWR}, but its evaluation results were much worse than the baseline system
for any simplification threshold $\varepsilon > 0.05$. The computational cost
of cubic spline interpolation is higher than linear interpolation and the
classification results are worse.


\chapter{Conclusion}\label{ch:Conclusion}
\section{Summary}
The aim of this bachelor's thesis was to build a recognition system that can
recognize many mathematical symbols with low error rates as well as to evaluate
which preprocessing steps and features help to improve the recognition rate.

All recognition systems were trained and evaluated with
$\num{\totalCollectedRecordings{}}$ recordings for \totalClassesAnalyzed{}
symbols. These recordings were collected by two crowdsourcing projects
(\href{http://detexify.kirelabs.org/classify.html}{Detexify} and
\href{write-math.com}{write-math.com}) and created with various devices. While
some recordings were created with standard touch devices such as tablets and
smartphones, others were created with the mouse.

\Glspl{MLP} were used for the classification task. Four baseline systems with
different numbers of hidden layers were used, as the number of hidden layer
influences the capabilities and problems of \glspl{MLP}. Furthermore, an error
measure MER was defined, which takes the top three \glspl{hypothesis} of the
classifier, merges symbols such as \verb+\sum+ ($\sum$) and \verb+\Sigma+
($\Sigma$) to equivalence classes, and then calculates the error.

All baseline systems used the same preprocessing queue. The recordings were
scaled to fit into a unit square, shifted to $(0,0)$, resampled with linear
interpolation so that every stroke had exactly 20~points which are spread
equidistant in time. The 80~($x,y$) coordinates of the first 4~strokes were
used to get exactly $160$ input features for every recording. The baseline
system $B_2$ has a MER error of $\SI{5.67}{\percent}$.

Three variations of the scale and shift algorithm, wild point filtering, stroke
connect, weighted average smoothing, and Douglas-Peucker smoothing were
evaluated. The evaluation showed that the scale and shift algorithm is
extremely important and the connect strokes algorithm improves the
classification. All other preprocessing algorithms either diminished the
classification performance or had less influence on it than the random
initialization of the \glspl{MLP} weights.

Adding two slightly rotated variants for each recording and hence tripling the
training set made the systems $B_3$ and $B_4$ perform much worse, but improved
the performance of the smaller systems.

The global features re-curvature, ink, stoke count and aspect ratio improved
the systems $B_1$--$B_3$, whereas the stroke center point feature made $B_2$
perform worse.

The learning rate and the momentum were evaluated. A learning rate of
$\eta=0.1$ and a momentum of $\alpha=0.9$ gave the best results. Newbob
training lead to much worse recognition rates. Denoising auto-encoders were
evaluated as one way to use pretraining, but by this the error rate increased
notably. However, supervised layer-wise pretraining improved the performance
decidedly.

The stroke connect algorithm was added to the preprocessing steps of the
baseline system as well as the re-curvature feature, the ink feature, the number
of strokes and the aspect ratio. The training setup of the baseline system was
changed to supervised layer-wise pretraining and the resulting model was trained
with a lower learning rate again. This optimized recognizer $B_{2,c}'$ had a MER
error of $\SI{3.96}{\percent}$. This means that the MER error dropped by over
$\SI{30}{\percent}$ in comparison to the baseline system $B_2$.

A MER error of $\SI{3.96}{\percent}$ makes the system usable for symbol lookup.
It could also be used as a starting point for the development of a
multiple-symbol classifier.

The aim of this bachelor's thesis was to develop a symbol recognition system
which is easy to use, fast and has high recognition rates as well as evaluating
ideas for single symbol classifiers. Some of those goals were reached. The
recognition system $B_{2,c}'$ evaluates new recordings in a fraction of a
second and has acceptable recognition rates. Many algorithms were evaluated.
However, there are still many other algorithms which could be evaluated and, at
the time of this work, the best classifier $B_{2,c}'$ is not publicly
available.

\section{Future Work}
The presented system for single-symbol recognition with \totalClassesAnalyzed{}
classes works well. However, there are still many other symbols that one might
want to classify, but which did not have enough training examples. This means
that one part of the future work will include collecting more training examples,
so that each class has at least 150 training examples.

Single-symbol recognition does already help \LaTeX{} users a lot, but multiple
symbol recognition is much more interesting. New hardware like
\href{http://www.isketchnote.com}{iSketchnote} or
\href{https://www.indiegogo.com/projects/equil-smartpen-2-real-ink-real-paper-digitized}{Equil~Smartpen~2}
can be used to improve the user experience of data input. It would be a
significant development if users could employ those improved input devices to
write complete formulas or systems of equations which a recognition system
would record, recognize, and optimize for the best typeset result.

User interviews and surveys should be made to see how users employ such
recognition systems, what they expect and if the system is useful for them. It
might especially be interesting to see which kind of input device is
comfortable for users and how the recorded data and the classification error
changes with different devices.

However, there is still a lot that could be tried for single-symbol recognition
with the mouse as an input device. Local features like bitmap environments
notably improved recognition rates in earlier work and should be evaluated
again and compared with the optimized recognizer $B_2'$. Different user
interfaces like the one shown in
\cref{fig:draft-single-symbol-context-interface} could be applied. Bottlenecks
could be added to the \gls{MLP} architecture. The $F_1$ score of preprocessing
steps could be calculated to learn optimal parameters for noise reduction.

Future work could also attempt to find recognition systems that have less
weights and similar recognition capabilities by inserting bottlenecks. If the
neural network becomes smaller, it could be possible to let users download it
in a JavaScript browser application and execute the classification directly on
the client with
\href{http://cs.stanford.edu/people/karpathy/convnetjs/index.html}{ConvNetJS}.

\cleardoublepage%
\phantomsection%
\addcontentsline{toc}{chapter}{\bibname}

\iflanguage{english}
{\bibliographystyle{IEEEtranSA}}	
{\bibliographystyle{babalpha-fl}}	

\bibliography{thesis}

\printglossaries%
\cleardoublepage%
\renewcommand{\chaptermark}[1]{\markboth{#1}{}}

\appendix

\iflanguage{english}
{\addchap{Appendix}}  
{\addchap{Anhang}}    

\section{Algorithms}\label{appendix:algorithms}
The following pseudo-code makes use of $0$-indexed lists and arrays.
The notation $abc[-1]$ means that the last element of the list or array $abc$
is accessed. The notation $abc[2:5]$ is called a \textit{slice} in Python.
It creates a new list from the list $abc$ that contains the elements with index
2, 3 and 4. The slice $abc[1:-1]$ means that a new list is created that contains
all elements except for the first of the list $abc$.

\begin{algorithm}[ht]
  \begin{algorithmic}
        \Function{resampling}{$pointlist$, $points\_per\_stroke$}
            \State $new\_pointlist \gets \Call{List}{~}$
            \ForAll{$stroke$ in $pointlist$}
                \State $new\_stroke \gets \Call{List}{~}$
                \If{\Call{count}{stroke} < 4}
                    \LineComment{Don't do anything if there are less than 4 points}
                    \State $new\_stroke \gets stroke$
                \Else
                    \State $x, y, t \gets \Call{List}{~}, \Call{List}{~}, \Call{List}{~}$
                    \ForAll{$point$ in $stroke$}
                        \State $x.\Call{append}{point['x']}$
                        \State $y.\Call{append}{point['y']}$
                        \State $t.\Call{append}{point['time']}$
                    \EndFor
                    \State $f_x \gets \Call{interpolate}{x, t}$ \Comment{This could be linear interpolation,}
                    \State $f_y \gets \Call{interpolate}{y, t}$ \Comment{cubic spline interpolation or something else}
                    \State $times \gets \Call{space\_linear}{t[0], t[-1]}$
                    \ForAll{$t$ in $times$}
                        \State $point \gets \Call{Point}{f_x(t), f_y(t), t}$
                        \State $new\_stroke.\Call{append}{point}$
                    \EndFor
                \EndIf
                \State $new\_pointlist.\Call{append}{new\_stroke}$
            \EndFor
            \Return $new\_pointlist$
        \EndFunction
  \end{algorithmic}
  \caption{Resampling}
  \label{alg:resampling}
\end{algorithm}

\begin{algorithm}[ht]
    \begin{algorithmic}
        \Function{scale\_and\_shift}{$pointlist$}
            \LineComment{Calculate bounding box}
            \State $min_x, min_y \gets pointlist[0][0]['x'], pointlist[0][0]['y']$
            \State $max_x, max_y \gets pointlist[0][0]['x'], pointlist[0][0]['y']$
            \State $min_t \gets pointlist[0][0]['t'], pointlist[0][0]['t']$
            \ForAll{$stroke$ in $pointlist$}
                \ForAll{$p$ in $stroke$}
                    \State $min_x \gets \Call{min}{min_x, p['x']}$
                    \State $max_x \gets \Call{max}{max_x, p['x']}$
                    \State $min_y \gets \Call{min}{min_y, p['y']}$
                    \State $max_y \gets \Call{max}{max_y, p['y']}$
                    \State $min_t \gets \Call{min}{min_t, p['t']}$
                \EndFor
            \EndFor

            \LineComment{Calculate parameters for scaling and shifting to $[-0.5, 0.5] \times [-0.5, 0.5]$}
            \State $width, height \gets max_x - min_x + 1, max_y - min_y + 1$
            \State $factor_x, factor_y \gets \frac{1}{width}, \frac{1}{height}$
            \State $factor \gets \min(factor_x, factor_y)$
            \State $add_x, add_y \gets \frac{width \cdot factor}{2}, \frac{height \cdot factor}{2}$

            \LineComment{Move every single point of a recording}
            \ForAll{$stroke$ in $pointlist$}
                \ForAll{$p$ in $stroke$}
                    \State $p['x'] \gets (p['x'] - min_x) \cdot factor - add_x$
                    \State $p['y'] \gets (p['y'] - min_y) \cdot factor - add_y$
                    \State $p['t'] \gets p['t'] - min_t$
                \EndFor
            \EndFor

            \Return $pointlist$
        \EndFunction
    \end{algorithmic}
\caption{Scale and shift a list of strokes to the $[-0.5, 0.5] \times [-0.5, 0.5]$ unit square}
\label{alg:scale-and-shift}
\end{algorithm}
\begin{algorithm}[ht]
  \begin{algorithmic}
        \Function{dot\_reduction}{$pointlist$, $\theta \in \mathbb{R}_{\geq 0}$}
            \State$new\_pointlist \gets \Call{List}{~}$
            \ForAll{$stroke$ in $pointlist$}
                \State$new\_stroke \gets stroke$
                \LineComment{Calculate maximum distance of two points in a stroke}
                \State$max\_distance \gets \Call{get\_max\_distance}{stroke}$
                \LineComment{Merge points of a stroke if the distance between them is below a threshold $\theta$}
                \If{$max\_distance < \theta$}
                    \State$p \gets \Call{get\_average\_point}{stroke}$
                    \State$new\_stroke \gets \Call{List}{p}$
                \EndIf%
                \State$new\_pointlist.\Call{append}{new\_stroke}$
            \EndFor%
            \Return$new\_pointlist$
        \EndFunction%
  \end{algorithmic}
  \caption{Dot reduction}
  \label{alg:dot-reduction}
\end{algorithm}
\begin{algorithm}[ht]
  \begin{algorithmic}
        \Function{dehook\_stroke}{$stroke$, $\theta \in \mathbb{R}_{\geq 0}$}
            \If{$\Call{count}{stroke} < 3$}
                \State\Return $stroke$
            \Else
                \State$new\_stroke \gets stroke[0: \Call{count}{stroke}-1]$ \Comment{Get everything but the last point}
                \State$stroke \gets stroke[\Call{count}{stroke}-3:]$ \Comment{get the last 3 points}
                \State$p \gets stroke[-1]$ \Comment{last point}
                \If{$\Call{calculate\_angle}{stroke} < \theta$}
                    \State$new\_stroke.\Call{append}{p}$
                \Else%
                    \State$new\_stroke \gets \Call{dehook\_stroke}{new\_stroke, \theta}$
                \EndIf%
            \EndIf%
            \Return$new\_pointlist$
        \EndFunction%
  \end{algorithmic}
  \caption{Dehooking}
  \label{alg:dehooking}
\end{algorithm}
\begin{algorithm}[ht]
  \begin{algorithmic}
        \Function{weighted\_average\_smoothing}{$pointlist$, $\theta = [\frac{1}{6}, \frac{4}{6}, \frac{1}{6}]$}
            \State $\theta \gets \frac{1}{\Call{sum}{\theta}} \cdot \theta$ \Comment{Normalize parameters to a sum of 1}
            \State $new\_pointlist \gets \Call{List}{~}$
            \ForAll{$stroke$ in $pointlist$}
                \State $tmp \gets \Call{List}{stroke[0]}$
                \State $new\_pointlist.\Call{append}{tmp}$
                \If{$\Call{count}{stroke} > 1$}
                    \For{$i \gets 1; i < \Call{count}{stroke}-1; i \gets i + 1$}
                        \State $p \gets \theta_0 \cdot stroke[i-1] + \theta_1 \cdot stroke[i] + \theta_2 \cdot stroke[i+1]$
                        \State $new\_pointlist[-1].\Call{append}{p}$
                    \EndFor
                    \State $new\_pointlist[-1].\Call{append}{stroke[-1]}$
                \EndIf
            \EndFor
            \Return $new\_pointlist$
        \EndFunction
  \end{algorithmic}
  \caption{Weighted average smoothing}
  \label{alg:weighted-average-smoothing}
\end{algorithm}
\begin{algorithm}[ht]
  \begin{algorithmic}
        \Function{douglas\_peucker}{$stroke$ as list of points, $\varepsilon \in \mathbb{R}_{\geq 0}$}
            \State$S \gets \Set{stroke[0], stroke[-1]}$
            \LineComment{Calculate point that is furthest away from line $(stroke[0], stroke[-1])$}
            \State$pi_{\max} \gets 0$ \Comment{Index of the point with highest distance}
            \State$d_{\max} \gets 0$ \Comment{Distance of the point with highest distance}
            \For{$i\gets1$; $i < \Call{count}{stroke} - 1$; $i \gets i+1$}
                \LineComment{Distance of the point $i$ to the line}
                \LineComment{defined by the first and the last point of stroke}
                \State$d \gets \Call{distance}{stroke[0], stroke[-1], stroke[i]}$
                \If{$d > d_{\max}$}
                    \State$d_{\max} \gets d$
                    \State$pi_{\max} \gets i$
                \EndIf%
            \EndFor%
            \LineComment{Recursively apply the algorithm}
            \If{$d_{\max} > \varepsilon$}
                \State$S_1 \gets \Call{douglas\_peucker}{stroke[0:pi_{\max}], \varepsilon}$
                \State$S_2 \gets \Call{douglas\_peucker}{stroke[pi_{\max}:-1], \varepsilon}$
                \State$S \gets S \cup S_1 \cup S_2$
            \EndIf%
            \State\Return$S$
        \EndFunction%
  \end{algorithmic}
  \caption{The Douglas-Peucker algorithm for stroke simplification.}
  \label{alg:douglas-peucker}
\end{algorithm}
\begin{algorithm}[ht]
  \begin{algorithmic}
    \State$a \gets \text{next from}~A$
    \State$b \gets \text{next from}~B$
    \State$d \gets \delta(a,b)$
    \State$a' \gets \text{next from}~A$
    \State$b' \gets \text{next from}~B$
    \While{points left in $A$ $\wedge$ points left in $B$}
      \State$l, m, r \gets \delta(a',b), \delta(a', b'), \delta(a, b')$
      \State$\mu \gets \min ~\{l, m, r\}$
      \State$d \gets d + \mu$
      \If{$l = \mu$}
        \State$a \gets a'$
        \State$a' \gets \text{next from}~A$
      \ElsIf{$r = \mu$}
        \State$b \gets b'$
        \State$b' \gets \text{next from}~B$
      \Else
        \State$a \gets a'$
        \State$b \gets b'$
        \State$a' \gets \text{next from}~A$
        \State$b' \gets \text{next from}~B$
      \EndIf%
    \EndWhile%
    \If{no points left in $A$}
      \ForAll{points $p$ in $B$}
        \State $d \gets d + \delta(a',p)$
      \EndFor%
    \ElsIf{no points left in $B$}
      \ForAll{points $p$ in $A$}
        \State $d \gets d + \delta(b',p)$
      \EndFor%
    \EndIf%
  \end{algorithmic}
  \caption{Greedy matching as described in \cite{Kirsch}}
\label{alg:greedy-matching}
\end{algorithm}
\clearpage

\section{Tables}\label{appendix:tables}

\begin{table}[ht]
    \centering
    \begin{tabular}{ll}
    \toprule
    \multicolumn{2}{l}{\textbf{Dot reduction}} \\\midrule
    Minimum distance threshold & $[0, \text{maximum point distance})$    \\[1ex]
    \multicolumn{2}{l}{\textbf{Wild point filter}} \\\midrule
    Maximum speed threshold    & $(0, \frac{\text{maximum point distance}}{\text{recording time}})$ \\[2ex]
    \multicolumn{2}{l}{\textbf{Dehook}} \\\midrule
    Maximum angle threshold    & $(0, 360]$                              \\[1ex]
    \multicolumn{2}{l}{\textbf{Smooth}} \\\midrule
    Smoothing factor $p_{i-1}$ & $[0, 1]$                           \\
    Smoothing factor $p_{i}$   & $[0, 1]$                           \\
    Smoothing factor $p_{i+1}$ & $[0, 1]$                           \\[1ex]
    \multicolumn{2}{l}{\textbf{Douglas Peucker}} \\\midrule
    Epsilon                    & $[0, \text{minimum point distance})$                           \\[1ex]
    \multicolumn{2}{l}{\textbf{Scale}} \\\midrule
    Size                       & $(0, \infty) \times (0, \infty)$        \\[1ex]
    \multicolumn{2}{l}{\textbf{Shift}} \\\midrule
    Center                     & $\Set{\text{true}, \text{false}}$       \\[1ex]
    Shift target               & $\mathbb{R}^2$                          \\[1ex]
    \multicolumn{2}{l}{\textbf{Stroke connect}} \\\midrule
    Minimum distance threshold & $[0, \text{maximum point distance})$    \\[1ex]
    \multicolumn{2}{l}{\textbf{Resample}} \\\midrule
    Type                       & $\Set{\text{linear}, \text{cubic}}$     \\
    Points per stroke          & $1, 2, \dots$                           \\
    \bottomrule
    \end{tabular}
    \caption {Preprocessing algorithms, their parameters and value ranges of
              those parameters. All of those algorithms are explained
              in \cref{sec:preprocessing}.}
    \label{table:preprocessing-summation}
\end{table}\clearpage
\begin{table}[ht]
    \centering
    \begin{tabular}{lc|lc}
        \toprule
        \multicolumn{2}{c}{Base symbol} & \multicolumn{2}{c}{equivalent symbols}\\
        \LaTeX         & Rendered       & \LaTeX                 & Rendered  \\\midrule
        \verb+\sum+    & $\sum$         & \verb+$\Sigma$+        & $\Sigma$\\
        \verb+\prod+   & $\prod$        & \verb+$\Pi$+           & $\Pi$\\
        ~              & ~              & \verb+$\sqcap$+        & $\sqcap$\\
        \verb+\coprod+ & $\coprod$      & \verb+$\amalg$+        & $\amalg$\\
        ~              & ~              & \verb+$\sqcup$+        & $\sqcup$\\
        \verb+\perp+   & $\perp$        & \verb+$\bot$+          & $\bot$\\
        \verb+\models+ & $\models$      & \verb+$\vDash$+        & $\vDash$\\
        \verb+|+       & |              & \verb+\mid+            & $\mid$  \\
        \verb+\Delta+  & $\Delta$       & \verb+$\triangle$+     & $\triangle$\\
        ~              & ~              & \verb+$\vartriangle$+  & $\vartriangle$\\
        \verb+\|+      & $\|$           & \verb+$\parallel$+     & $\parallel$\\
        \verb+\ohm+    & $\Omega$         & \verb+$\Omega$+      & $\Omega$\\
        \verb+\setminus+ & $\setminus$  & \verb+$\backslash$+    & $\backslash$\\
        \verb+\checked+ & {\mbox {\wasyfamily \char 8}} & \verb+$\checkmark$+    & $\checkmark$\\
        \verb+\&+      & $\&$           & \verb+$\with$+         & $\with$\\
        \verb+\#+      & $\#$           & \verb+$\sharp$+        & $\sharp$\\
        \verb+\S+      & $\S$           & \verb+$\mathsection$+  & $\mathsection$\\
        \verb+\nabla+  & $\nabla$       & \verb+\triangledown+   & $\triangledown$\\
        \verb+\lhd+    & $\lhd$         & \verb+$\triangleleft$+ & $\triangleleft$\\
        ~              & ~              & \verb+$\vartriangleleft$+ & $\vartriangleleft$\\
        \verb+\oiint+  & $\oiint$       & \verb+$\varoiint$+     & $\varoiint$\\
        \verb+\mathbb{R}+ & $\mathbb{R}$ & \verb+$\mathds{R}$+   & $\mathds{R}$\\
        \verb+\mathbb{Q}+ & $\mathbb{Q}$ & \verb+\mathds{Q}+     & $\mathds{Q}$\\
        \verb+\mathbb{Z}+ & $\mathbb{Z}$ & \verb+\mathds{Z}+     & $\mathds{Z}$\\
        \verb+\mathcal{A}+ & $\mathcal{A}$ & \verb+\mathscr{A}+  & $\mathscr{A}$\\
        \verb+\mathcal{D}+ & $\mathcal{D}$ & \verb+\mathscr{D}+  & $\mathscr{D}$\\
        \verb+\mathcal{N}+ & $\mathcal{N}$ & \verb+\mathscr{N}+  & $\mathscr{N}$\\
        \verb+\mathcal{R}+ & $\mathcal{R}$ & \verb+\mathscr{R}+  & $\mathscr{R}$\\
        \verb+\propto+ & $\propto$      & \verb+$\varpropto$+    & $\varpropto$\\
        \bottomrule
    \end{tabular}
    \caption{Symbols that cannot be distinguished in handwriting. Those symbols
             were used to define equivalence classes for an error measure MER
             which is introduced on \cpageref{ch:Evaluation}.}
    \label{table:difficult-symbols}
\end{table}


\begin{table}[ht]
    \centering
    \begin{tabular}{lc|lc}
        \toprule
        \LaTeX         & Rendered       & \LaTeX                 & Rendered  \\\midrule
        \verb+\alpha+  & $\alpha$       & \verb+\propto+         & $\propto$\\
        ~              & ~              & \verb+\ltimes+         & $\ltimes$\\
        \verb+0+       & 0              & \verb+O+               & O\\
        ~              & ~              & \verb+o+               & o\\
        ~              & ~              & \verb+\circ+           & $\circ$\\
        ~              & ~              & \verb+\degree+         & $^{\circ}$\\
        ~              & ~              & \verb+\fullmoon+       & {\mbox {\wasyfamily \char 35}}\\
        \verb+\epsilon+ & $\epsilon$    & \verb+$\varepsilon$+   & $\varepsilon$\\
        ~              & ~              & \verb+$\in$+           & $\in$\\
        ~              & ~              & \verb+$\mathcal{E}$+   & $\mathcal{E}$\\
        \verb+\Lambda+ & $\Lambda$      & \verb+$\wedge$+        & $\wedge$\\
        \verb+\emptyset+ & $\emptyset$  & \verb+\O+              & \O\\
        ~              & ~              & \verb+\o+              & \o\\
        ~              & ~              & \verb+$\diameter$+     & {\mbox {\wasyfamily \char 31}}\\
        ~              & ~              & \verb+$\varnothing$+   & $\varnothing$\\
        \verb+\rightarrow+ & $\rightarrow$ & \verb+$\longrightarrow$+   & $\longrightarrow$\\
        ~              & ~              & \verb+$\shortrightarrow$+ & $\shortrightarrow$\\
        \verb+\Rightarrow+ & $\Rightarrow$ & \verb+$\Longrightarrow$+   & $\Longrightarrow$\\
        \verb+\Leftrightarrow+ & $\Leftrightarrow$ & \verb+$\Longleftrightarrow$+   & $\Longleftrightarrow$\\
        \verb+\mapsto+ & $\mapsto$      & \verb+\longmapsto+     & $\longmapsto$\\
        \verb+\mathbb{1}+ & \includegraphics[height=12.3pt, keepaspectratio]{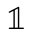}        & \verb+\mathds{1}+     & $\mathds{1}$\\
        \verb+\mathscr{L}+ & $\mathscr{L}$ & \verb+\mathcal{L}+  & $\mathcal{L}$\\
        \verb+\mathbb{Z}+ & $\mathbb{Z}$ & \verb+\mathcal{Z}+     & $\mathcal{Z}$\\
        \verb+\geq+    & $\geq$         & \verb+\geqslant+       & $\geqslant$\\
        ~              & ~              & \verb+\succeq+         & $\succeq$\\
        \verb+\leq+    & $\leq$         & \verb+\leqslant+       & $\leqslant$\\
        \verb+\pi+     & $\pi$          & \verb+\Pi+             & $\Pi$\\
        \verb+\psi+    & $\psi$         & \verb+\Psi+            & $\Psi$\\
        \verb+\phi+    & $\phi$         & \verb+\Phi+            & $\Phi$\\
        ~              & ~              & \verb+\emptyset+       & $\emptyset$\\
        \verb+\rho+    & $\rho$         & \verb+\varrho+         & $\varrho$\\
        \verb+\theta+  & $\theta$       & \verb+\Theta+          & $\Theta$\\
        \verb+\odot+   & $\odot$        & \verb+\astrosun+       & {\mbox {$\odot $}}\\
        \verb+\cdot+   & $\cdot$        & \verb+\bullet+         & $\bullet$\\
        \verb+x+       & $x$            & \verb+\times+          & $\times$\\
        ~              & ~              & \verb+X+               & $X$\\
        ~              & ~              & \verb+\chi+            & $\chi$\\
        ~              & ~              & \verb+\mathcal{X}+     & $\mathcal{X}$\\
        \verb+\beta+   & $\beta$        & \verb+\ss+             & \ss\\
        \verb+\male+   & {\mbox {\wasyfamily \char 26}} & \verb+\mars+ & {\leavevmode \lower 0.2ex\hbox {\wasyfamily \char 26}}\\
        \verb+\female+ & {\mbox {\wasyfamily \char 25}} & \verb+\venus+ & {\leavevmode \raise 0.2ex\hbox {\wasyfamily \char 25}}\\
        \verb+\bowtie+ & $\bowtie$      & \verb+\Bowtie+         & {\mbox {\wasyfamily \char 49}}\\
        \verb+\diamond+ & $\diamond$    & \verb+\diamondsuit+    & $\diamondsuit$\\
        ~              & ~              & \verb+\lozenge+        & $\lozenge$\\
        \verb+\dots+   & $\dots$        & \verb+\dotsc+          & $\dotsc$\\
        \verb+\mathcal{T}+ & $\mathcal{T}$ & \verb+\tau+         & $\tau$\\
        \bottomrule
    \end{tabular}
    \caption{Symbols that are extremely difficult to distinguish in handwriting (1).}
    \label{table:difficult-symbols-possible-but-difficult-1}
\end{table}

\begin{table}[ht]
    \centering
    \begin{tabular}{lc|lc}
        \toprule
        \LaTeX         & Rendered       & \LaTeX                 & Rendered  \\\midrule
        \verb+\mathcal{A}+ & $\mathcal{A}$ & \verb+A+               & $A$\\
        \verb+\mathcal{D}+ & $\mathcal{D}$ & \verb+D+               & $D$\\
        \verb+\mathcal{N}+ & $\mathcal{N}$ & \verb+N+               & $N$\\
        \verb+\mathcal{R}+ & $\mathcal{R}$ & \verb+R+               & $R$\\
        \verb+\varepsilon+ & $\varepsilon$ & \verb+\mathcal{E}+ & $\mathcal{E}$\\
        \bottomrule
    \end{tabular}
    \caption{Symbols that are extremely difficult to distinguish in handwriting (2).}
    \label{table:difficult-symbols-possible-but-difficult-2}
\end{table}


\begin{table}[ht]
    \centering
    \begin{tabular}{lc|lc}
        \toprule
        \LaTeX         & Rendered       & \LaTeX                 & Rendered  \\\midrule
        \verb+\dots+   & \dots          & \verb+\textellipsis+   & \textellipsis \\
        \verb+-+       & -              & \verb+\textendash+     & \textendash \\
        ~              & ~              & \verb+\textemdash+     & \textemdash \\
        ~              & ~              & \verb+\--+             & \-- \\
        ~              & ~              & \verb+\---+            & \--- \\
        ~              & ~              & \verb+\----+           & \---- \\
        \verb+\_+      & \_             & \verb+\textunderscore+ & \textunderscore \\
        \verb+i+       & i              & \verb+!`+              & !`\\
        ~              & ~              & \verb+\textexclamdown+ & \textexclamdown\\
        \verb+@+       & @              & \verb+$\MVAt$+         & $\MVAt$\\
        \verb+|+       & |              & \verb+\shortmid+       & \includegraphics[height=12.3pt, keepaspectratio]{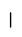} \\
        ~              & ~              & \verb+\textpipe+       & \includegraphics[height=12.3pt, keepaspectratio]{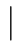} \\
        ~              & ~              & \verb+\textbar+        & \textbar \\
        \verb+\degree+ & {\ensuremath {^\circ }} & \verb+\textdegree+     & $^{\circ}$\\ 
        \bottomrule
    \end{tabular}
    \caption{\LaTeX{} symbols that were not evaluated, but also have confusion problems}
    \label{table:difficult-symbols-not-evaluated}
\end{table}\clearpage

\begin{table}[ht]
        \centering
            \begin{tabular}{lc|lc|lc}
                \toprule
                \multicolumn{2}{c}{Base symbol} & \multicolumn{2}{c}{\textbackslash{}n variant} & \multicolumn{2}{c}{\textbackslash{}not variant}\\
                \cmidrule{1-2}\cmidrule{3-4}\cmidrule{5-6}
                \LaTeX & Rendered & \LaTeX    & Rendered & \LaTeX      & Rendered \\
                \midrule
                \verb+=+ & $=$     & \verb+\neq+   & $\neq$ &\verb+\not=+ & $\not=$ \\
                \verb+\cong+ & $\cong$ & \verb+\ncong+ & $\ncong$ &\verb+\not\cong+ & $\not\cong$ \\
                \verb+\equiv+ & $\equiv$ & - & - &\verb+\not\equiv+ & $\not\equiv$ \\
                \verb+\in+ & $\in$ & \verb+\notin+ & $\notin$ &\verb+\not\in+ & $\not\in$ \\
                \verb+\vDash+ & $\vDash$ & \verb+\nvDash+ & $\nvDash$ &\verb+\not\vDash+ & $\not\vDash$ \\
                \verb+\mid+ & $\mid$ & \verb+\nmid+ & $\nmid$ &\verb+\not\mid+ & $\not\mid$ \\
                \verb+\exists+ & $\exists$ & \verb+\nexists+ & $\nexists$ &\verb+\not\exists+ & $\not\exists$ \\
                \verb+\subseteq+ & $\subseteq$ & \verb+\nsubseteq+ & $\nsubseteq$ &\verb+\not\subseteq+ & $\not\subseteq$ \\
                \verb+\rightarrow+ & $\rightarrow$ & \verb+\nrightarrow+ & $\nrightarrow$ &\verb+\not\rightarrow+ & $\not\rightarrow$ \\
                \verb+\Rightarrow+ & $\Rightarrow$ & \verb+\nRightarrow+ & $\nRightarrow$ &\verb+\not\Rightarrow+ & $\not\Rightarrow$ \\
        \bottomrule
    \end{tabular}

    \caption{\textbackslash{}n and \textbackslash{}not variants of symbols.}
    \label{table:not variants}
\end{table}

\begin{table}[H]
    \centering
    \begin{tabular}{crr|crr}
    \toprule
    Symbol                     & Mean & $\sigma$ & Symbol       & Mean & $\sigma$ \\\midrule
    \verb+\blacksquare+        & 9.22 & 3.86 & \verb+\male+               & 3.31 & 0.55 \\
    \verb+\blacktriangleright+ & 6.86 & 2.25 & \verb+\parr+               & 3.30 & 0.47 \\
    \verb+\bullet+             & 6.63 & 4.05 & \verb+\mathfrak{X}+        & 3.30 & 0.36 \\
    \verb+\boxtimes+           & 5.64 & 0.70 & \verb+\leftmoon+           & 3.29 & 0.52 \\
    \verb+\circledast+         & 5.27 & 0.70 & \verb+\sun+                & 3.28 & 0.68 \\
    \verb+\boxplus+            & 5.19 & 0.50 & \verb+\mathfrak{S}+        & 3.27 & 0.52 \\
    \verb+\otimes+             & 4.89 & 0.52 & \verb+\mathds{P}+          & 3.25 & 0.46 \\
    \verb+\circledR+           & 4.77 & 0.60 & \verb+\notin+              & 3.15 & 0.53 \\
    \verb+\oplus+              & 4.62 & 0.47 & \verb+\mars+               & 3.13 & 0.40 \\
    \verb+\star+               & 4.28 & 1.21 & \verb+\fullmoon+           & 3.05 & 0.25 \\
    \verb+\circledcirc+        & 4.28 & 0.60 & \verb+\degree+             & 3.04 & 0.33 \\
    \verb+\clubsuit+           & 4.20 & 1.83 & \verb+\mathds{1}+          & 3.01 & 0.67 \\
    \verb+\mathbb{Q}+          & 4.14 & 0.41 & \verb+\cong+               & 2.96 & 0.29 \\
    \verb+\oiint+              & 4.12 & 0.55 & \verb+\mathds{C}+          & 2.95 & 0.47 \\
    \verb+\copyright+          & 4.07 & 0.45 & \verb+\female+             & 2.68 & 0.55 \\
    \verb+\Bowtie+             & 3.92 & 0.46 & \verb+\venus+              & 2.64 & 0.32 \\
    \verb+\mathds{Q}+          & 3.89 & 0.53 & \verb+\ohm+                & 2.49 & 0.50 \\
    \verb+\mathfrak{M}+        & 3.84 & 0.89 & \verb+\celsius+            & 2.42 & 0.37 \\
    \verb+\mathds{R}+          & 3.83 & 0.50 & \verb+\sqrt{}+             & 1.89 & 0.38 \\
    \verb+\mathds{E}+          & 3.82 & 0.63 & \verb+\checked+            & 1.59 & 0.19 \\
    \verb+\mathds{Z}+          & 3.80 & 0.58 & \verb+\cdot+               & 0.77 & 2.19 \\
    \verb+\mathds{N}+          & 3.78 & 0.50 & \verb+\therefore+          & 0.46 & 1.04 \\
    \verb+\mathfrak{A}+        & 3.51 & 0.54 & \verb+\because+            & 0.36 & 1.01 \\
    \verb+\astrosun+           & 3.39 & 0.61 & \verb+\dotsc+              & 0.06 & 0.30 \\
    \bottomrule
    \end{tabular}
    \caption{Mean and standard deviation $\sigma$ of the ink feature of symbols
             that are not shown in \cref{fig:ink-mean-std-deviation}.}
    \label{table:ink-cut-out}
\end{table}
\begin{table}[H]
    \centering
    \begin{tabular}{crr|crr}
    \toprule
    Symbol                     & Mean    & $\sigma$ & Symbol                         & Mean    & $\sigma$ \\\midrule
    \verb+\sun+                & 9.46    & 2.18     &     \verb+\mathfrak{M}+        & 2.17    & 1.72 \\
    \verb+\mathds{E}+          & 4.78    & 1.56     &     \verb+\astrosun+           & 2.17    & 0.58 \\
    \verb+\mathds{1}+          & 3.40    & 0.85     &     \verb+\celsius+            & 2.05    & 0.29 \\
    \verb+\female+             & 3.35    & 1.88     &     \verb+\copyright+          & 2.03    & 0.22 \\
    \verb+\mathfrak{X}+        & 3.26    & 1.60     &     \verb+\diameter+           & 2.01    & 0.29 \\
    \verb+\male+               & 3.25    & 3.73     &     \verb+\mathfrak{A}+        & 1.45    & 0.92 \\
    \verb+\mathds{P}+          & 3.17    & 0.62     &     \verb+\Bowtie+             & 1.18    & 0.60 \\
    \verb+\mathds{Q}+          & 3.15    & 2.73     &     \verb+\leftmoon+           & 1.15    & 0.36 \\
    \verb+\mathds{R}+          & 3.15    & 0.77     &     \verb+\ohm+                & 1.05    & 0.29 \\
    \verb+\cong+               & 3.04    & 0.30     &     \verb+\mathfrak{S}+        & 1.05    & 0.27 \\
    \verb+\venus+              & 3.02    & 0.22     &     \verb+\parr+               & 1.05    & 0.22 \\
    \verb+\mathds{N}+          & 3.00    & 1.03     &     \verb+\sqrt{}+             & 1.04    & 0.29 \\
    \verb+\mars+               & 2.70    & 1.04     &     \verb+\checked+            & 1.04    & 0.22 \\
    \verb+\dotsc+              & 2.57    & 1.13     &     \verb+\degree+             & 1.03    & 0.16 \\
    \verb+\mathds{Z}+          & 2.19    & 0.78     &     \verb+\fullmoon+           & 1.02    & 0.14 \\
    \bottomrule
    \end{tabular}
    \caption{Mean and standard deviation $\sigma$ of the stroke count feature of
             symbols that are not shown in \cref{fig:stroke-count-mean-std-deviation}.}
    \label{table:stroke-count-cut-out}
\end{table}
\begin{table}[H]
    \centering
    \begin{tabular}{crr|crr}
    \toprule
    Symbol                      & Mean  & $\sigma$ & Symbol                    & Mean  & $\sigma$ \\\midrule
    \verb+-+                    & 33.09 & 31.82  & \verb+\mathfrak{S}+         & 1.04  & 0.25 \\
    \verb+\dots+                & 21.35 & 23.35  & \verb+\copyright+           & 1.04  & 0.19 \\
    \verb+\dotsc+               & 20.15 & 27.43  & \verb+\celsius+             & 1.01  & 0.25 \\
    \verb+\rightharpoonup+      & 5.06  & 2.28   & \verb+\diameter+            & 0.99  & 0.29 \\
    \verb+\multimap+            & 4.43  & 1.90   & \verb+\mars+                & 0.99  & 0.20 \\
    \verb+\longrightarrow+      & 4.33  & 2.27   & \verb+\mathfrak{A}+         & 0.97  & 0.33 \\
    \verb+\frown+               & 3.65  & 1.71   & \verb+\mathds{Q}+           & 0.96  & 0.21 \\
    \verb+\twoheadrightarrow+   & 3.60  & 1.31   & \verb+\mathfrak{X}+         & 0.93  & 0.26 \\
    \verb+\rightsquigarrow+     & 3.44  & 1.15   & \verb+\male+                & 0.92  & 0.40 \\
    \verb+\sim+                 & 3.41  & 1.16   & \verb+\mathds{C}+           & 0.90  & 0.21 \\
    \verb+\leadsto+             & 3.32  & 1.12   & \verb+\mathds{E}+           & 0.84  & 0.23 \\
    \verb+\ohm+                 & 1.61  & 0.53   & \verb+\parr+                & 0.75  & 0.15 \\
    \verb+\cong+                & 1.52  & 0.41   & \verb+\mathds{1}+           & 0.72  & 0.24 \\
    \verb+\sqrt{}+              & 1.50  & 0.56   & \verb+\mathds{N}+           & 0.72  & 0.18 \\
    \verb+\mathfrak{M}+         & 1.28  & 0.35   & \verb+\female+              & 0.71  & 0.38 \\
    \verb+\mathds{Z}+           & 1.18  & 0.29   & \verb+\mathds{R}+           & 0.71  & 0.17 \\
    \verb+\Bowtie+              & 1.15  & 0.33   & \verb+\mathds{P}+           & 0.64  & 0.21 \\
    \verb+\checked+             & 1.13  & 0.56   & \verb+\venus+               & 0.60  & 0.13 \\
    \bottomrule
    \end{tabular}
    \caption{Mean and standard deviation of the aspect ratio of symbols that are
             not shown in \cref{fig:aspect-ratio-mean-std-deviation}.}
    \label{table:aspect-cut-out}
\end{table}\clearpage
\subsection{Evaluated Symbols}
            \begin{longtable}{lc|lc}
                \toprule
                \LaTeX & Rendered & \LaTeX & Rendered \\
                \midrule
                \endhead
                \hline \multicolumn{4}{r}{{Continued on next page}} \\
                \endfoot
                \bottomrule
                \caption{112 symbols that were used for evaluation.}
                \endlastfoot
\verb+\&+ & $\&$ &\verb+\nmid+ & $\nmid$\\
\verb+\Im+ & $\Im$ &\verb+\nvDash+ & $\nvDash$\\
\verb+\Re+ & $\Re$ &\verb+\int+ & $\int$\\
\verb+\S+ & $\S$ &\verb+\fint+ & $\fint$\\
\verb+\Vdash+ & $\Vdash$ &\verb+\odot+ & $\odot$\\
\verb+\aleph+ & $\aleph$ &\verb+\oiint+ & $\oiint$\\
\verb+\amalg+ & $\amalg$ &\verb+\oint+ & $\oint$\\
\verb+\angle+ & $\angle$ &\verb+\varoiint+ & $\varoiint$\\
\verb+\ast+ & $\ast$ &\verb+\ominus+ & $\ominus$\\
\verb+\asymp+ & $\asymp$ &\verb+\oplus+ & $\oplus$\\
\verb+\backslash+ & $\backslash$ &\verb+\otimes+ & $\otimes$\\
\verb+\between+ & $\between$ &\verb+\parallel+ & $\parallel$\\
\verb+\blacksquare+ & $\blacksquare$ &\verb+\parr+ & $\parr$\\
\verb+\blacktriangleright+ & $\blacktriangleright$ &\verb+\partial+ & $\partial$\\
\verb+\bot+ & $\bot$ &\verb+\perp+ & $\perp$\\
\verb+\bowtie+ & $\bowtie$ &\verb+\pitchfork+ & $\pitchfork$\\
\verb+\boxdot+ & $\boxdot$ &\verb+\pm+ & $\pm$\\
\verb+\boxplus+ & $\boxplus$ &\verb+\prime+ & $\prime$\\
\verb+\boxtimes+ & $\boxtimes$ &\verb+\prod+ & $\prod$\\
\verb+\bullet+ & $\bullet$ &\verb+\propto+ & $\propto$\\
\verb+\checkmark+ & $\checkmark$ &\verb+\rangle+ & $\rangle$\\
\verb+\circ+ & $\circ$ &\verb+\rceil+ & $\rceil$\\
\verb+\circledR+ & $\circledR$ &\verb+\rfloor+ & $\rfloor$\\
\verb+\circledast+ & $\circledast$ &\verb+\rrbracket+ & $\rrbracket$\\
\verb+\circledcirc+ & $\circledcirc$ &\verb+\rtimes+ & $\rtimes$\\
\verb+\clubsuit+ & $\clubsuit$ &\verb+\sharp+ & $\sharp$\\
\verb+\coprod+ & $\coprod$ &\verb+\sphericalangle+ & $\sphericalangle$\\
\verb+\copyright+ & $\copyright$ &\verb+\sqcap+ & $\sqcap$\\
\verb+\dag+ & $\dag$ &\verb+\sqcup+ & $\sqcup$\\
\verb+\dashv+ & $\dashv$ &\verb+\sqrt{}+ & $\sqrt{}$\\
\verb+\diamond+ & $\diamond$ &\verb+\square+ & $\square$\\
\verb+\diamondsuit+ & $\diamondsuit$ &\verb+\star+ & $\star$\\
\verb+\div+ & $\div$ &\verb+\sum+ & $\sum$\\
\verb+\ell+ & $\ell$ &\verb+\times+ & $\times$\\
\verb+\flat+ & $\flat$ &\verb+\top+ & $\top$\\
\verb+\frown+ & $\frown$ &\verb+\triangle+ & $\triangle$\\
\verb+\guillemotleft+ & $\guillemotleft$ &\verb+\triangledown+ & $\triangledown$\\
\verb+\hbar+ & $\hbar$ &\verb+\triangleleft+ & $\triangleleft$\\
\verb+\heartsuit+ & $\heartsuit$ &\verb+\trianglelefteq+ & $\trianglelefteq$\\
\verb+\infty+ & $\infty$ &\verb+\triangleq+ & $\triangleq$\\
\verb+\langle+ & $\langle$ &\verb+\triangleright+ & $\triangleright$\\
\verb+\lceil+ & $\lceil$ &\verb+\uplus+ & $\uplus$\\
\verb+\lfloor+ & $\lfloor$ &\verb+\vDash+ & $\vDash$\\
\verb+\lhd+ & $\lhd$ &\verb+\varnothing+ & $\varnothing$\\
\verb+\lightning+ & $\lightning$ &\verb+\varpropto+ & $\varpropto$\\
\verb+\llbracket+ & $\llbracket$ &\verb+\vartriangle+ & $\vartriangle$\\
\verb+\lozenge+ & $\lozenge$ &\verb+\vdash+ & $\vdash$\\
\verb+\ltimes+ & $\ltimes$ &\verb+\with+ & $\with$\\
\verb+\mathds{1}+ & $\mathds{1}$ &\verb+\wp+ & $\wp$\\
\verb+\mathsection+ & $\mathsection$ &\verb+\wr+ & $\wr$\\
\verb+\mid+ & $\mid$ &\verb+\{+ & $\{$\\
\verb+\models+ & $\models$ &\verb+\|+ & $\|$\\
\verb+\mp+ & $\mp$ &\verb+\}+ & $\}$\\
\verb+\multimap+ & $\multimap$ &\verb+\vee+ & $\vee$\\
\verb+\nabla+ & $\nabla$ &\verb+\wedge+ & $\wedge$\\
\verb+\neg+ & $\neg$ &\verb+\barwedge+ & $\barwedge$
    \label{table:symbols-used-for-evaluation-0}
    \end{longtable}

\begin{table}[ht]
        \centering

            \begin{tabular}{lc|lc|lc|lc}
                \toprule
                \LaTeX & Rendered & \LaTeX & Rendered & \LaTeX & Rendered & \LaTeX & Rendered \\
                \midrule
\verb+\#+ & $\#$ &\verb+A+ & $A$ &\verb+S+ & $S$ &\verb+i+ & $i$\\
\verb+\$+ & $\$$ &\verb+B+ & $B$ &\verb+T+ & $T$ &\verb+j+ & $j$\\
\verb+\%+ & $\%$ &\verb+C+ & $C$ &\verb+U+ & $U$ &\verb+k+ & $k$\\
\verb+++ & $+$ &\verb+D+ & $D$ &\verb+V+ & $V$ &\verb+l+ & $l$\\
\verb+-+ & $-$ &\verb+E+ & $E$ &\verb+W+ & $W$ &\verb+m+ & $m$\\
\verb+/+ & $/$ &\verb+F+ & $F$ &\verb+X+ & $X$ &\verb+n+ & $n$\\
\verb+0+ & $0$ &\verb+G+ & $G$ &\verb+Y+ & $Y$ &\verb+o+ & $o$\\
\verb+1+ & $1$ &\verb+H+ & $H$ &\verb+Z+ & $Z$ &\verb+p+ & $p$\\
\verb+2+ & $2$ &\verb+I+ & $I$ &\verb+[+ & $[$ &\verb+q+ & $q$\\
\verb+3+ & $3$ &\verb+J+ & $J$ &\verb+]+ & $]$ &\verb+r+ & $r$\\
\verb+4+ & $4$ &\verb+K+ & $K$ &\verb+a+ & $a$ &\verb+s+ & $s$\\
\verb+5+ & $5$ &\verb+L+ & $L$ &\verb+b+ & $b$ &\verb+u+ & $u$\\
\verb+6+ & $6$ &\verb+M+ & $M$ &\verb+c+ & $c$ &\verb+v+ & $v$\\
\verb+7+ & $7$ &\verb+N+ & $N$ &\verb+d+ & $d$ &\verb+w+ & $w$\\
\verb+8+ & $8$ &\verb+O+ & $O$ &\verb+e+ & $e$ &\verb+x+ & $x$\\
\verb+9+ & $9$ &\verb+P+ & $P$ &\verb+f+ & $f$ &\verb+y+ & $y$\\
\verb+<+ & $<$ &\verb+Q+ & $Q$ &\verb+g+ & $g$ &\verb+z+ & $z$\\
\verb+>+ & $>$ &\verb+R+ & $R$ &\verb+h+ & $h$ &\verb+|+ & $|$\\

        \bottomrule
    \end{tabular}

    \caption{72 ASCII symbols that were used for evaluation, including all
             ten digits, the Latin alphabet in lower and upper case and
             a few more symbols.}
    \label{table:symbols-used-for-evaluation-1}
\end{table}

\begin{table}[ht]
        \centering
            \begin{tabular}{lc|lc|lc}
                \toprule
                \LaTeX & Rendered & \LaTeX & Rendered & \LaTeX & Rendered \\
                \midrule
\verb+\approx+ & $\approx$ &\verb+\geqslant+ & $\geqslant$ &\verb+\lesssim+ & $\lesssim$\\
\verb+\doteq+ & $\doteq$ &\verb+\neq+ & $\neq$ &\verb+\backsim+ & $\backsim$\\
\verb+\simeq+ & $\simeq$ &\verb+\not\equiv+ & $\not\equiv$ &\verb+\sim+ & $\sim$\\
\verb+\equiv+ & $\equiv$ &\verb+\preccurlyeq+ & $\preccurlyeq$ &\verb+\succ+ & $\succ$\\
\verb+\geq+ & $\geq$ &\verb+\preceq+ & $\preceq$ &\verb+\prec+ & $\prec$\\
\verb+\leq+ & $\leq$ &\verb+\succeq+ & $\succeq$ &\verb+\gtrless+ & $\gtrless$\\
\verb+\leqslant+ & $\leqslant$ &\verb+\gtrsim+ & $\gtrsim$ &\verb+\cong+ & $\cong$\\
        \bottomrule
    \end{tabular}

    \caption{21 symbols that were used for evaluation and indicate a
             relationship.}
    \label{table:symbols-used-for-evaluation-2}
\end{table}

\begin{table}[ht]
        \centering

            \begin{tabular}{lc|lc}
                \toprule
                \LaTeX & Rendered & \LaTeX & Rendered \\
                \midrule
\verb+\Downarrow+ & $\Downarrow$ &\verb+\nrightarrow+ & $\nrightarrow$\\
\verb+\Leftarrow+ & $\Leftarrow$ &\verb+\rightarrow+ & $\rightarrow$\\
\verb+\Leftrightarrow+ & $\Leftrightarrow$ &\verb+\rightleftarrows+ & $\rightleftarrows$\\
\verb+\Longleftrightarrow+ & $\Longleftrightarrow$ &\verb+\rightrightarrows+ & $\rightrightarrows$\\
\verb+\Longrightarrow+ & $\Longrightarrow$ &\verb+\rightsquigarrow+ & $\rightsquigarrow$\\
\verb+\Rightarrow+ & $\Rightarrow$ &\verb+\searrow+ & $\searrow$\\
\verb+\circlearrowleft+ & $\circlearrowleft$ &\verb+\shortrightarrow+ & $\shortrightarrow$\\
\verb+\circlearrowright+ & $\circlearrowright$ &\verb+\twoheadrightarrow+ & $\twoheadrightarrow$\\
\verb+\curvearrowright+ & $\curvearrowright$ &\verb+\uparrow+ & $\uparrow$\\
\verb+\downarrow+ & $\downarrow$ &\verb+\rightharpoonup+ & $\rightharpoonup$\\
\verb+\hookrightarrow+ & $\hookrightarrow$ &\verb+\rightleftharpoons+ & $\rightleftharpoons$\\
\verb+\leftarrow+ & $\leftarrow$ &\verb+\longmapsto+ & $\longmapsto$\\
\verb+\leftrightarrow+ & $\leftrightarrow$ &\verb+\mapsfrom+ & $\mapsfrom$\\
\verb+\longrightarrow+ & $\longrightarrow$ &\verb+\mapsto+ & $\mapsto$\\
\verb+\nRightarrow+ & $\nRightarrow$ &\verb+\leadsto+ & $\leadsto$\\
\verb+\nearrow+ & $\nearrow$ &\verb+\upharpoonright+ & $\upharpoonright$\\

        \bottomrule
    \end{tabular}

    \caption{32 arrow symbols that were used for evaluation.}
    \label{table:symbols-used-for-evaluation-3}
\end{table}

\begin{table}[ht]
        \centering

            \begin{tabular}{lc|lc|lc}
                \toprule
                \LaTeX & Rendered & \LaTeX & Rendered & \LaTeX & Rendered \\
                \midrule
\verb+\alpha+ & $\alpha$ &\verb+\xi+ & $\xi$ &\verb+\Xi+ & $\Xi$\\
\verb+\beta+ & $\beta$ &\verb+\pi+ & $\pi$ &\verb+\Pi+ & $\Pi$\\
\verb+\gamma+ & $\gamma$ &\verb+\rho+ & $\rho$ &\verb+\Sigma+ & $\Sigma$\\
\verb+\delta+ & $\delta$ &\verb+\sigma+ & $\sigma$ &\verb+\Phi+ & $\Phi$\\
\verb+\epsilon+ & $\epsilon$ &\verb+\tau+ & $\tau$ &\verb+\Psi+ & $\Psi$\\
\verb+\zeta+ & $\zeta$ &\verb+\phi+ & $\phi$ &\verb+\Omega+ & $\Omega$\\
\verb+\eta+ & $\eta$ &\verb+\chi+ & $\chi$ &\verb+\varepsilon+ & $\varepsilon$\\
\verb+\theta+ & $\theta$ &\verb+\psi+ & $\psi$ &\verb+\varkappa+ & $\varkappa$\\
\verb+\iota+ & $\iota$ &\verb+\omega+ & $\omega$ &\verb+\varpi+ & $\varpi$\\
\verb+\kappa+ & $\kappa$ &\verb+\Gamma+ & $\Gamma$ &\verb+\varrho+ & $\varrho$\\
\verb+\lambda+ & $\lambda$ &\verb+\Delta+ & $\Delta$ &\verb+\varphi+ & $\varphi$\\
\verb+\mu+ & $\mu$ &\verb+\Theta+ & $\Theta$ &\verb+\vartheta+ & $\vartheta$\\
\verb+\nu+ & $\nu$ &\verb+\Lambda+ & $\Lambda$ &\verb+ + & $ $\\

        \bottomrule
    \end{tabular}

    \caption{All Greek letters and some variations of Greek letters were
             used for evaluation. 38 of them are in this table, the rest
             is identical to Latin letters.}
    \label{table:symbols-used-for-evaluation-4}
\end{table}

\begin{table}[ht]
        \centering

            \begin{tabular}{lc|lc|lc}
                \toprule
                \LaTeX & Rendered & \LaTeX & Rendered & \LaTeX & Rendered \\
                \midrule
\verb+\mathcal{A}+ & $\mathcal{A}$ &\verb+\mathcal{T}+ & $\mathcal{T}$ &\verb+\mathds{Z}+ & $\mathds{Z}$\\
\verb+\mathcal{B}+ & $\mathcal{B}$ &\verb+\mathcal{U}+ & $\mathcal{U}$ &\verb+\mathfrak{A}+ & $\mathfrak{A}$\\
\verb+\mathcal{C}+ & $\mathcal{C}$ &\verb+\mathcal{X}+ & $\mathcal{X}$ &\verb+\mathfrak{M}+ & $\mathfrak{M}$\\
\verb+\mathcal{D}+ & $\mathcal{D}$ &\verb+\mathcal{Z}+ & $\mathcal{Z}$ &\verb+\mathfrak{S}+ & $\mathfrak{S}$\\
\verb+\mathcal{E}+ & $\mathcal{E}$ &\verb+\mathbb{H}+ & $\mathbb{H}$ &\verb+\mathfrak{X}+ & $\mathfrak{X}$\\
\verb+\mathcal{F}+ & $\mathcal{F}$ &\verb+\mathbb{N}+ & $\mathbb{N}$ &\verb+\mathscr{A}+ & $\mathscr{A}$\\
\verb+\mathcal{G}+ & $\mathcal{G}$ &\verb+\mathbb{Q}+ & $\mathbb{Q}$ &\verb+\mathscr{C}+ & $\mathscr{C}$\\
\verb+\mathcal{H}+ & $\mathcal{H}$ &\verb+\mathbb{R}+ & $\mathbb{R}$ &\verb+\mathscr{D}+ & $\mathscr{D}$\\
\verb+\mathcal{L}+ & $\mathcal{L}$ &\verb+\mathbb{Z}+ & $\mathbb{Z}$ &\verb+\mathscr{E}+ & $\mathscr{E}$\\
\verb+\mathcal{M}+ & $\mathcal{M}$ &\verb+\mathds{C}+ & $\mathds{C}$ &\verb+\mathscr{F}+ & $\mathscr{F}$\\
\verb+\mathcal{N}+ & $\mathcal{N}$ &\verb+\mathds{E}+ & $\mathds{E}$ &\verb+\mathscr{H}+ & $\mathscr{H}$\\
\verb+\mathcal{O}+ & $\mathcal{O}$ &\verb+\mathds{N}+ & $\mathds{N}$ &\verb+\mathscr{L}+ & $\mathscr{L}$\\
\verb+\mathcal{P}+ & $\mathcal{P}$ &\verb+\mathds{P}+ & $\mathds{P}$ &\verb+\mathscr{P}+ & $\mathscr{P}$\\
\verb+\mathcal{R}+ & $\mathcal{R}$ &\verb+\mathds{Q}+ & $\mathds{Q}$ &\verb+\mathscr{S}+ & $\mathscr{S}$\\
\verb+\mathcal{S}+ & $\mathcal{S}$ &\verb+\mathds{R}+ & $\mathds{R}$ &\verb+ + & $ $\\

        \bottomrule
    \end{tabular}

    \caption{44 variants of Latin letters were used for evaluation.}
    \label{table:symbols-used-for-evaluation-5}
\end{table}

\begin{table}[ht]
        \centering

            \begin{tabular}{lc|lc|lc}
                \toprule
                \LaTeX & Rendered & \LaTeX & Rendered & \LaTeX & Rendered \\
                \midrule
\verb+\therefore+ & $\therefore$ &\verb+\cdot+  & $\cdot$  &\verb+\dots+  & $\dots$\\
\verb+\because+   & $\because$   &\verb+\vdots+ & $\vdots$ &\verb+\ddots+ & $\ddots$\\
\verb+\dotsc+     & $\dotsc$     &\verb+ +      &          &\verb+ +      &  \\
        \bottomrule
    \end{tabular}

    \caption{7 symbols that contain only dots were used for evaluation.}
    \label{table:symbols-used-for-evaluation-6}
\end{table}

\begin{table}[ht]
        \centering
            \begin{tabular}{lc|lc|lc|lc|lc}
                \toprule
                \LaTeX & R & \LaTeX & R & \LaTeX & R &  \LaTeX & R & \LaTeX & R \\
                \midrule
\verb+\AA+ & {\r A}                                                           &\verb+\L+       & \includegraphics[height=12.3pt, keepaspectratio]{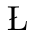} &\verb+\male+     & {\mbox {\wasyfamily \char 26}}                         &\verb+\ohm+      &  $\Omega $                     &\verb+\sun+       & {\mbox {\wasyfamily \char 46}} \\
\verb+\AE+ & \includegraphics[height=12.3pt, keepaspectratio]{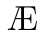} &\verb+\O+       & \includegraphics[height=12.3pt, keepaspectratio]{symbols/L.pdf} &\verb+\mars+     & {\leavevmode \lower 0.2ex\hbox {\wasyfamily \char 26}} &\verb+\fullmoon+ & {\mbox {\wasyfamily \char 35}} &\verb+\degree+    & {\ensuremath {^\circ }}\\
\verb+\aa+ & {\r a}                                                           &\verb+\o+       & \includegraphics[height=12.3pt, keepaspectratio]{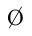} &\verb+\female+   & {\mbox {\wasyfamily \char 25}}                         &\verb+\leftmoon+ & {\mbox {\wasyfamily \char 36}} &\verb+\iddots+    & \includegraphics[height=12.3pt, keepaspectratio]{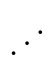}\\
\verb+\ae+ & \includegraphics[height=12.3pt, keepaspectratio]{symbols/AE.pdf} &\verb+\Bowtie+  & {\mbox {\wasyfamily \char 49}}                                  &\verb+\venus+    & {\leavevmode \raise 0.2ex\hbox {\wasyfamily \char 25}} &\verb+\checked+  & {\mbox {\wasyfamily \char 8}}  &\verb+\diameter+  & {\mbox {\wasyfamily \char 31}} \\
\verb+\ss+ & \includegraphics[height=12.3pt, keepaspectratio]{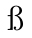} &\verb+\celsius+ & $^\circ \mathrm {C}$                                            &\verb+\astrosun+ & {\mbox {$\odot $}}                                     &\verb+\pounds+   & \textsterling                  &\verb+\mathbb{1}+ & \includegraphics[height=12.3pt, keepaspectratio]{symbols/mathbb-1.pdf}\\
        \bottomrule
    \end{tabular}
    \caption{25 symbols that were used for evaluation.}
    \label{table:symbols-used-for-evaluation-7}
\end{table}

\begin{table}[ht]
        \centering

            \begin{tabular}{lc|lc|lc}
                \toprule
                \LaTeX & Rendered & \LaTeX & Rendered & \LaTeX & Rendered \\
                \midrule
\verb+\cup+ & $\cup$ &\verb+\varsubsetneq+ & $\varsubsetneq$ &\verb+\exists+ & $\exists$\\
\verb+\cap+ & $\cap$ &\verb+\nsubseteq+ & $\nsubseteq$ &\verb+\nexists+ & $\nexists$\\
\verb+\emptyset+ & $\emptyset$ &\verb+\sqsubseteq+ & $\sqsubseteq$ &\verb+\forall+ & $\forall$\\
\verb+\setminus+ & $\setminus$ &\verb+\subseteq+ & $\subseteq$ &\verb+\in+ & $\in$\\
\verb+\supset+ & $\supset$ &\verb+\subsetneq+ & $\subsetneq$ &\verb+\ni+ & $\ni$\\
\verb+\subset+ & $\subset$ &\verb+\supseteq+ & $\supseteq$ &\verb+\notin+ & $\notin$\\

        \bottomrule
    \end{tabular}

    \caption{18 set related symbols that were used for evaluation.}
    \label{table:symbols-used-for-evaluation-8}
\end{table}
\clearpage\clearpage
\subsection{Evaluation Results}
\begin{table}[h]
    \centering
    \begin{tabular}{crrr}
    \toprule
    \multirow{2}{*}{System}  & \multicolumn{3}{c}{Classification error}\\
          & std                    & TOP3                  & merged \\\midrule
    $B_1$ & $\SI{23.34}{\percent}$ & $\SI{6.80}{\percent}$ & $\SI{6.64}{\percent}$ \\
    $B_1$ & $\SI{23.12}{\percent}$ & $\SI{6.71}{\percent}$ & $\SI{6.58}{\percent}$ \\
    $B_1$ & $\SI{23.44}{\percent}$ & $\SI{6.72}{\percent}$ & $\SI{6.57}{\percent}$ \\
    $B_1$ & $\SI{23.18}{\percent}$ & $\SI{6.67}{\percent}$ & $\SI{6.54}{\percent}$ \\
    $B_1$ & $\SI{23.08}{\percent}$ & $\SI{6.75}{\percent}$ & $\SI{6.64}{\percent}$ \\\midrule
    $B_2$ & $\SI{21.51}{\percent}$ & $\SI{5.75}{\percent}$ & $\SI{5.67}{\percent}$ \\
    $B_2$ & $\SI{21.45}{\percent}$ & $\SI{5.68}{\percent}$ & $\SI{5.60}{\percent}$ \\
    $B_2$ & $\SI{21.80}{\percent}$ & $\SI{5.74}{\percent}$ & $\SI{5.66}{\percent}$ \\
    $B_2$ & $\SI{21.83}{\percent}$ & $\SI{5.75}{\percent}$ & $\SI{5.68}{\percent}$ \\
    $B_2$ & $\SI{21.58}{\percent}$ & $\SI{5.75}{\percent}$ & $\SI{5.66}{\percent}$ \\\midrule
    $B_3$ & $\SI{21.93}{\percent}$ & $\SI{5.74}{\percent}$ & $\SI{5.64}{\percent}$ \\
    $B_3$ & $\SI{22.28}{\percent}$ & $\SI{5.82}{\percent}$ & $\SI{5.75}{\percent}$ \\
    $B_3$ & $\SI{21.80}{\percent}$ & $\SI{5.74}{\percent}$ & $\SI{5.58}{\percent}$ \\
    $B_3$ & $\SI{21.74}{\percent}$ & $\SI{5.50}{\percent}$ & $\SI{5.41}{\percent}$ \\
    $B_3$ & $\SI{21.54}{\percent}$ & $\SI{5.50}{\percent}$ & $\SI{5.41}{\percent}$ \\\midrule
    $B_4$ & $\SI{23.88}{\percent}$ & $\SI{6.12}{\percent}$ & $\SI{6.04}{\percent}$ \\
    $B_4$ & $\SI{24.84}{\percent}$ & $\SI{6.44}{\percent}$ & $\SI{6.21}{\percent}$ \\
    $B_4$ & $\SI{23.84}{\percent}$ & $\SI{6.17}{\percent}$ & $\SI{6.02}{\percent}$ \\
    $B_4$ & $\SI{23.93}{\percent}$ & $\SI{6.31}{\percent}$ & $\SI{6.13}{\percent}$ \\
    $B_4$ & $\SI{23.19}{\percent}$ & $\SI{5.98}{\percent}$ & $\SI{5.83}{\percent}$ \\
    \bottomrule
    \end{tabular}
    \caption{The influence of random weight initialization. This table is
             summed up on \cpageref{sec:random-weight-initialization}.}
    \label{table:baseline-systems-random-initializations}
\end{table}

\begin{table}[h]
    \centering
    \begin{tabular}{lrrrrrr}
    \toprule
    \multirow{2}{*}{System}          & \multicolumn{6}{c}{Classification error}\\
    \cmidrule(l){2-7}
               & TOP1                   & change  & TOP3    & change  & MER  & change  \\\midrule
    $B_1$      & $\SI{23.34}{\percent}$ &         &  $\SI{6.80}{\percent}$ &              &  $\SI{6.64}{\percent}$ & \\
    $B_2$      & \underline{$\SI{21.51}{\percent}$} &  &  $\SI{5.75}{\percent}$&          &  $\SI{5.67}{\percent}$ & \\
    $B_3$      & $\SI{21.93}{\percent}$ &         &  $\SI{5.74}{\percent}$ &              &  $\SI{5.64}{\percent}$ & \\
    $B_4$      & $\SI{23.88}{\percent}$ &         &  $\SI{6.12}{\percent}$ &              &  $\SI{6.04}{\percent}$ & \\\midrule
    $B_{1,NSS}$& $\SI{40.51}{\percent}$ & $\SI{17.17}{\percent}$ & $\SI{17.83}{\percent}$ & $\SI{11.03}{\percent}$ & $\SI{17.16}{\percent}$ & $\SI{10.52}{\percent}$ \\
    $B_{2,NSS}$& $\SI{32.27}{\percent}$ & $\SI{10.76}{\percent}$ & $\SI{10.19}{\percent}$ & $\SI{+4.44}{\percent}$ & $\SI{10.07}{\percent}$ & $\SI{+4.40}{\percent}$ \\
    $B_{3,NSS}$& $\SI{34.31}{\percent}$ & $\SI{12.38}{\percent}$ & $\SI{11.66}{\percent}$ & $\SI{+5.92}{\percent}$ & $\SI{11.52}{\percent}$ & $\SI{+5.88}{\percent}$ \\
    $B_{4,NSS}$& $\SI{60.33}{\percent}$ & $\SI{36.45}{\percent}$ & $\SI{32.34}{\percent}$ & $\SI{26.22}{\percent}$ & $\SI{31.34}{\percent}$ & $\SI{25.30}{\percent}$ \\\midrule
    $B_{1,I2}$ & $\SI{22.75}{\percent}$ & $\SI{-0.59}{\percent}$ &  $\SI{6.40}{\percent}$ & $\SI{-0.40}{\percent}$ &  $\SI{6.24}{\percent}$ & $\SI{-0.40}{\percent}$ \\
    $B_{2,I2}$ & $\SI{21.52}{\percent}$ & $\SI{+0.01}{\percent}$ &  $\SI{5.42}{\percent}$ & $\SI{-0.33}{\percent}$ &  $\SI{5.31}{\percent}$ & $\SI{-0.36}{\percent}$ \\
    $B_{3,I2}$ & $\SI{21.73}{\percent}$ & $\SI{-0.20}{\percent}$ & \underline{$\SI{5.20}{\percent}$} & $\SI{-0.54}{\percent}$ & \underline{$\SI{5.09}{\percent}$} & $\SI{-0.55}{\percent}$ \\
    $B_{4,I2}$ & $\SI{28.20}{\percent}$ & $\SI{+4.32}{\percent}$ &  $\SI{7.77}{\percent}$ & $\SI{+1.65}{\percent}$ &  $\SI{7.29}{\percent}$ & $\SI{+1.25}{\percent}$ \\\midrule
    $B_{1,I3}$ & $\SI{22.77}{\percent}$ & $\SI{-0.57}{\percent}$ &  $\SI{6.67}{\percent}$ & $\SI{-0.13}{\percent}$ &  $\SI{6.53}{\percent}$ & $\SI{-0.11}{\percent}$ \\
    $B_{2,I3}$ & $\SI{21.86}{\percent}$ & $\SI{+0.35}{\percent}$ &  $\SI{5.87}{\percent}$ & $\SI{+0.12}{\percent}$ &  $\SI{5.80}{\percent}$ & $\SI{+0.13}{\percent}$ \\
    $B_{3,I3}$ & $\SI{21.80}{\percent}$ & $\SI{-0.13}{\percent}$ &  $\SI{5.95}{\percent}$ & $\SI{+0.21}{\percent}$ &  $\SI{5.84}{\percent}$ & $\SI{+0.20}{\percent}$ \\
    $B_{4,I3}$ & $\SI{23.55}{\percent}$ & $\SI{-0.33}{\percent}$ &  $\SI{6.11}{\percent}$ & $\SI{-0.01}{\percent}$ &  $\SI{6.03}{\percent}$ & $\SI{-0.01}{\percent}$ \\
    \bottomrule
    \end{tabular}
    \caption{The baseline models $B_1$--$B_4$ were tested with all three
             implementations of the scale and shift preprocessing algorithm.
             After every error score is indicated how much the system changed in
             comparison to its baseline. A change of $\SI{-0.05}{\percent}$
             means that the system improved by $\SI{0.05}{\percent}$ compared to
             its baseline system. The column \enquote{change} was left blank as
             implementation~1 was used in the baseline systems. Implementation~2
             does not center the recording and implementation~3 does center the
             recording on both axes. The $NSS$ models used \textit{n}o
             \textit{s}cale and \textit{s}hift algorithm. The results of this
             table are discussed on \cpageref{scale-and-shift-implementations}.}
\label{table:scale-and-shift-variations}
\end{table}

\begin{table}[h]
    \centering
    \begin{tabular}{lrrrrrr}
    \toprule
    \multirow{2}{*}{System} & \multicolumn{6}{c}{Classification error}\\
    \cmidrule(l){2-7}
                                         & TOP1                   & change                 & TOP3                   & change                 & MER                & change \\\midrule
    $B_{1,\theta_{sc} = \SI{5}{\pixel}}$ & $\SI{23.27}{\percent}$ & $\SI{-0.07}{\percent}$ &  $\SI{6.50}{\percent}$ & $\SI{-0.30}{\percent}$ & $\SI{6.37}{\percent}$ & $\SI{-0.27}{\percent}$ \\
    $B_{2,\theta_{sc} = \SI{5}{\pixel}}$ & $\SI{21.20}{\percent}$ & $\SI{-0.31}{\percent}$ &  $\SI{5.59}{\percent}$ & $\SI{-0.16}{\percent}$ & $\SI{5.50}{\percent}$ & $\SI{-0.17}{\percent}$ \\
    $B_{3,\theta_{sc} = \SI{5}{\pixel}}$ & $\SI{21.80}{\percent}$ & $\SI{-0.13}{\percent}$ &  $\SI{5.54}{\percent}$ & $\SI{-0.20}{\percent}$ & $\SI{5.47}{\percent}$ & $\SI{-0.17}{\percent}$ \\
    $B_{4,\theta_{sc} = \SI{5}{\pixel}}$ & $\SI{24.29}{\percent}$ & $\SI{+0.41}{\percent}$ &  $\SI{6.10}{\percent}$ & $\SI{-0.02}{\percent}$ & $\SI{5.94}{\percent}$ & $\SI{-0.10}{\percent}$ \\\midrule
    $B_{1,\theta_{sc} = \SI{10}{\pixel}}$& $\SI{23.17}{\percent}$ & $\SI{-0.17}{\percent}$ &  $\SI{6.61}{\percent}$ & $\SI{-0.19}{\percent}$ & $\SI{6.47}{\percent}$ & $\SI{-0.17}{\percent}$ \\
    $B_{2,\theta_{sc} = \SI{10}{\pixel}}$& \underline{$\SI{20.97}{\percent}$} & $\SI{-0.54}{\percent}$ &  $\SI{5.43}{\percent}$ & $\SI{-0.32}{\percent}$ & $\SI{5.34}{\percent}$ & $\SI{-0.33}{\percent}$ \\
    $B_{3,\theta_{sc} = \SI{10}{\pixel}}$& $\SI{21.34}{\percent}$ & $\SI{-0.59}{\percent}$ & \underline{$\SI{5.42}{\percent}$} & $\SI{-0.32}{\percent}$ & \underline{$\SI{5.33}{\percent}$} & $\SI{-0.31}{\percent}$ \\
    $B_{4,\theta_{sc} = \SI{10}{\pixel}}$& $\SI{23.50}{\percent}$ & $\SI{-0.38}{\percent}$ &  $\SI{6.11}{\percent}$ & $\SI{-0.01}{\percent}$ & $\SI{5.81}{\percent}$ & $\SI{-0.23}{\percent}$ \\\midrule
    $B_{1,\theta_{sc} = \SI{20}{\pixel}}$& $\SI{22.81}{\percent}$ & $\SI{-0.53}{\percent}$ &  $\SI{6.28}{\percent}$ & $\SI{-0.52}{\percent}$ & $\SI{6.19}{\percent}$ & $\SI{-0.45}{\percent}$ \\
    $B_{2,\theta_{sc} = \SI{20}{\pixel}}$& $\SI{21.61}{\percent}$ & $\SI{+0.10}{\percent}$ &  $\SI{5.79}{\percent}$ & $\SI{+0.04}{\percent}$ & $\SI{5.69}{\percent}$ & $\SI{+0.02}{\percent}$ \\
    $B_{3,\theta_{sc} = \SI{20}{\pixel}}$& $\SI{21.71}{\percent}$ & $\SI{-0.22}{\percent}$ &  $\SI{5.55}{\percent}$ & $\SI{-0.19}{\percent}$ & $\SI{5.45}{\percent}$ & $\SI{-0.19}{\percent}$ \\
    $B_{4,\theta_{sc} = \SI{20}{\pixel}}$& $\SI{24.36}{\percent}$ & $\SI{+0.48}{\percent}$ &  $\SI{6.23}{\percent}$ & $\SI{+0.11}{\percent}$ & $\SI{5.93}{\percent}$ & $\SI{-0.11}{\percent}$ \\
    \bottomrule
    \end{tabular}
    \caption{The baseline models $B_1$--$B_4$ with additionally applied stroke
             connect algorithm, before the scale and shift algorithm with
             different thresholds $\theta_{sc}$. The results of this table
             are discussed on \cpageref{subsec:stroke-connect}.}
\label{table:stroke-connect-evaluation}
\end{table}

\begin{table}[h]
    \centering
    \begin{tabular}{lrrrrrrr}
    \toprule
    \multirow{2}{*}{System}  & \multicolumn{6}{c}{Classification error}\\
    \cmidrule(l){2-7}
                                          & TOP1                   & change                  & TOP3                   & change                 & MER                & change \\\midrule
    $B_{1,\varepsilon=0.05,\text{linear}}$& $\SI{22.87}{\percent}$ & $\SI{-0.47}{\percent}$ & $\SI{6.68}{\percent}$ & $\SI{-0.12}{\percent}$ & $\SI{6.55}{\percent}$ & $\SI{-0.09}{\percent}$\\
    $B_{2,\varepsilon=0.05,\text{linear}}$& \underline{$\SI{21.24}{\percent}$} & $\SI{-0.27}{\percent}$ & \underline{$\SI{5.67}{\percent}$} & $\SI{-0.08}{\percent}$ & \underline{$\SI{5.57}{\percent}$} & $\SI{-0.10}{\percent}$\\
    $B_{3,\varepsilon=0.05,\text{linear}}$& $\SI{21.88}{\percent}$ & $\SI{-0.05}{\percent}$ & $\SI{6.01}{\percent}$ & $\SI{+0.27}{\percent}$ & $\SI{5.92}{\percent}$ & $\SI{+0.28}{\percent}$\\
    $B_{4,\varepsilon=0.05,\text{linear}}$& $\SI{23.84}{\percent}$ & $\SI{-0.04}{\percent}$ & $\SI{6.58}{\percent}$ & $\SI{+0.46}{\percent}$ & $\SI{6.25}{\percent}$ & $\SI{+0.21}{\percent}$\\\midrule
    $B_{1,\varepsilon=0.05,\text{cubic}}$ & $\SI{25.26}{\percent}$ & $\SI{+1.92}{\percent}$ & $\SI{8.77}{\percent}$ & $\SI{+1.97}{\percent}$ & $\SI{8.73}{\percent}$ & $\SI{+2.09}{\percent}$\\
    $B_{2,\varepsilon=0.05,\text{cubic}}$ & $\SI{23.84}{\percent}$ & $\SI{+1.91}{\percent}$ & $\SI{7.59}{\percent}$ & $\SI{+1.84}{\percent}$ & $\SI{7.54}{\percent}$ & $\SI{+1.87}{\percent}$\\
    $B_{3,\varepsilon=0.05,\text{cubic}}$ & $\SI{23.95}{\percent}$ & $\SI{+2.02}{\percent}$ & $\SI{7.49}{\percent}$ & $\SI{+1.75}{\percent}$ & $\SI{7.42}{\percent}$ & $\SI{+1.78}{\percent}$\\
    $B_{4,\varepsilon=0.05,\text{cubic}}$ & $\SI{29.47}{\percent}$ & $\SI{+6.13}{\percent}$ & $\SI{9.96}{\percent}$ & $\SI{+3.84}{\percent}$ & $\SI{9.68}{\percent}$ & $\SI{+3.64}{\percent}$\\\midrule
    $B_{1,\varepsilon=0.1,\text{linear}}$ & $\SI{23.81}{\percent}$ & $\SI{+0.47}{\percent}$ & $\SI{7.04}{\percent}$ & $\SI{+0.24}{\percent}$ & $\SI{6.91}{\percent}$ & $\SI{+0.27}{\percent}$\\
    $B_{2,\varepsilon=0.1,\text{linear}}$ & $\SI{22.02}{\percent}$ & $\SI{+0.51}{\percent}$ & $\SI{5.95}{\percent}$ & $\SI{+0.20}{\percent}$ & $\SI{5.88}{\percent}$ & $\SI{+0.21}{\percent}$\\
    $B_{3,\varepsilon=0.1,\text{linear}}$ & $\SI{22.10}{\percent}$ & $\SI{+0.17}{\percent}$ & $\SI{5.80}{\percent}$ & $\SI{+0.06}{\percent}$ & $\SI{5.72}{\percent}$ & $\SI{+0.08}{\percent}$\\
    $B_{4,\varepsilon=0.1,\text{linear}}$ & $\SI{25.05}{\percent}$ & $\SI{+1.17}{\percent}$ & $\SI{6.78}{\percent}$ & $\SI{+0.66}{\percent}$ & $\SI{6.42}{\percent}$ & $\SI{+0.38}{\percent}$\\\midrule
    $B_{1,\varepsilon=0.2,\text{linear}}$ & $\SI{28.08}{\percent}$ & $\SI{+4.74}{\percent}$ & $\SI{8.60}{\percent}$ & $\SI{+1.80}{\percent}$ & $\SI{8.48}{\percent}$ & $\SI{+1.84}{\percent}$\\
    $B_{2,\varepsilon=0.2,\text{linear}}$ & $\SI{25.82}{\percent}$ & $\SI{+4.31}{\percent}$ & $\SI{7.38}{\percent}$ & $\SI{+1.63}{\percent}$ & $\SI{7.24}{\percent}$ & $\SI{+1.57}{\percent}$\\
    $B_{3,\varepsilon=0.2,\text{linear}}$ & $\SI{26.72}{\percent}$ & $\SI{+4.79}{\percent}$ & $\SI{7.42}{\percent}$ & $\SI{+1.68}{\percent}$ & $\SI{7.27}{\percent}$ & $\SI{+1.63}{\percent}$\\
    $B_{4,\varepsilon=0.2,\text{linear}}$ & $\SI{28.36}{\percent}$ & $\SI{+4.48}{\percent}$ & $\SI{7.90}{\percent}$ & $\SI{+1.78}{\percent}$ & $\SI{7.76}{\percent}$ & $\SI{+1.72}{\percent}$\\\midrule
    $B_{1,\varepsilon=0.2,\text{cubic}}$  & $\SI{30.98}{\percent}$ & $\SI{+7.64}{\percent}$ &$\SI{10.77}{\percent}$ & $\SI{+3.97}{\percent}$ &$\SI{10.62}{\percent}$ & $\SI{+3.98}{\percent}$\\
    $B_{2,\varepsilon=0.2,\text{cubic}}$  & $\SI{28.54}{\percent}$ & $\SI{+7.03}{\percent}$ & $\SI{9.16}{\percent}$ & $\SI{+3.41}{\percent}$ & $\SI{9.06}{\percent}$ & $\SI{+3.39}{\percent}$\\
    $B_{3,\varepsilon=0.2,\text{cubic}}$  & $\SI{28.94}{\percent}$ & $\SI{+7.01}{\percent}$ & $\SI{8.82}{\percent}$ & $\SI{+3.08}{\percent}$ & $\SI{8.66}{\percent}$ & $\SI{+3.02}{\percent}$\\
    $B_{4,\varepsilon=0.2,\text{cubic}}$  & $\SI{32.80}{\percent}$ & $\SI{+8.92}{\percent}$ &$\SI{10.25}{\percent}$ & $\SI{+4.13}{\percent}$ & $\SI{9.85}{\percent}$ & $\SI{+3.81}{\percent}$\\
    \bottomrule
    \end{tabular}
    \caption{The evaluation results of Douglas-Peucker smoothing show that a
             strong simplification (a high $\varepsilon$ value) gives much worse
             results. Cubic spline interpolation performed much worse than
             linear interpolation. Those results are explained on
             \cpageref{subsec:douglas-peucker-smoothing-evaluation}.}
\label{table:douglas-peucker-smoothing-evaluation}
\end{table}

\begin{table}[h]
    \centering
    \begin{tabular}{lrrrrrr}
    \toprule
    \multirow{2}{*}{System}  & \multicolumn{6}{c}{Classification error}\\
    \cmidrule(l){2-7}
                       & TOP1                   & change                 & TOP3                   & change                 & MER                 & change \\\midrule
    $B_{1, \eta=0.05}$ & $\SI{24.58}{\percent}$ & $\SI{+1.24}{\percent}$ & $\SI{ 7.95}{\percent}$ & $\SI{+1.15}{\percent}$ & $\SI{ 7.70}{\percent}$ & $\SI{+1.06}{\percent}$\\
    $B_{1, \eta=0.1}$  & $\SI{23.34}{\percent}$ &                        & $\SI{ 6.80}{\percent}$ &                        & $\SI{ 6.64}{\percent}$ & \\
    $B_{1, \eta=0.2}$  & $\SI{23.41}{\percent}$ & $\SI{+0.07}{\percent}$ & $\SI{ 6.66}{\percent}$ & $\SI{-0.14}{\percent}$ & $\SI{ 6.64}{\percent}$ & $\SI{+0.00}{\percent}$\\
    $B_{1, \eta=1}$    & $\SI{30.80}{\percent}$ & $\SI{+7.46}{\percent}$ & $\SI{12.09}{\percent}$ & $\SI{+5.29}{\percent}$ & $\SI{11.36}{\percent}$ & $\SI{+4.72}{\percent}$\\\midrule
    $B_{2, \eta=0.05}$ & $\SI{22.37}{\percent}$ & $\SI{+0.86}{\percent}$ & $\SI{ 6.24}{\percent}$ & $\SI{+0.49}{\percent}$ & $\SI{ 6.14}{\percent}$ & $\SI{+0.47}{\percent}$\\
    $B_{2, \eta=0.1}$  & \underline{$\SI{21.51}{\percent}$} &            & $\SI{ 5.75}{\percent}$ &                        & $\SI{ 5.67}{\percent}$ & \\
    $B_{2, \eta=0.2}$  & $\SI{22.39}{\percent}$ & $\SI{+0.88}{\percent}$ & $\SI{ 6.02}{\percent}$ & $\SI{+0.27}{\percent}$ & $\SI{ 5.95}{\percent}$ & $\SI{+0.28}{\percent}$\\
    $B_{2, \eta=1}$    & $\SI{30.27}{\percent}$ & $\SI{+8.76}{\percent}$ & $\SI{12.80}{\percent}$ & $\SI{+7.05}{\percent}$ & $\SI{11.29}{\percent}$ & $\SI{+5.62}{\percent}$\\\midrule
    $B_{3, \eta=0.05}$ & $\SI{22.81}{\percent}$ & $\SI{+0.88}{\percent}$ & $\SI{ 6.01}{\percent}$ & $\SI{+0.27}{\percent}$ & $\SI{ 5.89}{\percent}$ & $\SI{+0.25}{\percent}$\\
    $B_{3, \eta=0.1}$  & $\SI{21.93}{\percent}$ &                        & \underline{$\SI{ 5.74}{\percent}$} &            & \underline{$\SI{ 5.64}{\percent}$} & \\
    $B_{3, \eta=0.2}$  & $\SI{21.77}{\percent}$ & $\SI{-0.16}{\percent}$ & $\SI{ 5.83}{\percent}$ & $\SI{+0.09}{\percent}$ & $\SI{ 5.71}{\percent}$ & $\SI{+0.07}{\percent}$\\
    $B_{3, \eta=1}$    & $\SI{90.67}{\percent}$ & $\SI{68.74}{\percent}$ & $\SI{86.98}{\percent}$ & $\SI{81.24}{\percent}$ & $\SI{86.12}{\percent}$ & $\SI{80.48}{\percent}$\\\midrule
    $B_{4, \eta=0.05}$ & $\SI{25.23}{\percent}$ & $\SI{+1.94}{\percent}$ & $\SI{ 6.86}{\percent}$ & $\SI{+0.74}{\percent}$ & $\SI{ 6.74}{\percent}$ & $\SI{+0.70}{\percent}$\\
    $B_{4, \eta=0.1}$  & $\SI{23.88}{\percent}$ &                        & $\SI{ 6.12}{\percent}$ &                        & $\SI{ 6.04}{\percent}$ & \\
    $B_{4, \eta=0.2}$  & $\SI{23.29}{\percent}$ & $\SI{-0.59}{\percent}$ & $\SI{ 6.14}{\percent}$ & $\SI{+0.02}{\percent}$ & $\SI{ 5.98}{\percent}$ & $\SI{-0.06}{\percent}$\\
    $B_{4, \eta=1}$    & $\SI{99.17}{\percent}$ & $\SI{75.29}{\percent}$ & $\SI{98.85}{\percent}$ & $\SI{92.73}{\percent}$ & $\SI{98.85}{\percent}$ & $\SI{92.81}{\percent}$\\
    \bottomrule
    \end{tabular}
    \caption{Evaluation results of the systems $B_1$ -- $B_4$ with adjusted
             learning rates $\eta$. The column \enquote{change} was left blank
             for the baseline systems ($\eta = 0.1$), as this value will only be
             different from exactly $0$ due to random weight initialization. The
             results of this table are explained on
             \cpageref{subsec:learning-rate}.}
\label{table:learning-rate-choices}
\end{table}\clearpage

\section{Figures}\label{appendix:figures}

\setcounter{figure}{0}

\subsection{Scatterplots of Features}
\begin{figure}[H]
    \centering
    \includegraphics*[width=\textwidth]{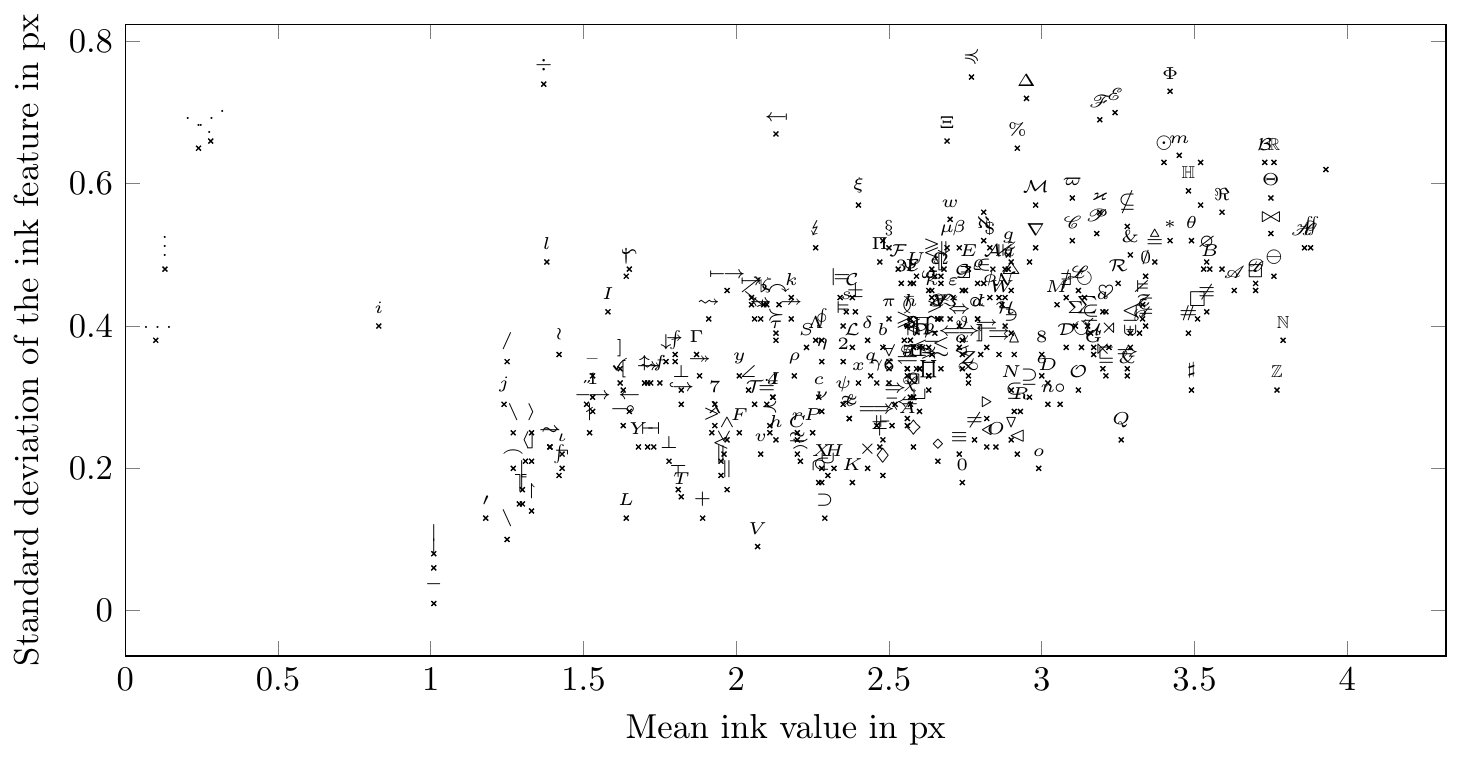}
    \caption{Mean-standard deviation scatterplot of the ink feature.
             Some symbols were excluded from this plot. They are listed in
             \cref{table:ink-cut-out}.
             This type of scatterplot was introduced on
             \cpageref{sec:features-evaluation}.}
    \label{fig:ink-mean-std-deviation}
\end{figure}

\begin{figure}[H]
    \centering
    \includegraphics*[width=\textwidth]{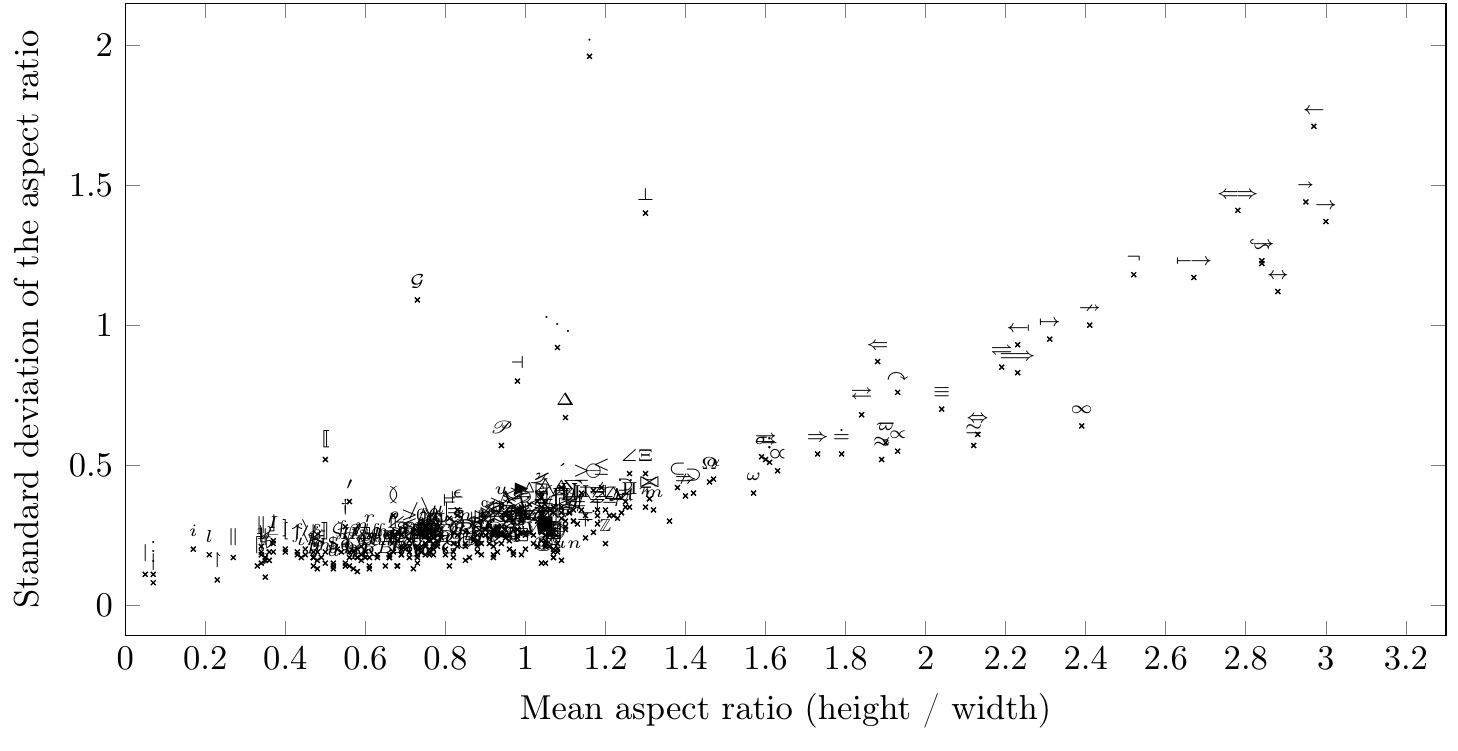}
    \caption{Mean-standard deviation scatterplot of the aspect ratio feature.
             This type of scatterplot was introduced on
             \cpageref{sec:features-evaluation}.}
    \label{fig:aspect-ratio-mean-std-deviation}
\end{figure}\clearpage

\section{Creative Users}\label{appendix:creative-users}

The following drawings made some creative users:
\begin{figure}[h]
    \centering
    \subfloat[ID 286218]{
        \includegraphics[height=0.2\linewidth, keepaspectratio]{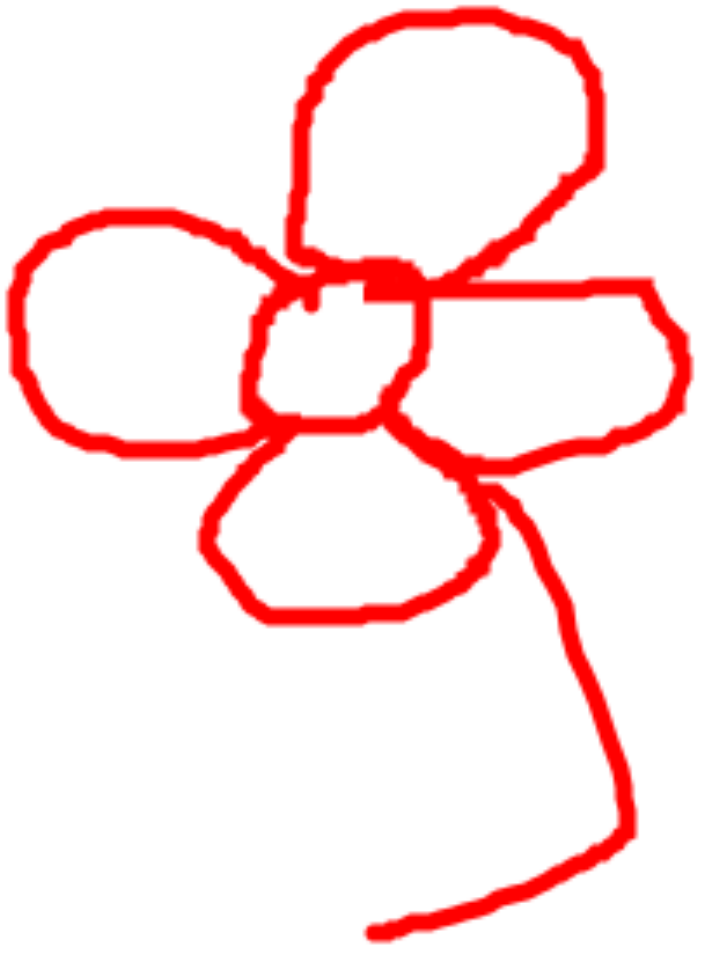}
        \label{fig:raw-data-id-286218}
    }%
    \subfloat[ID 271124]{
        \includegraphics[height=0.2\linewidth, keepaspectratio]{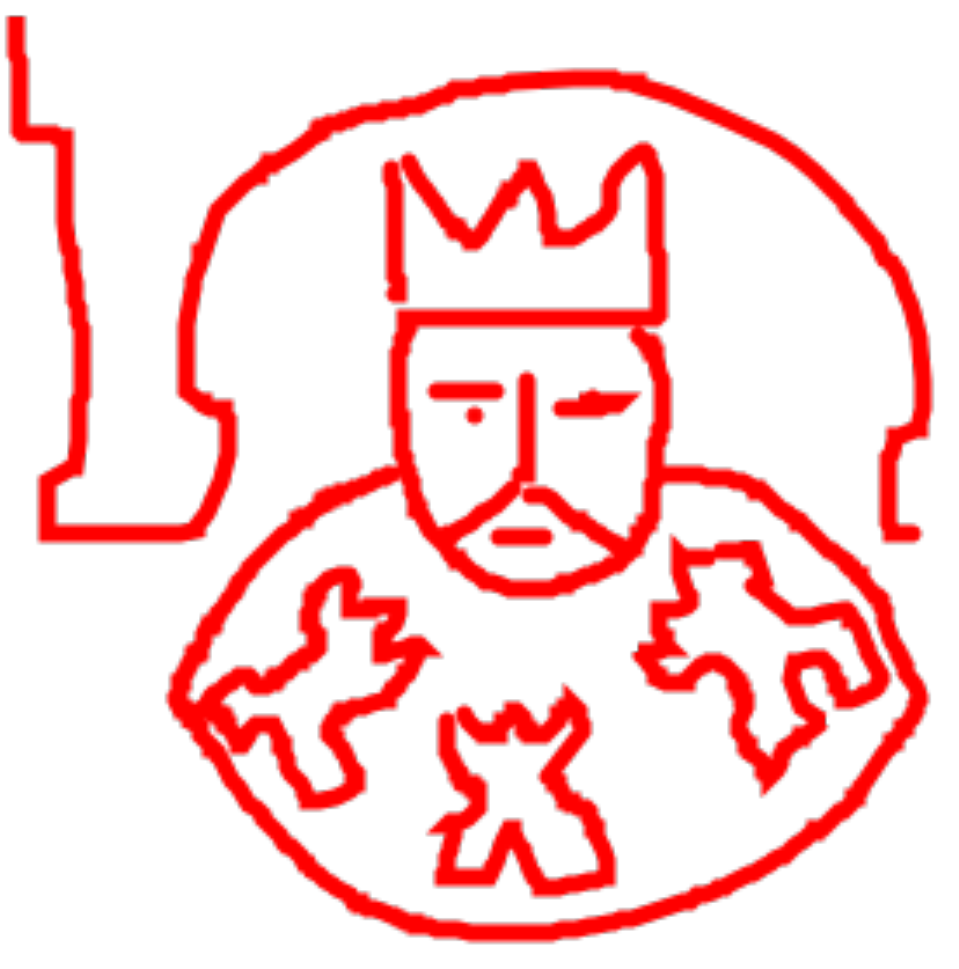}
        \label{fig:raw-data-id-271124}
    }%
    \subfloat[ID 278421]{
        \includegraphics[height=0.2\linewidth, keepaspectratio]{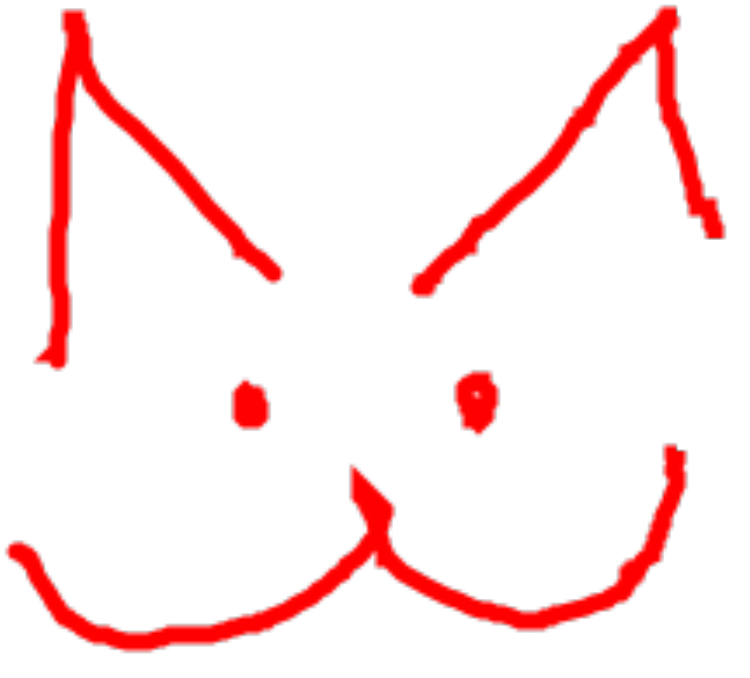}
        \label{fig:raw-data-id-278421}
    }%
    \subfloat[ID 280080]{
        \includegraphics[height=0.2\linewidth, keepaspectratio]{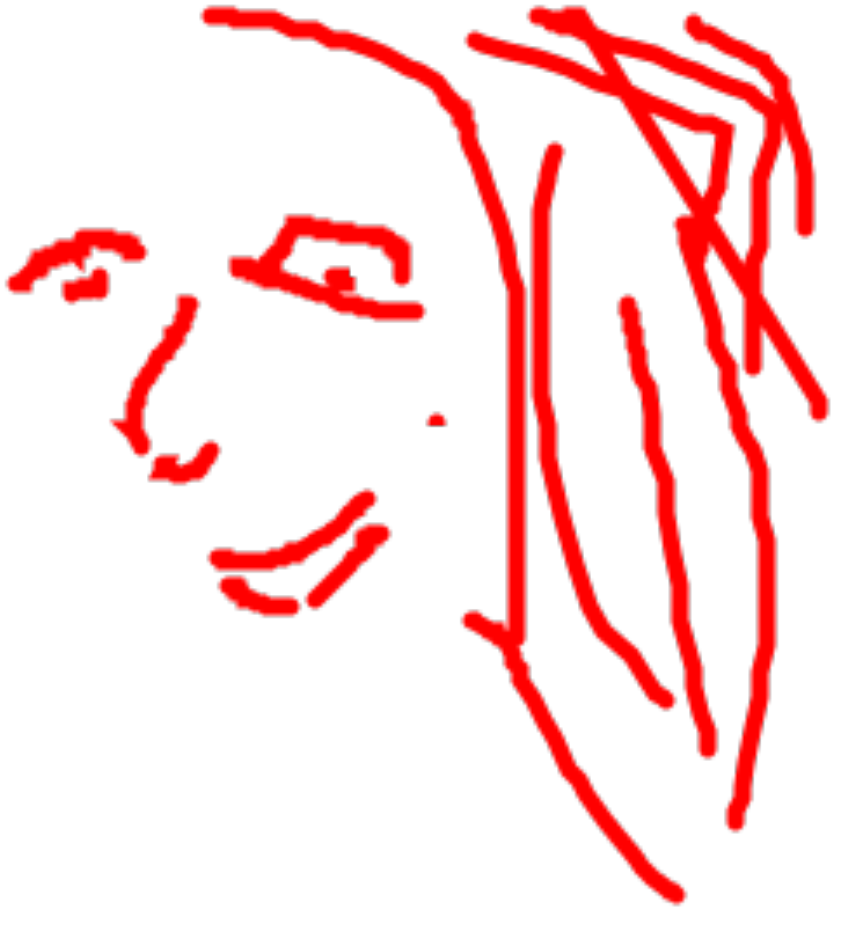}
        \label{fig:raw-data-id-280080}
    }

    \subfloat[ID 284768]{
        \includegraphics[height=0.2\linewidth, keepaspectratio]{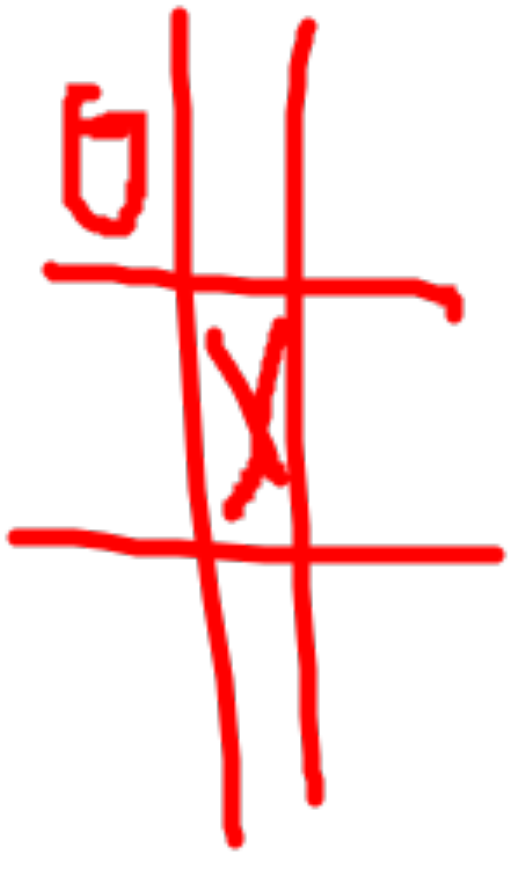}
        \label{fig:raw-data-id-284768}
    }%
    \subfloat[ID 282203]{
        \includegraphics[height=0.2\linewidth, keepaspectratio]{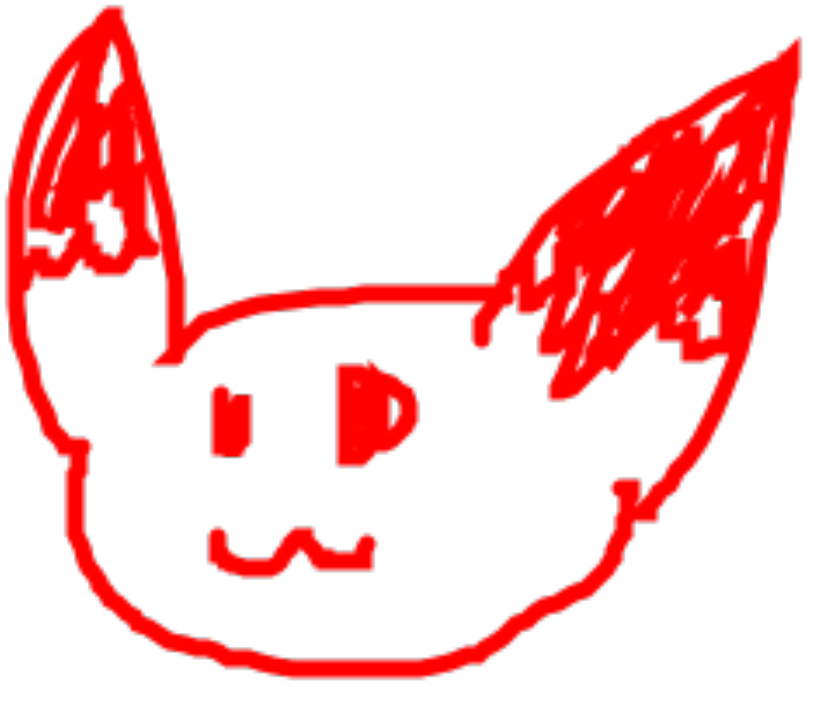}
        \label{fig:raw-data-id-282203}
    }%
    \subfloat[ID 266998]{
        \hspace*{1em}
        \includegraphics[height=0.2\linewidth, keepaspectratio]{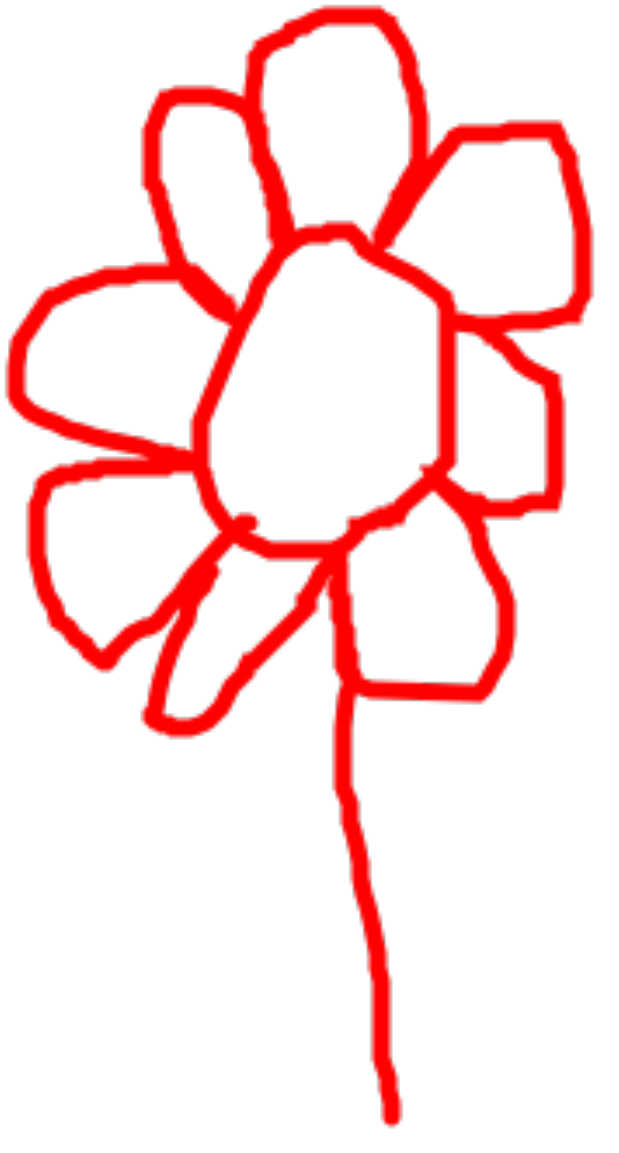}
        \hspace*{1em}
        \label{fig:raw-data-id-266998}
    }%
    \subfloat[ID 247571]{
        \hspace*{1em}
        \includegraphics[height=0.2\linewidth, keepaspectratio]{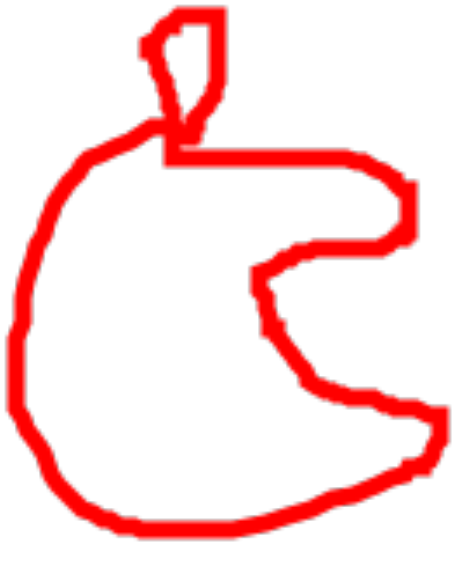}
        \hspace*{1em}
        \label{fig:raw-data-id-247571}
    }

    \subfloat[ID 208279]{
        \includegraphics[height=0.2\linewidth, keepaspectratio]{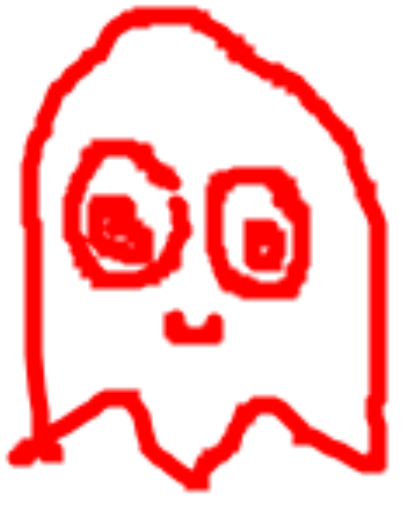}
        \label{fig:raw-data-id-208279}
    }%
    \subfloat[ID 191364]{
        \includegraphics[height=0.2\linewidth, keepaspectratio]{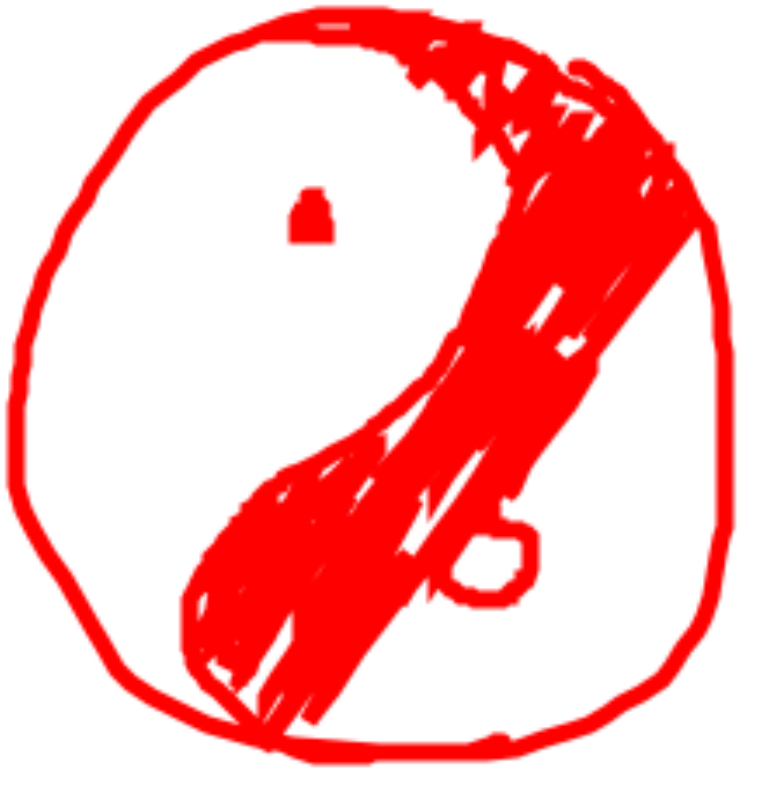}
        \label{fig:raw-data-id-191364}
    }%
    \subfloat[ID 215135]{
        \hspace*{1em}
        \includegraphics[height=0.2\linewidth, keepaspectratio]{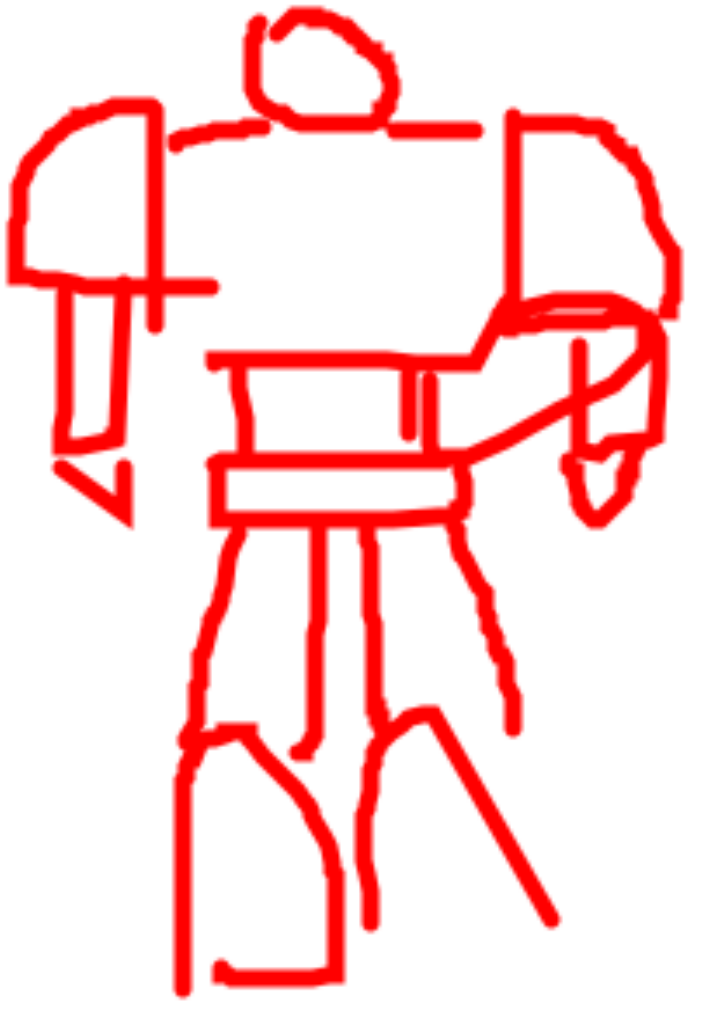}
        \hspace*{1em}
        \label{fig:raw-data-id-215135}
    }%
    \subfloat[ID 218757]{
        \hspace*{1em}
        \includegraphics[height=0.2\linewidth, keepaspectratio]{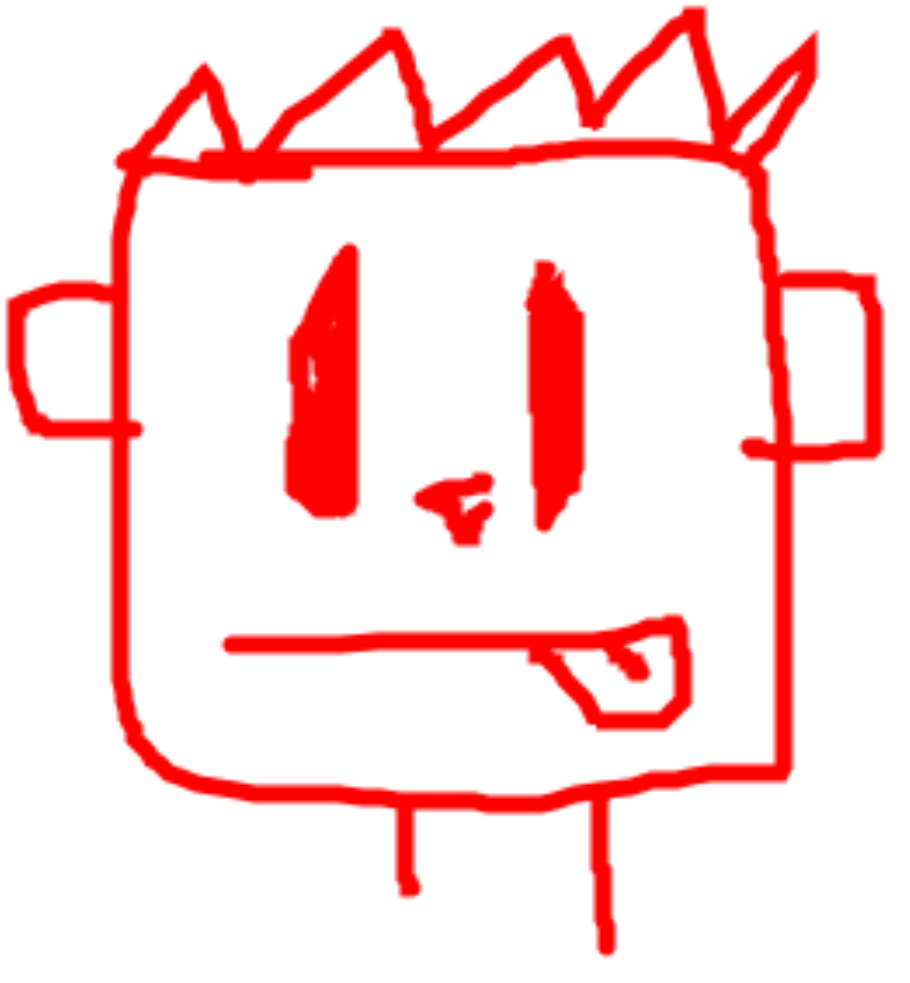}
        \hspace*{1em}
        \label{fig:raw-data-id-218757}
    }

    \subfloat[ID 167020]{
        \includegraphics[height=0.2\linewidth, keepaspectratio]{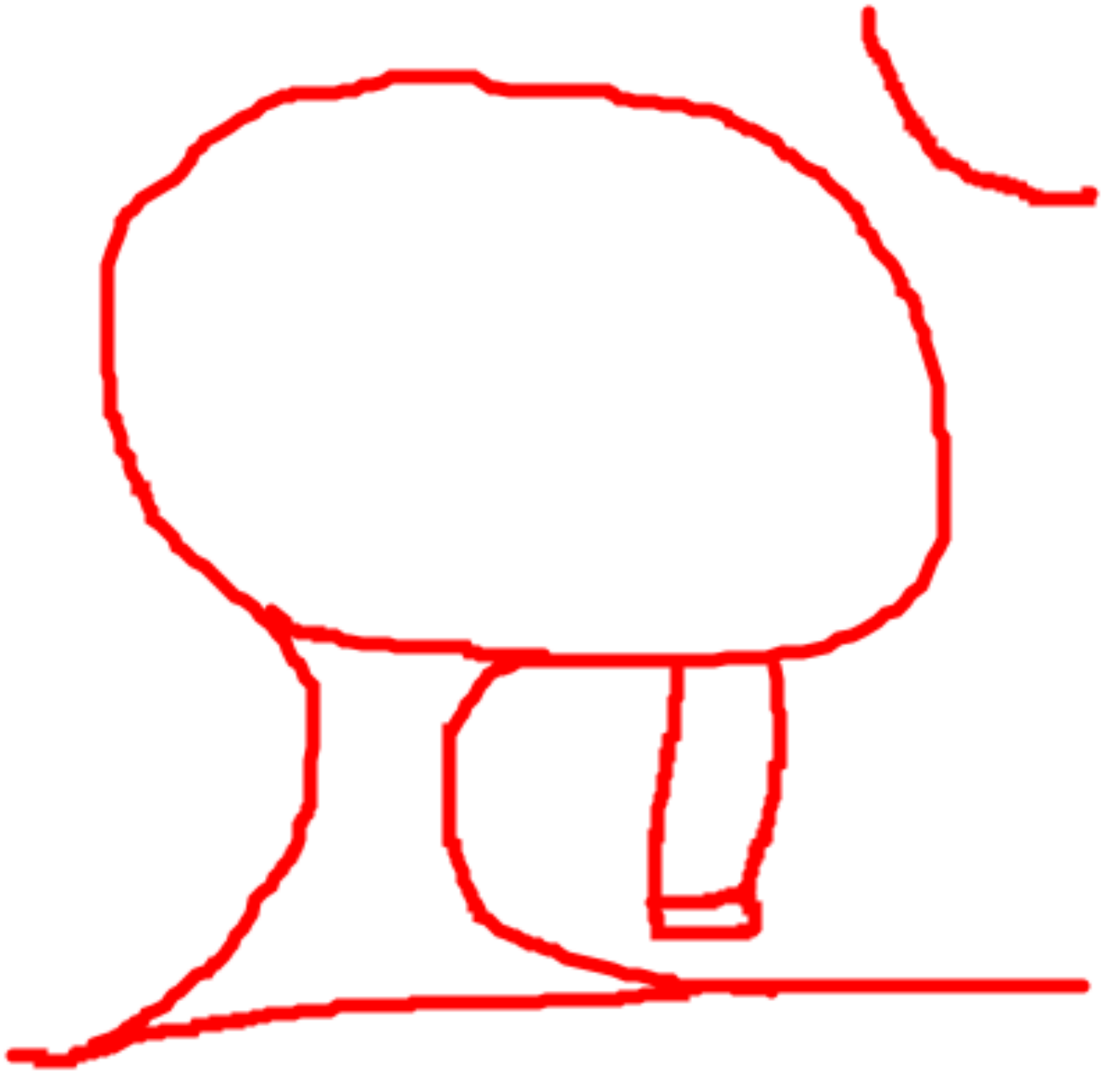}
        \label{fig:raw-data-id-167020}
    }%
    \subfloat[ID 230694]{
        \includegraphics[height=0.2\linewidth, keepaspectratio]{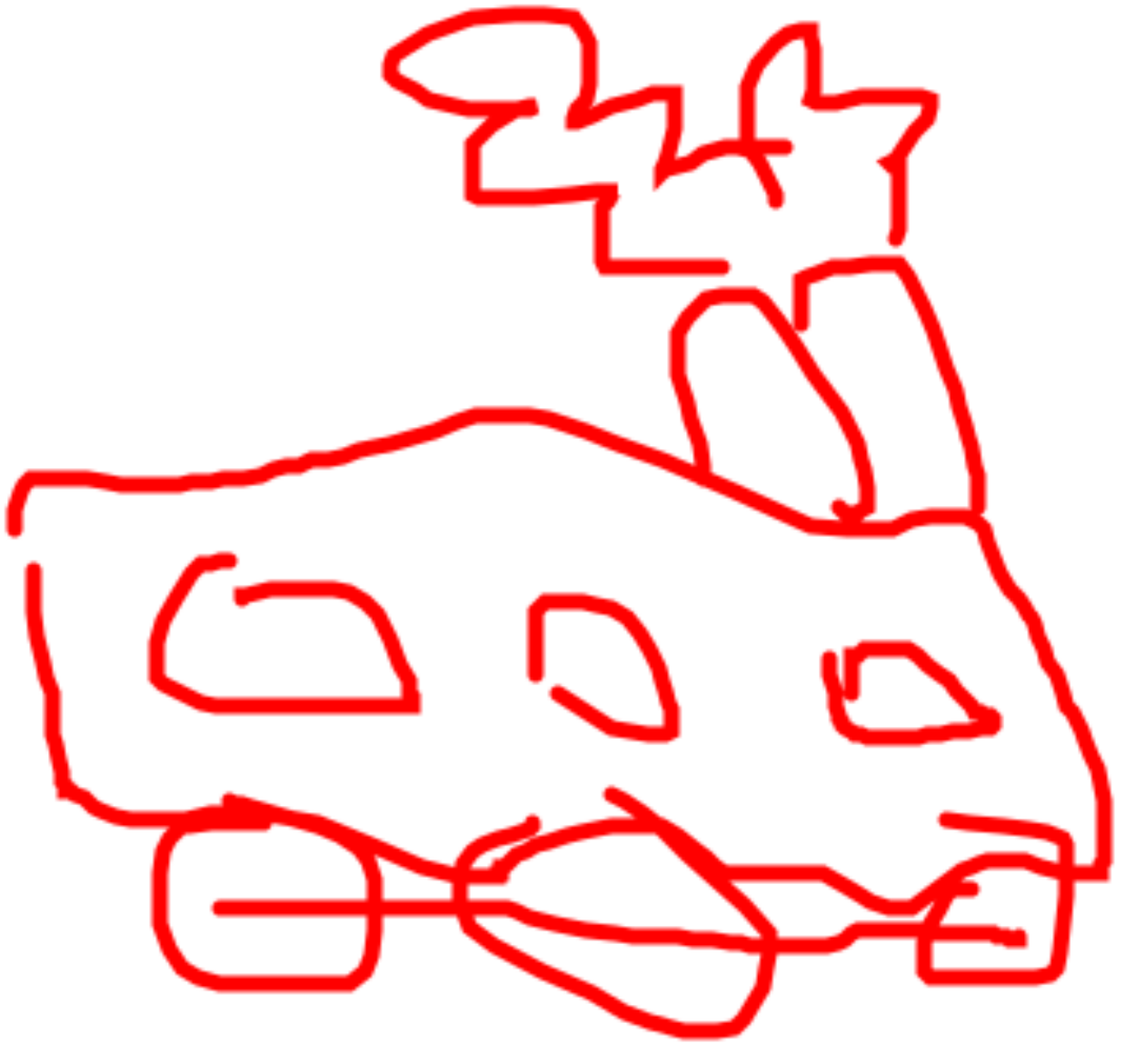}
        \label{fig:raw-data-id-230694}
    }%
    \subfloat[ID 230995]{
        \hspace*{1em}
        \includegraphics[height=0.2\linewidth, keepaspectratio]{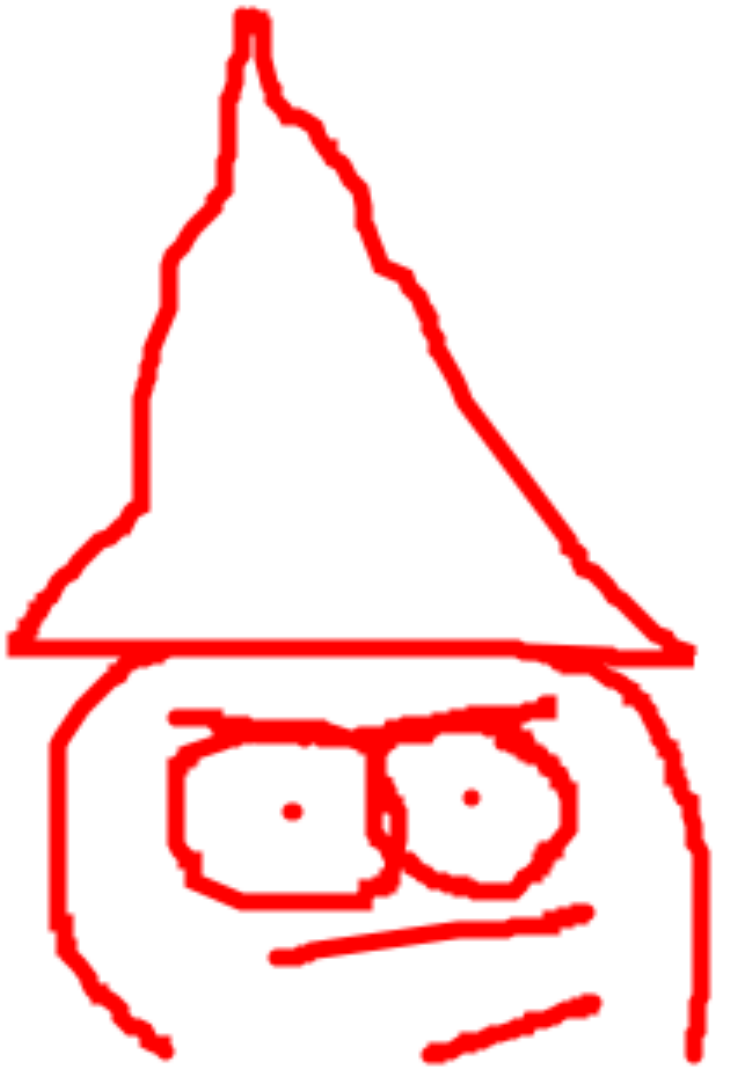}
        \hspace*{1em}
        \label{fig:raw-data-id-230995}
    }%
    \subfloat[ID 233035]{
        \includegraphics[height=0.2\linewidth, keepaspectratio]{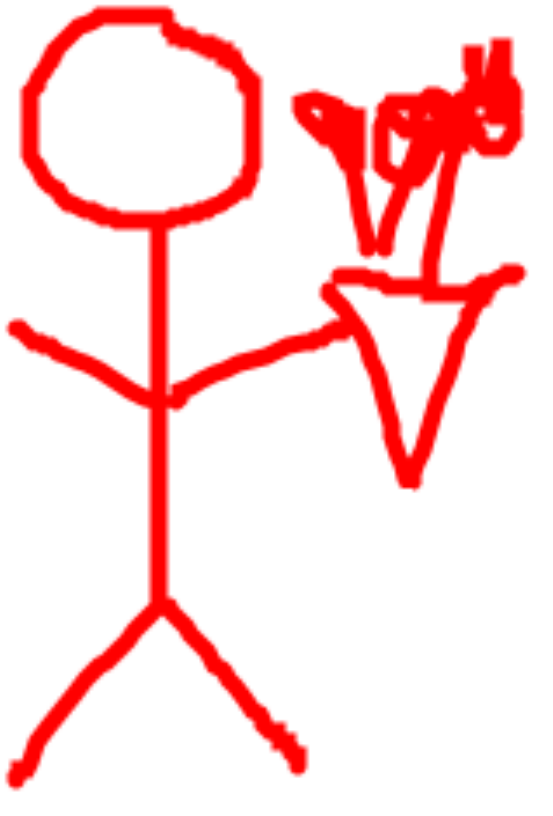}
        \label{fig:raw-data-id-233035}
    }

    \caption{Images drawn by creative users.}
    \label{fig:creative-users}
\end{figure}
\clearpage

\section{Raw Data Example}\label{appendix:raw-data-example}

The following code shows the recording with ID 292927 as it is stored in
the database. It is a JSON string that contains a list of strokes. Every
stroke is a list of control points where every control point has the \texttt{x}
and \texttt{y} coordinates as well as the \texttt{time}.

The symbol that was drawn is a $\subseteq$. So it has two strokes. This
recording has 145 control points.

The Unix time of the 28th of April 2014, 3~p.m. UTC would be $\num{1398636000}$.
The Unix time of $\num{1411732873010}$ is the number of milliseconds since 1970.
It is the 26th of September 2014 at 12:01:13~p.m. UTC.

\begin{lstlisting}[language=json,caption={292927.json},]
[[{"x":657,"y":600,"time":1411732873010},
  {"x":656,"y":600,"time":1411732873056},
  {"x":654,"y":599,"time":1411732873064},
  {"x":651,"y":599,"time":1411732873072},
  {"x":650,"y":598,"time":1411732873078},
  {"x":646,"y":598,"time":1411732873086},
  {"x":643,"y":598,"time":1411732873094},
  {"x":638,"y":598,"time":1411732873102},
  {"x":634,"y":598,"time":1411732873110},
  {"x":629,"y":598,"time":1411732873118},
  {"x":626,"y":597,"time":1411732873126},
  {"x":620,"y":597,"time":1411732873134},
  {"x":616,"y":597,"time":1411732873142},
  {"x":614,"y":597,"time":1411732873150},
  {"x":612,"y":596,"time":1411732873158},
  {"x":606,"y":596,"time":1411732873164},
  {"x":603,"y":596,"time":1411732873172},
  {"x":599,"y":596,"time":1411732873180},
  {"x":596,"y":596,"time":1411732873188},
  {"x":592,"y":597,"time":1411732873196},
  {"x":589,"y":599,"time":1411732873204},
  {"x":585,"y":599,"time":1411732873212},
  {"x":582,"y":600,"time":1411732873220},
  {"x":579,"y":600,"time":1411732873228},
  {"x":577,"y":602,"time":1411732873236},
  {"x":574,"y":603,"time":1411732873242},
  {"x":572,"y":605,"time":1411732873258},
  {"x":571,"y":605,"time":1411732873266},
  {"x":569,"y":607,"time":1411732873274},
  {"x":567,"y":608,"time":1411732873282},
  {"x":565,"y":609,"time":1411732873290},
  {"x":560,"y":611,"time":1411732873298},
  {"x":557,"y":613,"time":1411732873306},
  {"x":556,"y":613,"time":1411732873314},
  {"x":553,"y":615,"time":1411732873320},
  {"x":552,"y":616,"time":1411732873331},
  {"x":551,"y":616,"time":1411732873336},
  {"x":550,"y":617,"time":1411732873345},
  {"x":548,"y":620,"time":1411732873352},
  {"x":548,"y":622,"time":1411732873360},
  {"x":546,"y":623,"time":1411732873368},
  {"x":545,"y":627,"time":1411732873376},
  {"x":545,"y":629,"time":1411732873384},
  {"x":544,"y":633,"time":1411732873392},
  {"x":543,"y":636,"time":1411732873400},
  {"x":542,"y":642,"time":1411732873406},
  {"x":540,"y":647,"time":1411732873414},
  {"x":539,"y":653,"time":1411732873422},
  {"x":538,"y":657,"time":1411732873430},
  {"x":537,"y":659,"time":1411732873438},
  {"x":536,"y":664,"time":1411732873446},
  {"x":535,"y":669,"time":1411732873454},
  {"x":535,"y":670,"time":1411732873462},
  {"x":534,"y":674,"time":1411732873470},
  {"x":534,"y":675,"time":1411732873478},
  {"x":533,"y":680,"time":1411732873486},
  {"x":532,"y":684,"time":1411732873492},
  {"x":532,"y":689,"time":1411732873500},
  {"x":532,"y":690,"time":1411732873508},
  {"x":532,"y":691,"time":1411732873516},
  {"x":533,"y":693,"time":1411732873524},
  {"x":535,"y":695,"time":1411732873540},
  {"x":535,"y":696,"time":1411732873556},
  {"x":536,"y":696,"time":1411732873564},
  {"x":537,"y":697,"time":1411732873570},
  {"x":539,"y":698,"time":1411732873578},
  {"x":540,"y":699,"time":1411732873594},
  {"x":542,"y":700,"time":1411732873602},
  {"x":544,"y":700,"time":1411732873610},
  {"x":549,"y":701,"time":1411732873618},
  {"x":550,"y":701,"time":1411732873626},
  {"x":553,"y":701,"time":1411732873634},
  {"x":556,"y":701,"time":1411732873642},
  {"x":559,"y":701,"time":1411732873650},
  {"x":562,"y":701,"time":1411732873656},
  {"x":565,"y":701,"time":1411732873664},
  {"x":568,"y":701,"time":1411732873672},
  {"x":571,"y":702,"time":1411732873680},
  {"x":572,"y":702,"time":1411732873688},
  {"x":576,"y":702,"time":1411732873696},
  {"x":579,"y":702,"time":1411732873705},
  {"x":587,"y":702,"time":1411732873713},
  {"x":591,"y":702,"time":1411732873720},
  {"x":594,"y":702,"time":1411732873728},
  {"x":601,"y":702,"time":1411732873736},
  {"x":606,"y":702,"time":1411732873742},
  {"x":610,"y":702,"time":1411732873750},
  {"x":615,"y":702,"time":1411732873758},
  {"x":618,"y":702,"time":1411732873766},
  {"x":622,"y":702,"time":1411732873774},
  {"x":627,"y":702,"time":1411732873782},
  {"x":630,"y":702,"time":1411732873790},
  {"x":632,"y":702,"time":1411732873798},
  {"x":636,"y":702,"time":1411732873806},
  {"x":639,"y":702,"time":1411732873814},
  {"x":642,"y":702,"time":1411732873820},
  {"x":642,"y":701,"time":1411732873828},
  {"x":644,"y":701,"time":1411732873836},
  {"x":645,"y":701,"time":1411732873852},
  {"x":646,"y":701,"time":1411732873868},
  {"x":648,"y":700,"time":1411732873876},
  {"x":649,"y":700,"time":1411732873884},
  {"x":651,"y":700,"time":1411732873892},
  {"x":653,"y":700,"time":1411732873900},
  {"x":656,"y":700,"time":1411732873906},
  {"x":657,"y":700,"time":1411732873914},
  {"x":658,"y":700,"time":1411732873922}],
 [{"x":524,"y":741,"time":1411732874446},
  {"x":526,"y":741,"time":1411732874462},
  {"x":527,"y":741,"time":1411732874470},
  {"x":529,"y":741,"time":1411732874478},
  {"x":532,"y":740,"time":1411732874484},
  {"x":537,"y":740,"time":1411732874492},
  {"x":539,"y":740,"time":1411732874500},
  {"x":543,"y":740,"time":1411732874508},
  {"x":548,"y":740,"time":1411732874516},
  {"x":550,"y":740,"time":1411732874524},
  {"x":558,"y":740,"time":1411732874532},
  {"x":567,"y":740,"time":1411732874540},
  {"x":575,"y":740,"time":1411732874548},
  {"x":580,"y":740,"time":1411732874556},
  {"x":587,"y":740,"time":1411732874564},
  {"x":591,"y":740,"time":1411732874570},
  {"x":599,"y":740,"time":1411732874578},
  {"x":602,"y":740,"time":1411732874586},
  {"x":610,"y":739,"time":1411732874594},
  {"x":615,"y":739,"time":1411732874602},
  {"x":621,"y":738,"time":1411732874610},
  {"x":628,"y":738,"time":1411732874618},
  {"x":633,"y":738,"time":1411732874626},
  {"x":638,"y":738,"time":1411732874634},
  {"x":646,"y":738,"time":1411732874642},
  {"x":652,"y":738,"time":1411732874650},
  {"x":655,"y":738,"time":1411732874656},
  {"x":661,"y":738,"time":1411732874664},
  {"x":664,"y":738,"time":1411732874672},
  {"x":671,"y":738,"time":1411732874680},
  {"x":676,"y":738,"time":1411732874688},
  {"x":681,"y":739,"time":1411732874696},
  {"x":686,"y":740,"time":1411732874704},
  {"x":692,"y":741,"time":1411732874712},
  {"x":697,"y":742,"time":1411732874720},
  {"x":702,"y":742,"time":1411732874728},
  {"x":705,"y":742,"time":1411732874734},
  {"x":706,"y":742,"time":1411732874742}]]
\end{lstlisting}


\section{HWRT Handbook}\label{appendix:hwrt-handbook}

The Python package \texttt{hwrt} can be installed via pip:

\begin{verbatim}
# pip install hwrt
\end{verbatim}

The toolkit requires a configuration file \texttt{~/.hwrtrc} that contains
your projects root folder and the name of your neural network toolkit:

\begin{verbatim}
root: /home/moose/Downloads/write-math
nntoolkit: programname
\end{verbatim}

After that, it can be checked via command line if the installation worked:

\begin{verbatim}
$ hwrt --version
hwrt 0.1.150
\end{verbatim}

The project development hosted on \url{https://github.com/MartinThoma/hwrt}.

\texttt{hwrt 0.1.X} works in your projects root folder. Inside of
\texttt{project root} it looks for the following folders

\begin{itemize}
    \item \texttt{raw-datasets}: Flat folder that contains one \texttt{info.yml}
          and the raw datasets as \texttt{.pickle} files. Pickle is the
          standard way to serialize objects in Python.
    \item \texttt{preprocessed}: Folder that contains other folders. Each folder
          describes one specific way to preprocess data as well as the raw data
          source within a \texttt{info.yml} and contains the preprocessed files
          as \texttt{.pickle} files.
    \item \texttt{feature-files}: Folder that contains other folders. Each
          folder describes a set of features and the data source that should be
          used within a \texttt{info.yml}. The feature-files are created in
          those folders in the \texttt{.pfile} format.
    \item \texttt{models}:  Folder that contains other folders. Each
          folder contains an \texttt{info.yml} that describes the feature file
          data source, the model and how to train the model.
\end{itemize}

The \gls{YAML} configuration file for the preprocessing queue,
\texttt{info.yml}, looks like this:

\begin{verbatim}
data-source: archive/raw-datasets/2014-08-26-20-14-handwriting_datasets-raw.pickle
queue:
  - RemoveDuplicateTime: null
  - StrokeConnect:
      - minimum_distance: 10
  - ScaleAndShift:
      - max_width: 1.0
      - max_height: 1.0
      - center: true
  - SpaceEvenlyPerStroke:
      - kind: linear
      - number: 20
  - ScaleAndShift:
      - max_width: 1.0
      - max_height: 1.0
      - center: true
\end{verbatim}

The \texttt{queue} is ordered and can contain duplicate elements. All features
that are classes in \texttt{hwrt/preprocessing.py} can be used in this list. The
\texttt{data-source} is relative to the project root folder.

The \gls{YAML} configuration file for features, \texttt{info.yml}, looks like this:

\begin{verbatim}
data-source: archive/preprocessed/c2
data-multiplication:
  - Multiply:
      - nr: 1
features:
  - ConstantPointCoordinates:
      - strokes: 4
      - points_per_stroke: 20
      - fill_empty_with: 0
      - pen_down: false
  - ReCurvature:
      - strokes: 4
  - Ink: null
  - StrokeCount: null
  - AspectRatio: null
\end{verbatim}

All features that are classes in \texttt{hwrt/features.py} can be used in this
list.

The model \texttt{info.yml} looks like this:

\begin{verbatim}
data-source: archive/feature-files/c2
training: '{{nntoolkit}} train --epochs 1000 --learning-rate 0.1
    --momentum 0.1
    {{training}} {{validation}}
    {{testing}} < {{src_model}} > {{target_model}} 2>> {{target_model}}.log'
model:
    type: mlp
    topology: 167:500:500:369
\end{verbatim}

The training parameter makes use of templates. \verb+{{nntoolkit}}+ gets
replaced by the string that was specified in $\sim$\verb+/.hwrtrc+,
\verb+{{training}}+ gets replaced by the training pfile, \verb+{{validation}}+
gets replaced by the validation pfile and \verb+{{testing}}+ gets
replaced by the testing pfile. The training algorithm looks for \texttt{model-[number].json}
and replace \verb+{{src_model}}+ by the latest model path.
\verb+{{target_model}}+ gets replaced by \texttt{model-[number+1].json}.

\section{Website}\label{appendix:website}

\begin{figure}[H]
    \centering
    \subfloat[Webpage where users can record their handwriting]{
        \hspace*{3em}
        \includegraphics[height=0.4\linewidth, keepaspectratio]{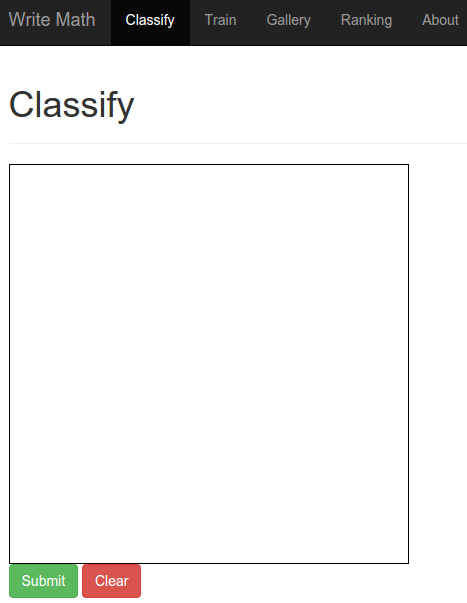}
        \hspace*{3em}
\label{fig:website-classify}
    }%
    \subfloat[Gallery page where the user can see what was drawn and what is
             unclassified]{
        \includegraphics*[height=0.4\linewidth,keepaspectratio]{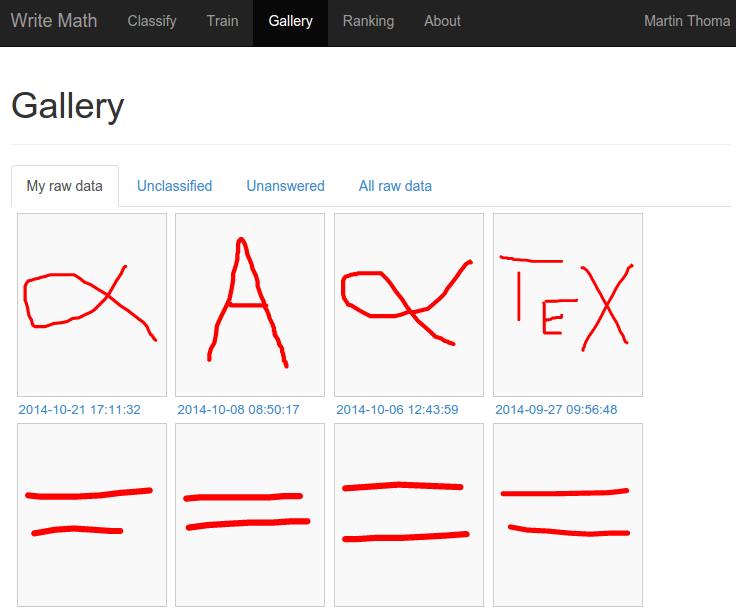}
\label{fig:website-gallery}
    }%
    \caption{Screenshots of different pages of \texttt{write-math.com}}
\label{fig:website-screenshots}
\end{figure}

\begin{figure}[H]
    \centering
    \includegraphics*[width=0.5\textwidth,keepaspectratio]{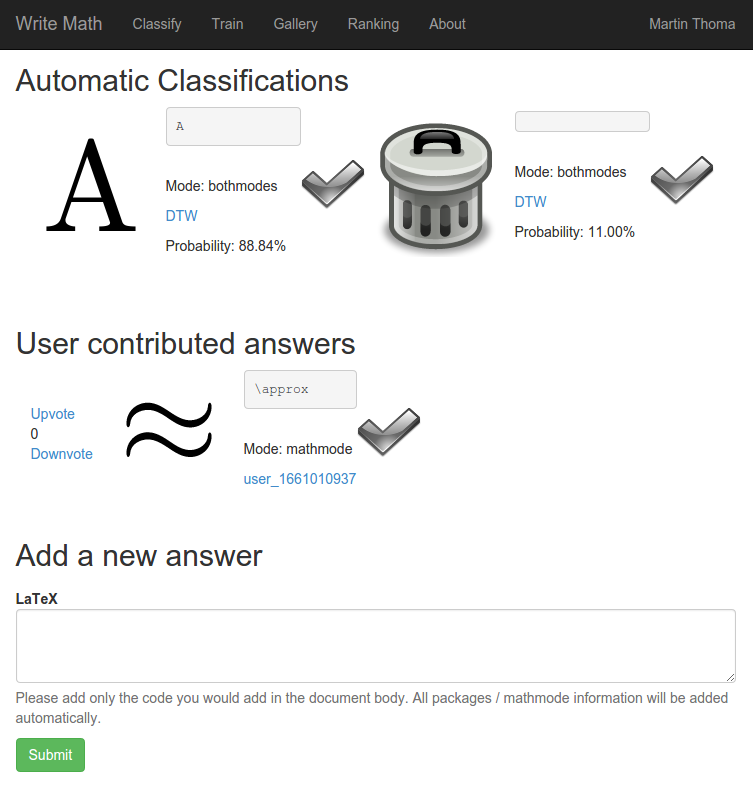}
    \caption{Page on which the user can see a recording and which symbols were
             suggested by automatic classifiers as well has human classifiers.
             The human classifications can get accepted and rated.}
\label{fig:website-classification-interface}
\end{figure}

\end{document}